\documentclass[sigconf]{acmart}

\def\BibTeX{{\rm B\kern-.05em{\sc i\kern-.025em b}\kern-.08emT\kern-.1667em\lower.7ex\hbox{E}\kern-.125emX}}

\copyrightyear{2024}
\acmYear{2024}
\setcopyright{rightsretained}
\acmConference[KDD '24]{Proceedings of the 30th ACM SIGKDD Conference on Knowledge Discovery and Data Mining}{August 25--29, 2024}{Barcelona, Spain}
\acmBooktitle{Proceedings of the 30th ACM SIGKDD Conference on Knowledge Discovery and Data Mining (KDD '24), August 25--29, 2024, Barcelona, Spain}\acmDOI{10.1145/3637528.3671758}
\acmISBN{979-8-4007-0490-1/24/08}

\usepackage{amsmath,amsfonts,bm}

\def\eqref#1{equation~\ref{#1}}

\def\1{\bm{1}}

\DeclareMathAlphabet{\mathsfit}{\encodingdefault}{\sfdefault}{m}{sl}
\SetMathAlphabet{\mathsfit}{bold}{\encodingdefault}{\sfdefault}{bx}{n}

\DeclareMathOperator*{\argmax}{arg\,max}

\usepackage{url}
\usepackage[utf8]{inputenc} %
\usepackage[T1]{fontenc}    %
\usepackage{booktabs}       %
\usepackage{amsfonts}       %
\usepackage{nicefrac}       %
\usepackage{microtype}      %
\usepackage{xcolor}         %
\usepackage{graphicx}
\usepackage{tabularx}
\usepackage{adjustbox}
\usepackage{multirow}
\usepackage{amsmath}
\usepackage{comment}
\usepackage{algorithm}
\usepackage{algpseudocode}
\usepackage{makecell}
\usepackage{subfigure}
\usepackage{wrapfig}
\let\oldsubfigure\subfigure
\renewcommand{\subfigure}[2][]{
  \oldsubfigure[\scriptsize\centering #1]{#2}
}

\usepackage{ulem}
\usepackage{subfloat}
\usepackage{algpseudocode}
\usepackage{xcolor}
\makeatletter
\newcommand{\bluecomment}[1]{\textcolor{blue}{\(\triangleright\) \textit{#1}}}
\algrenewcommand\algorithmiccomment[1]{\bluecomment{#1}}
\makeatother

\newtheorem{definition}{Definition}
\usepackage{cleveref}
\usepackage{enumitem}

\newcommand{\dshadow}{D_{s}}

\begin{document}

\title{Where Have You Been? A Study of Privacy Risk for Point-of-Interest Recommendation}

\author{Kunlin Cai}
\email{kunlin96@g.ucla.edu}
\affiliation{%
  \institution{University of California, Los Angeles}
  \country{}
}

\author{Jinghuai Zhang}
\email{jinghuai1998@g.ucla.edu}
\affiliation{%
  \institution{University of California, Los Angeles}
  \country{}
}

\author{Zhiqing Hong}
\email{zh252@cs.rutgers.edu}
\affiliation{%
  \institution{Rutgers University}
  \country{}
}

\author{William Shand}
\email{wss2ec@g.ucla.edu}
\affiliation{%
  \institution{University of California, Los Angeles}
  \country{}
}

\author{Guang Wang}
\email{guang@cs.fsu.edu}
\affiliation{%
  \institution{Florida State University}
  \country{}
}

\author{Desheng Zhang}
\email{desheng@cs.rutgers.edu}
\affiliation{%
  \institution{Rutgers University}
  \country{}
}

\author{Jianfeng Chi}
\email{jianfengchi@meta.com}
\affiliation{%
  \institution{Meta}
  \country{}
}

\author{Yuan Tian}
\email{yuant@ucla.edu}
\affiliation{%
  \institution{University of California, Los Angeles}
  \country{}
}

\renewcommand{\shortauthors}{Cai, et al.}

\begin{abstract}
As location-based services (LBS) have grown in popularity, more human mobility data has been collected. The collected data can be used to build machine learning (ML) models for LBS to enhance their performance and improve overall experience for users. However, the convenience comes with the risk of privacy leakage since this type of data might contain sensitive information related to user identities, such as home/work locations. Prior work focuses on protecting mobility data privacy during transmission or prior to release, lacking the privacy risk evaluation of mobility data-based ML models. To better understand and quantify the privacy leakage in mobility data-based ML models, we design a privacy attack suite containing data extraction and membership inference attacks tailored for point-of-interest (POI) recommendation models, one of the most widely used mobility data-based ML models. These attacks in our attack suite assume different adversary knowledge and aim to extract different types of sensitive information from mobility data, providing a holistic privacy risk assessment for POI recommendation models. Our experimental evaluation using two real-world mobility datasets demonstrates that current POI recommendation models are vulnerable to our attacks. We also present unique findings to understand what types of mobility data are more susceptible to privacy attacks. Finally, we evaluate defenses against these attacks and highlight future directions and challenges. Our attack suite is released at \textcolor{blue}{\url{https://github.com/KunlinChoi/POIPrivacy}}
\end{abstract}

\ccsdesc[300]{Security and privacy~Privacy-preserving protocols}
\ccsdesc[500]{Information systems~Location based services}
\keywords{POI recommendation; privacy-preserving machine learning; data extraction; membership inference}

\thanks{Correspondence to: Kunlin Cai, Yuan Tian, and Jianfeng Chi. Work unrelated to Meta.}

\maketitle

\section{Introduction} %

With the development and wide usage of mobile and wearable devices, large volumes of human mobility data are collected to support location-based services (LBS), such as traffic management~\citep{traffic_nips2020,lan2022dstagnn_traffic}, store location selection~\citep{site_rec_kdd17}, and point-of-interest (POI) recommendation~\citep{sun2020go, 10.1145/3477495.3531983}. In particular, POI recommendation involves relevant POI suggestions to users for future visits based on personal preferences using ML techniques~\citep{islam2020survey}, which has recently gained much research attention\footnote{From 2017 to 2023, there are more than 111 papers on POI recommendation built upon mobility data collected by location service providers~\citep{wang2023survey}.}. 
POI recommendation models have also been integrated into popular services such as Yelp and Google Maps to assist users in making informed decisions about the next destination to visit.
However, mobility data collected to train POI recommendation models are highly sensitive as they can leak users' sensitive information such as their social relationships, trip purposes, and identities~\cite {blumberg2009locational}.

\begin{figure}[h]
    \centering
    \includegraphics[width=0.43\textwidth]{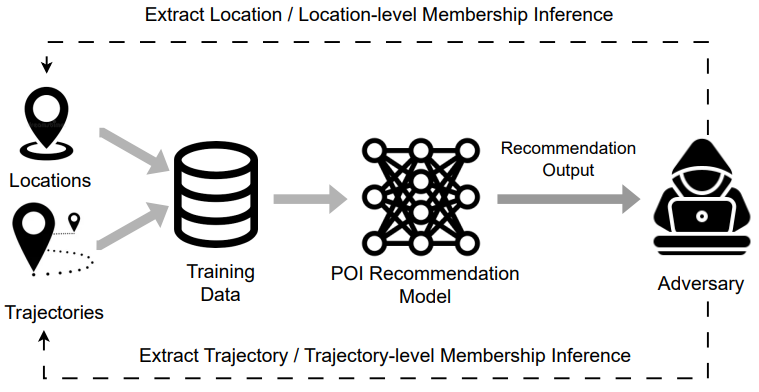}
    \vspace{-3mm}
    \caption{Our attack suite highlights the privacy concerns in POI recommendation models. In particular, we demonstrate that an adversary can extract or infer membership information of locations or trajectories in the training dataset.}
    \label{fig:IntroFigure}
    \vspace{-5mm}
\end{figure}
 
Although there are a significant number of studies~\citep{gedik2005location,krumm2007inference,andres2013geo,shokri2013hiding} on mobility data privacy, the existing research primarily focuses on analyzing attacks and evaluations within the context of mobility data transmission and release processes.
For example, previous studies have demonstrated the linkages of mobility data from various side channels, including social networks~\citep{henne2013snapme, hassan2018analysis}, open-source datasets~\citep{gambs2014anonymization,powar2023sok}, and network packets~\citep{jiang2007preserving, vratonjic2014location}. The linkages between these side channels can lead to the identification of individuals. As a result, efforts to protect mobility data have primarily concentrated on data aggregations and releases~\citep{gedik2005location,meyerowitz2009hiding,bordenabe2014optimal}.
These studies neglect the risk of adversaries extracting sensitive attributes or properties from the ML models (e.g., POI recommendation models) that use mobility data for training, which are inherently susceptible to privacy attacks~\citep{shokri2017membership,carlini2022membership}.

Evaluating privacy risks in POI recommendation models remains challenging because existing attack and defense mechanisms are ineffective due to the unique features of mobility data.
Previous privacy attacks have mainly focused on ML models trained with image and text data~\citep{shokri2017membership,10.1145/2810103.2813677,carlini2019secret}, where each data point can uniquely identify itself. However, mobility data, such as locations, are less semantically unique without the context.
Moreover, mobility data is special in that it contains multimodal spatial and temporal information, which describes each individual's movements and behavior patterns over time. All existing attacks fail to construct meaningful context and leverage spatial-temporal information, resulting in their failures when applied to POI recommendations.
Furthermore, existing defense mechanisms~\citep{ACGMM16, shi2022just, SCLJY22} have mainly been tested on classification models trained with image or text data. Given the task and data are significantly different, the effectiveness of defense mechanisms is unknown when applied to POI recommendation. %

In this paper, we design a comprehensive privacy attack suite to study the privacy leakage in POI recommendation models trained with mobility data. 
Specifically, our privacy attack suite contains the two most popular kinds of privacy attacks on machine learning models, data extraction and membership inference attacks, to assess the privacy vulnerabilities of POI recommendation models at both \textbf{location} and \textbf{trajectory} levels. 
In contrast to privacy attacks for image and text data, the attacks in our attack suite are tailored for mobility data and aim to extract different types of sensitive information based on practical adversary knowledge.

We perform experiments on three representative POI recommendation models trained on two mobility datasets. We demonstrate that POI recommendation models are vulnerable to our designed data extraction and membership inference attacks. We further provide an in-depth analysis to understand what factors affect the attack performance and contribute to the effectiveness of the attacks. 
Based on our analysis, we discover that the effect of data outliers exists in privacy attacks against POI
recommendations, making training examples with certain types of users, locations, and trajectories particularly vulnerable to the attacks in the attack suite.
Finally, We test several existing defenses and find that they do not effectively thwart our attacks with negligible utility loss, which calls for better methods to defend against our attacks.

\noindent
\textbf{Contributions:}~~ 
\begin{itemize}[leftmargin=*]
\item %
We introduce a novel privacy attack suite that incorporates unique characteristics of mobility data (e.g., spatial-temporal information) into the attack design. In particular, we target a previously under-defended attack surface: neural-network-based POI recommendation. To the best of our knowledge, our work is the first to comprehensively evaluate the privacy risks in POI recommendation models using inference attacks from both location and trajectory levels.
\item We conduct extensive experiments on state-of-the-art POI recommendation models and datasets to demonstrate that POI recommendation models are vulnerable to data extraction and membership inference attacks in our attack suite.
\item We provide an in-depth analysis to understand what unique factors in mobility data make them vulnerable to privacy attacks. We also explore the reason regarding how our attack design works and test existing defenses against our attacks. Our analysis identifies the challenges and future directions for developing privacy-preserving POI recommendation models.
\end{itemize}

\section{Background}
\subsection{Point-of-Interest Recommendation}
POI recommendation has recently gained much attention due to its importance in many business applications~\citep{islam2020survey}, such as user experience personalization and resource optimization. Initially, researchers focused on feature engineering and algorithms such as Markov chain~\citep{zhang2014lore,chen2014nlpmm}, matrix factorization algorithms~\citep{lian2014geomf,cheng2013you}, and Bayesian personalized ranking~\citep{he2017category,zhao2016stellar} for POI recommendation. However, more recent studies have shifted their attention towards employing neural networks like RNN~\citep{Liu2016PredictingTN,yang2020location}, LSTM~\citep{kong2018hst,sun2020go}, and self-attention models~\citep{luo2021stan,lian2020geography}. 
Neural networks can better learn from spatial-temporal correlation in mobility data (e.g., check-ins) to predict users' future locations and thus outperform other POI recommendation algorithms by a large margin. 
Meanwhile, this could introduce potential privacy leakage.
Thus, we aim to design an attack suite to measure the privacy risks of neural-network-based POI recommendations systematically.

We first provide the basics of POI recommendations and notations used throughout this paper. Let $\mathcal{U}$ be the user space, $\mathcal{L}$ be the location space, and $\mathcal{T}$ be the timestamp space. A POI recommendation model takes the observed trajectory of a user as input and predicts the next POI that will be visited, which is formulated as $f_\theta: \mathcal{U} \times \mathcal{L}^{n} \times \mathcal{T}^{n} \rightarrow  \mathbb{R}^{\lvert \mathcal{L} \rvert}$. Here, the length of the input trajectory is $n$. We denote a user by its user ID $u \in \mathcal{U}$ for simplicity.
For an input trajectory with $n$ check-ins, we
denote its \emph{trajectory sequence} as $x_{T}^{0:n-1}=\{(l_0, t_0), \dots, (l_{n-1},t_{n-1})\}$, where $l_i \in \mathcal{L}$ and $t_i \in \mathcal{T}$ indicate the POI location and corresponding time interval of $i$-th check-in. Also, the \emph{location sequence} of this trajectory is denoted as $x_{L}^{0:n-1}=\{l_0, \dots, l_{n-1}\}$. The POI recommendation model predicts the next location $l_{n}$ (also denoted as $y$ by convention) by outputting the logits of all the POIs. Then, the user can select the POI with the highest logit as its prediction $\hat{y}$, where $\hat{y} = \argmax f_\theta(u, x_{T}^{0:n-1})$. Given the training set $D_{\text{tr}}$ sampled from an underlying distribution ${\mathcal{D}}$, the model weights are optimized to minimize the prediction loss on the overall training data, i.e.,
$
         \min_{\theta} \frac{1}{|D_{\text{tr}}|} \sum_{ (u, x_{T}^{{0:n-1}}, y)\in D_{\text{tr}}} \ell(f_\theta(u, x_{T}^{0:n-1}), y),
$
where $\ell$ is the cross-entropy loss, i.e., $\ell(f_\theta(u, x_{T}^{0:n-1}), y) = -\log(f_\theta(u, x_{T}^{0:n-1}))_y$. 
The goal of the training process is to maximize the performance of the model on the unseen test dataset $D_{te} \in \mathcal{D}$, which is drawn from the same distribution as the training data. During inference, this prediction $\hat{y}$ is then compared to the next real location label $l_{n}$ to compute the prediction accuracy. 
The performance evaluation of POI recommendation models typically employs metrics such as top-$k$ accuracy (e.g., $k=1, 5, 10$).


\subsection{Threat Models}
\label{sec:threat_models}

\noindent
\textbf{Adversary Objectives.}~~
To understand the potential privacy leakage of training data in POI recommendation models, we design the following four attacks from the two most common privacy attack families: membership inference attack~\cite{shokri_2017_MIA} and data extraction attacks~\cite{carlini2021extracting}, based on the characteristics of the mobility data for POI recommendation, namely \textit{common location extraction} (\textsc{LocExtract}), \textit{training trajectory extraction} (\textsc{TrajExtract}), \textit{location-level membership inference attack} (\textsc{LocMIA}), and \textit{trajectory-level membership inference attack} (\textsc{TrajMIA}). 
These four attacks aim to extract or infer different sensitive information about a user in the POI recommendation model training data.

\textsc{LocExtract} focuses on extracting a user's most frequently visited location;
\textsc{TrajExtract} extracts a user's location sequence with a certain length given a starting location;
\textsc{LocMIA} infers whether a user has been to a location and used for training;
\textsc{TrajMIA} infers where a trajectory sequence has been used for training.

\noindent
\textbf{Adversary Knowledge.}~~For all attacks, we assume the attacker has access to the query interface of the victim model. Specifically, the attacker can query the victim model with the target user and obtain the corresponding output logits. This assumption is realistic in two scenarios: 
(1) A malicious third-party entity is granted access to the POI model query API hosted by the model owner (e.g., location service providers like Foursquare or Yelp) for specific businesses such as personalized advertisement. This scenario is well-recognized by~\cite{4sq_news2, 4sq_place, xin2021outoftown}.
(2) The retention period of the training data expires. Still, the model owner keeps the model and an adversary (e.g., a malicious insider of location service providers) can extract or infer the sensitive information using our attack suite, even if the training data have been deleted. In this scenario, the model owner may violate privacy regulations such as GDPR~\citep{gdpr}.

Depending on different attack objectives, the adversary also possesses different auxiliary knowledge.
In particular,
for \textsc{TrajExtract}, we assume the attacker can query the victim model with a starting location $l_0$ that the target user visited. This assumption is reasonable because an attacker can use real-world observation~\citep{vicente2011location, srivatsa2012deanonymizing}, \textsc{LocExtract}, and \textsc{LocMIA} as cornerstones.
As for \textsc{LocMIA} and \textsc{TrajMIA}, we assume the attacker has access to a shadow dataset following the standard settings of membership inference attacks~\citep{shokri2017membership,carlini2022membership}. 
\section{Attack Suite}
Our privacy attack suite includes the two most prevalent types of privacy attacks on machine learning models: data extraction and membership inference attacks. These are used to evaluate the privacy vulnerabilities of POI recommendation models at both location and trajectory levels. The subsequent sections detail the technical approaches and design of attacks, taking into account the unique aspects of mobility data.

\subsection{Data Extraction Attacks} 
\label{sec:training-data-extraction}

Our data extraction attacks are rooted in the idea that victim models display varying levels of memorization in different subsets of training data. By manipulating the spatial-temporal information in the queries, the attacker can extract users' locations or trajectories that these victim models predominantly memorize.

\noindent
\textbf{\textsc{LocExtract}}~~
Common location extraction attack ($\textsc{LocExtract}$) aims to extract a user's most frequently visited location in the victim model training, i.e.,  
$$
    \textsc{LocExtract}(f_\theta,u) \rightarrow \hat{l}_{top1},\dots,\hat{l}_{topk}.
$$
The attack takes the victim model $f_\theta$ and the target user $u$ as the inputs and generates $k$ predictions $\hat{l}_{top1},\dots,\hat{l}_{topk}$ to extract the most frequently visited location of user $u$. 
The attack is motivated by our key observation: querying POI recommendation models with a random location reveals that these models tend to ``over-learn'' a user's most frequently visited locations, making these locations more likely to appear in the model output.
For example, we randomly choose 10 users and query the victim model using 100 randomly selected locations. Of these queries, 32.5\% yield the most frequent location for the target user. Yet, these most common locations are present in only 18.7\% of these users' datasets.

In \textsc{LocExtract}, we first generate a set of different random inputs for a specific user and use them to make iterative queries to the victim model. Each query returns the prediction logits with a length of $|\mathcal{L}|$ outputted by the victim model. 
The larger the logit value, the more confident the model is in predicting the corresponding location as the next POI. Therefore, by iterating queries to the model given a target user and aggregating the logit values of all queries, the most visited location is more likely to have a large logit value after aggregation. 
Here, we use a soft voting mechanism, i.e., averaging the logit values of all queries, as the aggregation function (see also Sec.~\ref{result:parameter} for the comparison with different aggregation functions).
With the resulting mean logits, we output the top-$k$ locations with $k$ largest logit values as the attack results. 
Algorithm~\ref{alg:common_point} gives the outline of \textsc{LocExtract}. Though the attack is straightforward, it is effective and can be a stepping stone for \textsc{TrajExtract} in our attack suite.

\noindent
\textbf{\textsc{TrajExtract}}~~
Our training trajectory extraction attack \\($\textsc{TrajExtract}$) aims to extract the location sequence $x_L^{0:n-1} = \{l_0,\dots,l_{n-1}\}$ in a training trajectory of user $u$ with a length of $n$ from the victim model $f_\theta$. Formally,
$$
\textsc{TrajExtract}(f_\theta,u,l_0,n) \rightarrow \hat{x}_{L_0}^{0:n-1}, \dots, \hat{x}_{L_\beta}^{0:n-1},
$$
where $\hat{x}_{L_0}^{0:n-1}, \dots, \hat{x}_{L_\beta}^{0:n-1}$ indicate the top-$\beta$ extracted location sequences by the attack.

The key idea of the training trajectory extraction attack is to identify the location sequence with the lowest log perplexity, as models tend to demonstrate lower log perplexity when they see trained data. 
We denote log perplexity as:
$$
\label{formula:ppl2}
       \textsc{PPL}_{f_\theta}(u,x_T^{0:n-1}) = -\log{\mathrm{Pr}_{f_\theta}}(u,x_T^{0:n-1})
       = -\sum_{i=0}^{n-1}\log\mathrm{Pr}_{f_\theta}(u, x_T^{0:i-1}),
$$
where $\mathrm{Pr}_{f_\theta}(\cdot)$ is the likelihood of observing $x_T^{0:n-1}$ with user $u$ under the victim model $f_\theta$. In order to get the lowest log perplexity of location sequences with a length of $n$, we have to enumerate all possible location sequences. However, in the context of POI recommendation, there are $\mathcal{O}(\lvert \mathcal{L}\rvert^{n-1})$ possible location sequences for a given user. $\lvert \mathcal{L} \rvert$ equals the number of unique POIs within the mobility dataset and can include thousands of options. Thus, the cost of calculating the log perplexity of all location sequences can be very high.
To this end, we use beam search to extract the location sequences with both time and space complexity $\mathcal{O}(\lvert \mathcal{L}\rvert \times n \times \beta)$, where $\beta$ is the beam size. In particular, to extract a trajectory of length $n$, we iteratively query the victim model using a set of candidate trajectories with a size of $\beta$ and update the candidate trajectories until the extraction finishes. 
As highlighted in the prior work~\citep{fan2018hierarchical}, when using beam search to determine the final outcome of a sequential neural network, there is a risk of generating non-diverse outputs and resembling the training data sequence. 
However, in our scenario, this property can be leveraged as an advantage in \textsc{TrajExtract}, as our primary objective revolves around extracting the training location sequence with higher confidence.  
As a final remark, both \textsc{LocExtract} and \textsc{TrajExtract} need a query timestamp to query the victim model, and we will show the effects of the timestamp in our experiments.
Algorithm~\ref{alg:TrainingTraj} in Appendix~\ref{app:attack} gives the detailed steps of \textsc{TrajExtract}.

\subsection{Membership Inference Attacks}
\label{sec:membership-inference-attack}

Membership inference attack ($\textsc{MIA}$) %
aims to determine whether a target data sample is used in the model training. We extend the notion to infer whether certain sensitive information (e.g., user-location pair $(u, l)$ and trajectory sequence $(u, x_T)$) of the user's data is involved in the training of the victim model $f_\theta$. Since POI recommendation models use multi-modal sequential data as inputs and adversaries lack sufficient information to construct a complete input, we propose attack designs to manipulate the spatial-temporal information in queries to enhance effectiveness of attacks.
The membership inference attack can be formulated as follow:
$$
\textsc{MIA}(f_\theta,X_{target},\dshadow) \rightarrow \{\text{member},\text{nonmember}\},
$$
where $X_{target}$ represents the target sensitive information ($X_{target}=(u, l)$ in \textsc{LocMIA} and $X_{target}=(u, x_T)$ in \textsc{TrajMIA}), and $\dshadow$ is the shadow dataset owned by the adversary.

To effectively infer the membership of a given $X_{target}$, we adapt the state-of-the-art membership inference attack -- likelihood ratio attack (LiRA)~\citep{carlini2022membership} to the context of POI recommendation. The key insight of LiRA is that the model parameters trained with $X_{target}$ differ from those trained without it, and the effect of the model parameter on a data sample can be well approximated using a loss value.
By conducting a hypothesis test on the distributions of the loss values, we can identify if the victim model is trained with the $X_{target}$ or not. LiRA consists of four steps: (1) train multiple shadow models, (2) query the shadow models trained with $X_{target}$ and without $X_{target}$ to obtain two distributions, (3) query the victim model $X_{target}$ to obtain the output logits, and (4) conduct a $\Lambda$ hypothesis test to infer the membership of the $X_{target}$ based on the two distributions and the query results. Due to the space limit, we defer the details of LiRA to Appendix~\ref{app:attack}.

\noindent
\textbf{\textsc{LocMIA}}~~In this attack, the adversary aims to determine whether a given user $u$ has visited a location $l$ in the training data. However, it is not feasible to directly apply LiRA to \textsc{LocMIA} as the victim model takes the trajectory sequences as inputs, but the adversary only has a target location without the needed sequential context. 
In particular, \textsc{LocMIA} needs the auxiliary inputs to calculate the membership confidence score since this process cannot be completed only using $X_{target}=(u, l)$.  
This attack is a stark contrast to $\textsc{MIA}$ for image/text classification tasks where the $X_{target}$ itself is sufficient to compute the membership confidence score.

To this end, we design a spatial-temporal model query algorithm (Algorithm~\ref{alg:LocQuery} in Appendix~\ref{app:attack}) to tailor LiRA to \textsc{LocMIA} and optimize membership confidence score calculation. 
The idea behind the algorithm is that if a particular user has been to a certain POI location, the model might ``unintentionally'' memorize its neighboring POI locations and the corresponding timestamp in the training data.
Motivated by this, each time we query the models (e.g., the victim and shadow models), we generate $n_l$ random locations and $n_t$ fixed-interval timestamps. 
To obtain stable and precise membership confidence scores, we first average the corresponding confidence scores at the target location by querying with $n_l$ locations at the same timestamp. While the adversary does not possess the ground truth timestamp linked with the target POI for queries, the adversary aims to mimic a query close to the real training data. To achieve this, we repeat the same procedure of querying different locations for $n_t$ timestamps and take the maximum confidence scores among the $n_t$ averaged confidence scores as the final membership inference score for the target example. 
Algorithm~\ref{alg:lira_loc} gives the outline of LiRA in terms of \textsc{LocMIA}, and the lines marked with red are specific to \textsc{LocMIA}.

\noindent
\textbf{\textsc{TrajMIA}}~~
The attack aims to determine whether a trajectory is used in the training data of the victim model.
Unlike \textsc{LocMIA}, $X_{target}=(u, x_T)$ suffices to calculate the membership confidence score in LiRA, and we do not need any auxiliary inputs. 
To fully leverage information of the target example querying the victim model and improve the attack performance, we also utilize the $n-2$ intermediate outputs and the final output from the sequence $x_T$ with a length of $n$ to compute the membership confidence score, i.e., we take the average of all $n-1$ outputs. This change improves the attack performance as the intermediate outputs provide additional membership information for each point in the target trajectory. The purple lines in Algorithm~\ref{alg:lira_loc} highlight steps specific to
\textsc{TrajMIA}.

\subsection{Practical Implications of the Attack Suite}

Our attack suite is designed as an integrated framework focusing on the basic units of mobility data %
-- locations and trajectories. It contains two prevalent types of privacy attacks: data extraction and membership inference attacks. 
Each attack in our attack suite targets a specific type of mobility data and could serve as a privacy auditing tool~\citep{jagielski2020auditing}. 
They can also be used to infer additional sensitive information in mobility data: 
\begin{itemize}[leftmargin=*]
\item \textsc{LocExtract} extracts %
a user's most common location. Combined with the semantics of the POI, we may infer the user's address such as work address, which is closely related to user identity;
\item \textsc{TrajExtract} can be further used to infer user trajectories and identify trip purposes by analyzing the POIs visited during a journey~\citep{meng2017travel}; 
\item \textsc{LocMIA} can determine the membership of multiple POIs, thereby facilitating the inference of a user's activity range and social connections in~\citet{cho2011friendship,ren2023your};
\item \textsc{TrajMIA} infers if a user's trajectory is in the training dataset, which can serve as an auditing tool to examine the privacy leakage by assuming a worst-case adversary.
\end{itemize}

\begin{table}[t!]
\renewcommand{\arraystretch}{1.2}
\fontsize{8.5}{8.5}\selectfont
\centering
\caption{The performance of victim models.}
\vspace{-2mm}
\begin{tabular}{cccc}
\toprule
Dataset & Model & Top-$1$ ACC & Top-$10$ ACC \\
\midrule
\multirow{3}{*}{\textsc{4sq}} & \textsc{GETNext} & 0.34 & 0.71 \\ 
& \textsc{LSTPM} & 0.25 & 0.67 \\ 
& \textsc{RNN} & 0.24 & 0.68 \\
\midrule
\multirow{3}{*}{\textsc{Gowalla}} & \textsc{GETNext} & 0.16 & 0.48\\ 
& \textsc{LSTPM} & 0.15 & 0.39 \\ 
& \textsc{RNN} & 0.10 & 0.26 \\ 
\bottomrule
\end{tabular}
\label{model_utility}
\vspace{-3mm}
\end{table}

\section{Experiments}
We empirically evaluate the proposed attack suite to answer the following research questions: 
(1) What's the performance of the proposed attacks in extracting or inferring the sensitive information from POI recommendation models (Sec.~\ref{exp:overall_results})?
(2) What unique factors (e.g., user, location, trajectory) in mobility data correlate with the attack performance (Sec.~\ref{sec:dataspec})?
(3) How do different attack designs (e.g., spatial-temporal querying for membership inference attacks) improve the attack performance (Sec.~\ref{rq3})?

\subsection{Experimental Setup}

We briefly describe the datasets, models, and evaluation metrics used in our experiments. Due to the space limit, we defer the details of datasets (e.g., statistics of each dataset), data pre-processing pipeline, default training and attack parameters to Appendix~\ref{app:detail_exp}.

\textbf{Datasets}~~ 
\textit{Following the literature}~\citep{10.1145/3477495.3531983,kong2018hst}, we comprehensively evaluate four privacy attacks on two POI recommendation benchmarks: FourSquare (\textsc{4sq})~\citep{yang2014modeling} and \textsc{Gowalla}~\citep{cho2011friendship}.

\textbf{Models}~~ 
We experiment with three representative POI recommendation models, including \textsc{GETNext}~\citep{10.1145/3477495.3531983}, \textsc{LSTPM}~\citep{sun2020go}, and \textsc{RNN}~\citep{libcity}. 
Note that \textsc{GETNext} and \textsc{LSTPM} are the state-of-the-art POI recommendation methods based on the transformer and hierarchical \textsc{LSTM}, respectively. We also include \textsc{RNN} since it is a commonly used baseline for POI recommendation.

\textbf{Evaluation Metrics}~~ We use the top-$k$ extraction attack success rate (ASR) to evaluate the effectiveness of data extraction attacks. For \textsc{LocExtract}, the top-$k$ ASR is defined as $| U_{\mathrm{extracted}} | / |\mathcal{U}|$, where $U_{\mathrm{extracted}}$ is the set of users whose most visited locations are in the top-$k$ predictions outputted by our attack; For \textsc{TrajExtract} the top-$k$ ASR is $|\text{correct extractions}| / |\text{all } (u,l_0) \text{ pairs} |$, where correct extractions are $(u,l_0)$ pairs with top-$k$ extracted results matching an exact location sequence in the training data.

For \textsc{LocMIA} and \textsc{TrajMIA},
we utilize the commonly employed metrics for evaluating membership inference attacks, namely the area under the curve (AUC), average-case ``accuracy'' (ACC), and true positive rate (TPR) versus false positive rate (FPR) in the low-false positive rate regime. 
Our primary focus is the TPR versus FPR metric in the low-false positive rate regime because evaluating membership inference attacks should prioritize the worst-case privacy setting rather than average-case metrics, as emphasized in~\citep{carlini2022membership}.

\subsection{Experimental Results and Analysis}

\subsubsection{Attack performance (RQ1)} 
\label{exp:overall_results}

\begin{figure*}[t]
    \centering
    \subfigure [\textsc{LocExtract} (\textsc{4sq})]
    {\includegraphics[width=0.22\textwidth]{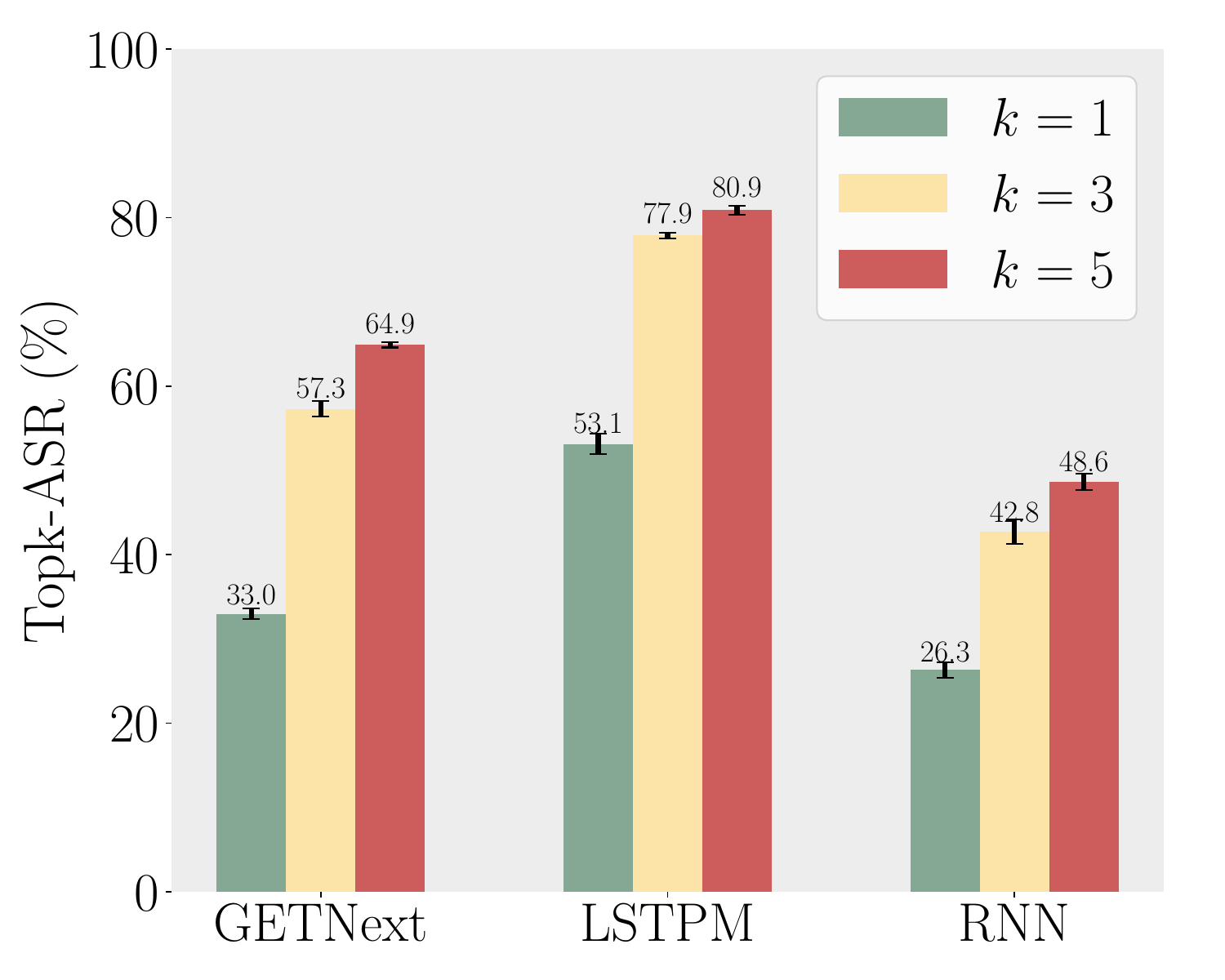}
    \label{fig:Attack1_results_4sq}}
    \subfigure[\textsc{LocExtract} (\textsc{Gowalla})] {\includegraphics[width=0.22\textwidth]{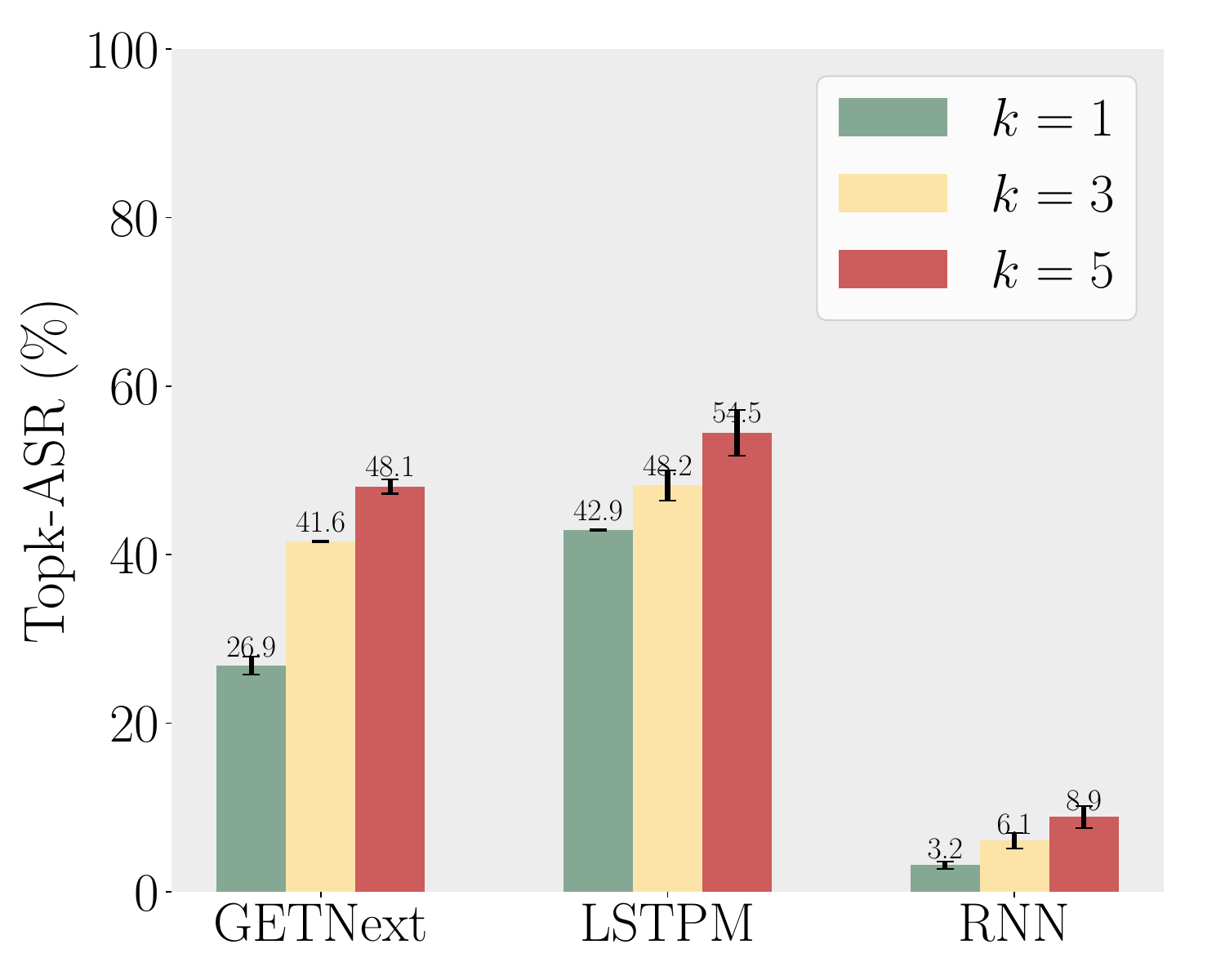}
    \label{fig:Attack1_results_gow}}
    \subfigure[\textsc{TrajExtract} (\textsc{4sq})] {\includegraphics[width=0.22\textwidth]{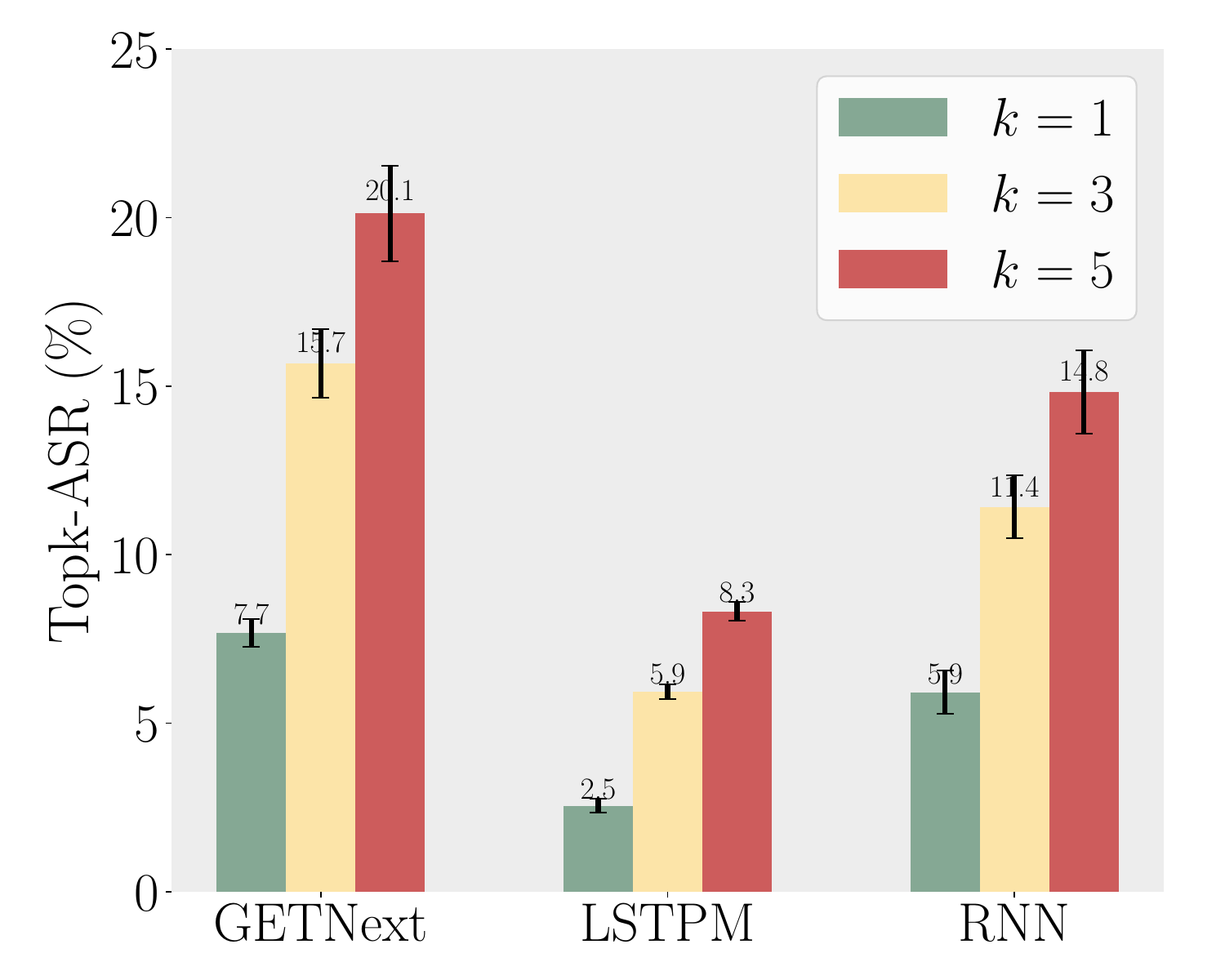}
    \label{fig:Attack2_results_4sq}}
    \subfigure [\textsc{TrajExtract} (\textsc{Gowalla})] {\includegraphics[width=0.22\textwidth]{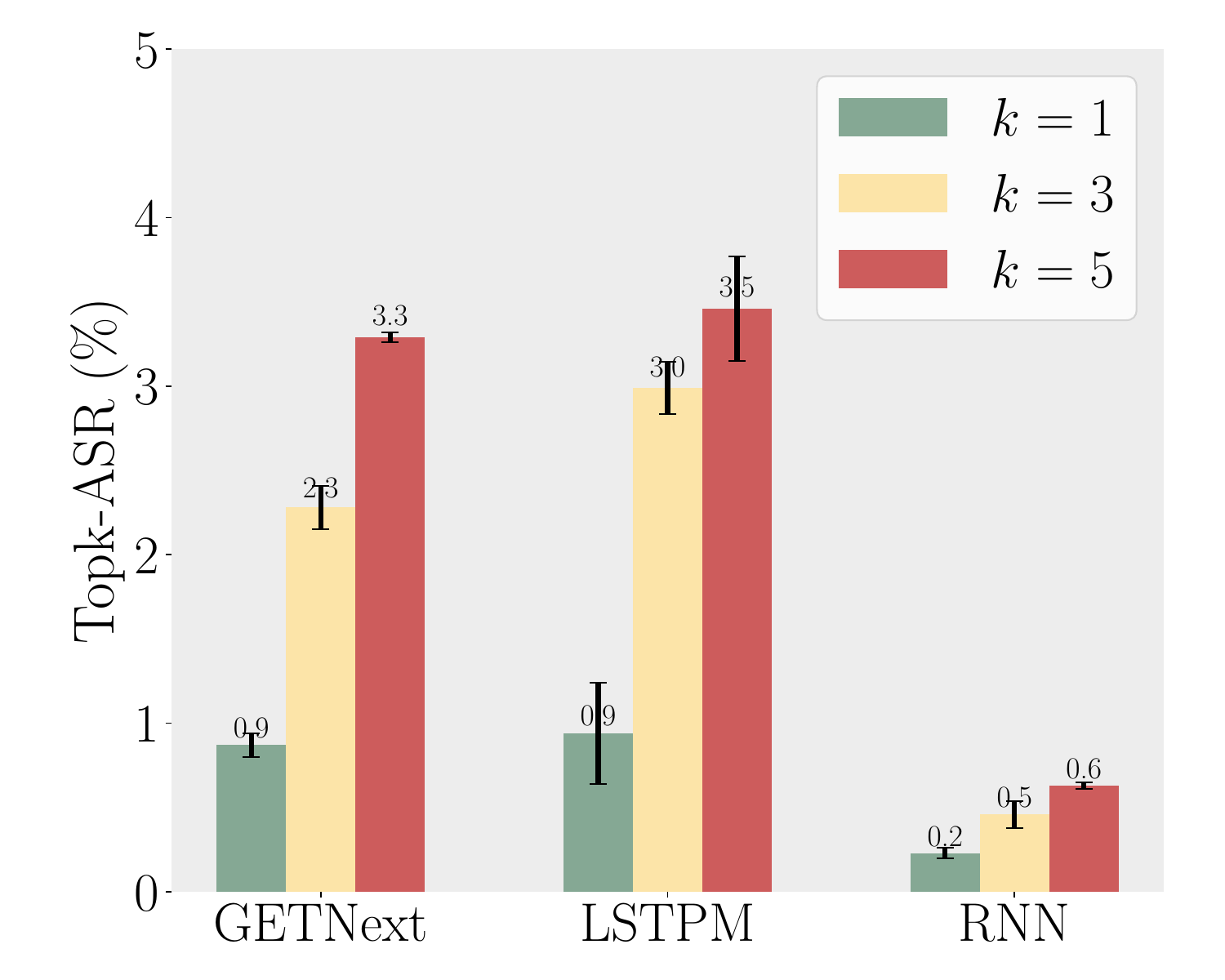}
    \label{fig:Attack2_results_gow}}
    \vspace{-4mm}
    \caption{Attack performance of data extraction attacks (\textsc{LocExtract} and \textsc{TrajExtract}) on three victim models and two mobility datasets.}
    \label{fig:extraction_results}
    \vspace{-3mm}
\end{figure*}

\begin{figure*}[t]
    \centering
    \subfigure [\textsc{LocMIA} (\textsc{4sq})] {\includegraphics[width=0.22\textwidth]{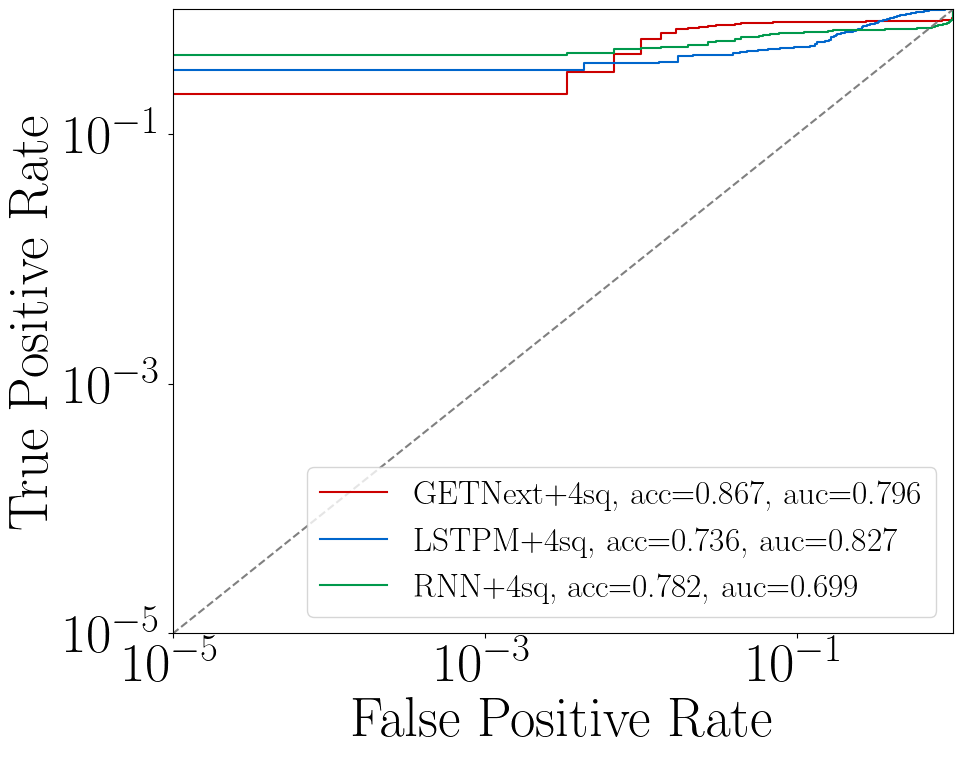}
    \label{fig:mia_results_3_4sq}}
    \subfigure [\textsc{LocMIA} (\textsc{Gowalla})] {\includegraphics[width=0.22\textwidth]{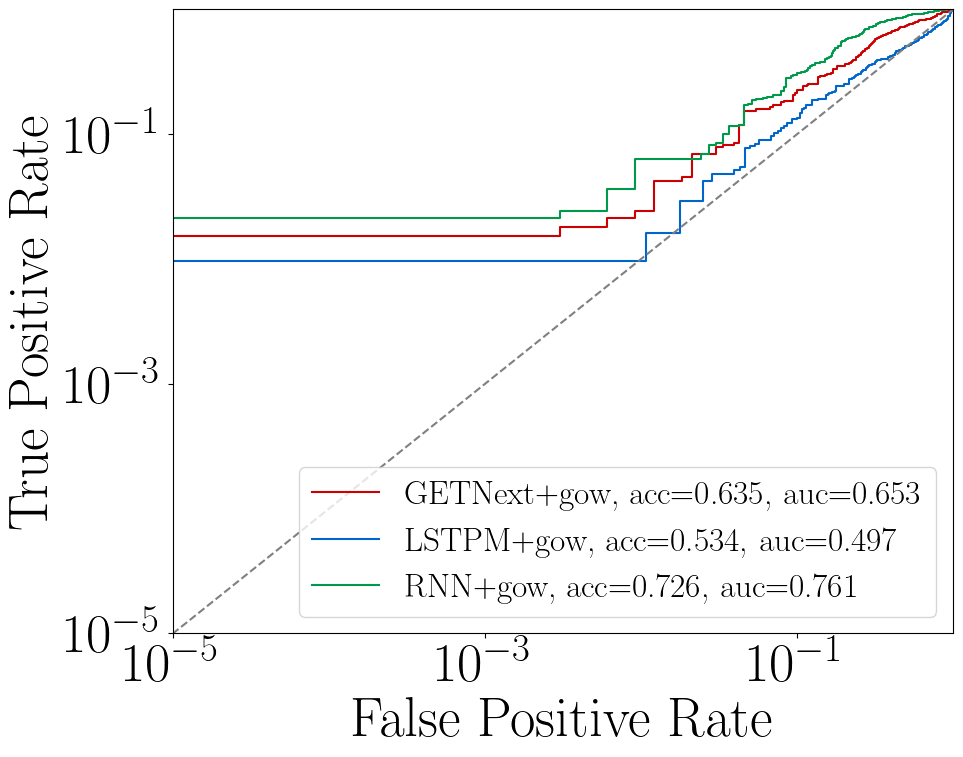}
    \label{fig:mia_results_3_gow}}
    \subfigure [\textsc{TrajMIA} (\textsc{4sq})] {\includegraphics[width=0.22\textwidth]{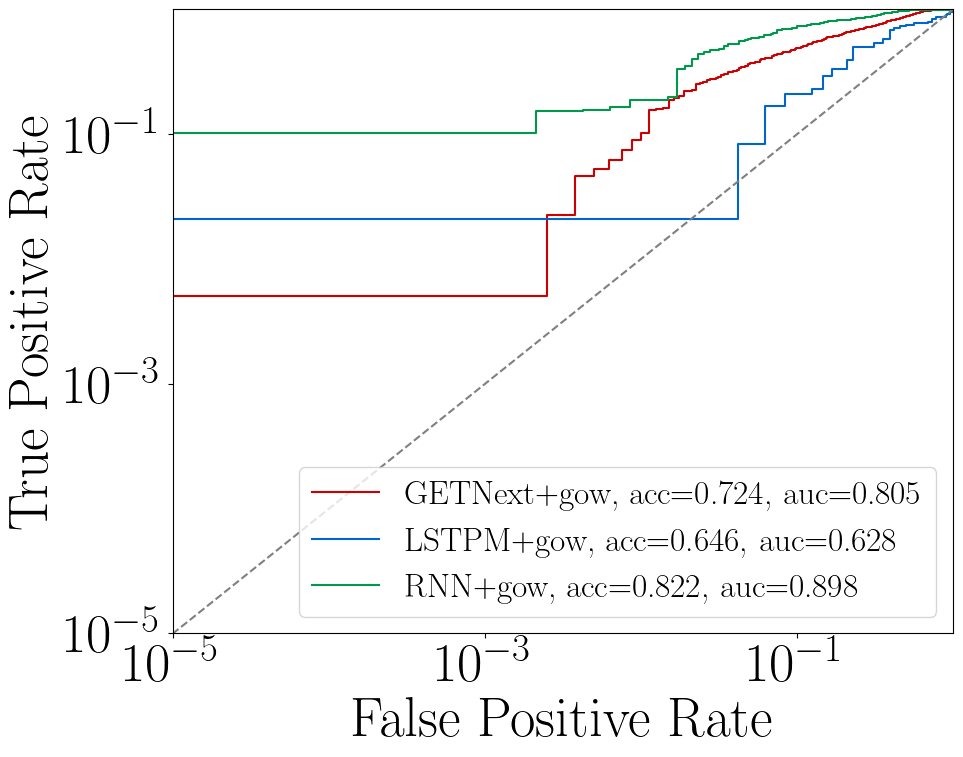}
    \label{fig:mia_results_4_4sq}}
    \subfigure [\textsc{TrajMIA} (\textsc{Gowalla})] {\includegraphics[width=0.22\textwidth]{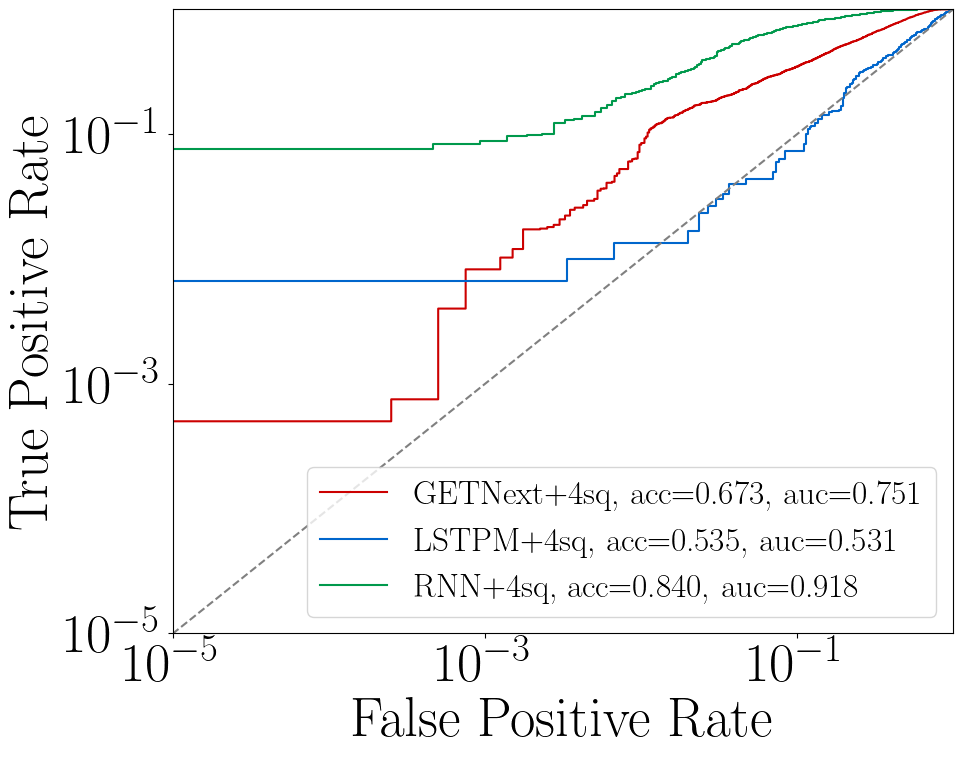}
    \label{fig:mia_results_4_gow}}
    \vspace{-4mm}
    \caption{
    Attack performance of (\textsc{LocMIA} and \textsc{TrajMIA}) on three victim models and two POI recommendation datasets. The diagonal line indicates the random guess baseline.
    } 
    \vspace{-3mm}
    \label{fig:mia_results}
\end{figure*}

Figures~\ref{fig:extraction_results} and~\ref{fig:mia_results} visualize the attack performance of data extraction and membership inference attacks, respectively. In Figure~\ref{fig:extraction_results}, we observe that \textsc{LocExtract} and \textsc{TrajExtract} can effectively extract users' most common locations and trajectories across various model architectures and datasets as the attack performance is significantly better than the random guess baseline, i.e., $1/|\mathcal{L}|$ ($0.04\%$ for \textsc{LocExtract}) and $1/|\mathcal{L}|^{n-1}$($10^{-8}\%$ for \textsc{TrajExtract}). Likewise, as shown in Figure~\ref{fig:mia_results}, \textsc{LocMIA} and \textsc{TrajMIA} successfully determine the membership of a specific user-location pair or trajectory, significantly outperforming the random guess baseline (represented by the diagonal line in both figures).

The attack performance also demonstrates that trajectory-level attacks are significantly more challenging than location-level attacks, evident from the better performance of \textsc{LocExtract} and \textsc{LocMIA} compared to \textsc{TrajExtract} and \textsc{TrajMIA} for data extraction and membership inference. We suspect this is because POI recommendation models are primarily designed to predict a single location. In contrast, our trajectory-level attacks aim to extract or infer a trajectory encompassing multiple consecutive locations. The experiment results also align with the findings that longer trajectories are less vulnerable to our attacks (see Figures~\ref{fig:trajlen-att2} and~\ref{fig:trajmia_querylen} in Appendix~\ref{result:parameter}).

The attack performance also differs across different model architectures and datasets. We see a general trend of privacy-utility trade-off in POI recommendation models based on the model performance of the victim model in Table~\ref{model_utility}: with better victim model performance comes better attack performance. While this is a common trend, it might not hold in some cases. For example, the \textsc{MIA} performance against \textsc{RNN} is sometimes better than \textsc{GETNext} and \textsc{LSTPM} performances. 
This might be because \textsc{GETNext} and \textsc{LSTPM} improve upon \textsc{RNN} by better leveraging spatial-temporal information 
in the mobility datasets. However, the adversary cannot use the exact spatial-temporal information in shadow model training since the adversary cannot access that information. 
This result can be inspiring in that even though spatial-temporal information can effectively improve attack performance, victim models that better utilize spatial-temporal information are still more resilient to MIAs. Future studies should also consider this characteristic when designing attacks or privacy-preserving POI recommendation models with better privacy-utility trade-offs.

\subsubsection{Factors in mobility data that make it vulnerable to the attacks (RQ2)}
\label{sec:dataspec}

\begin{figure}[h]
        \subfigure [\# of Users Visited]{\includegraphics[width=0.21\textwidth]{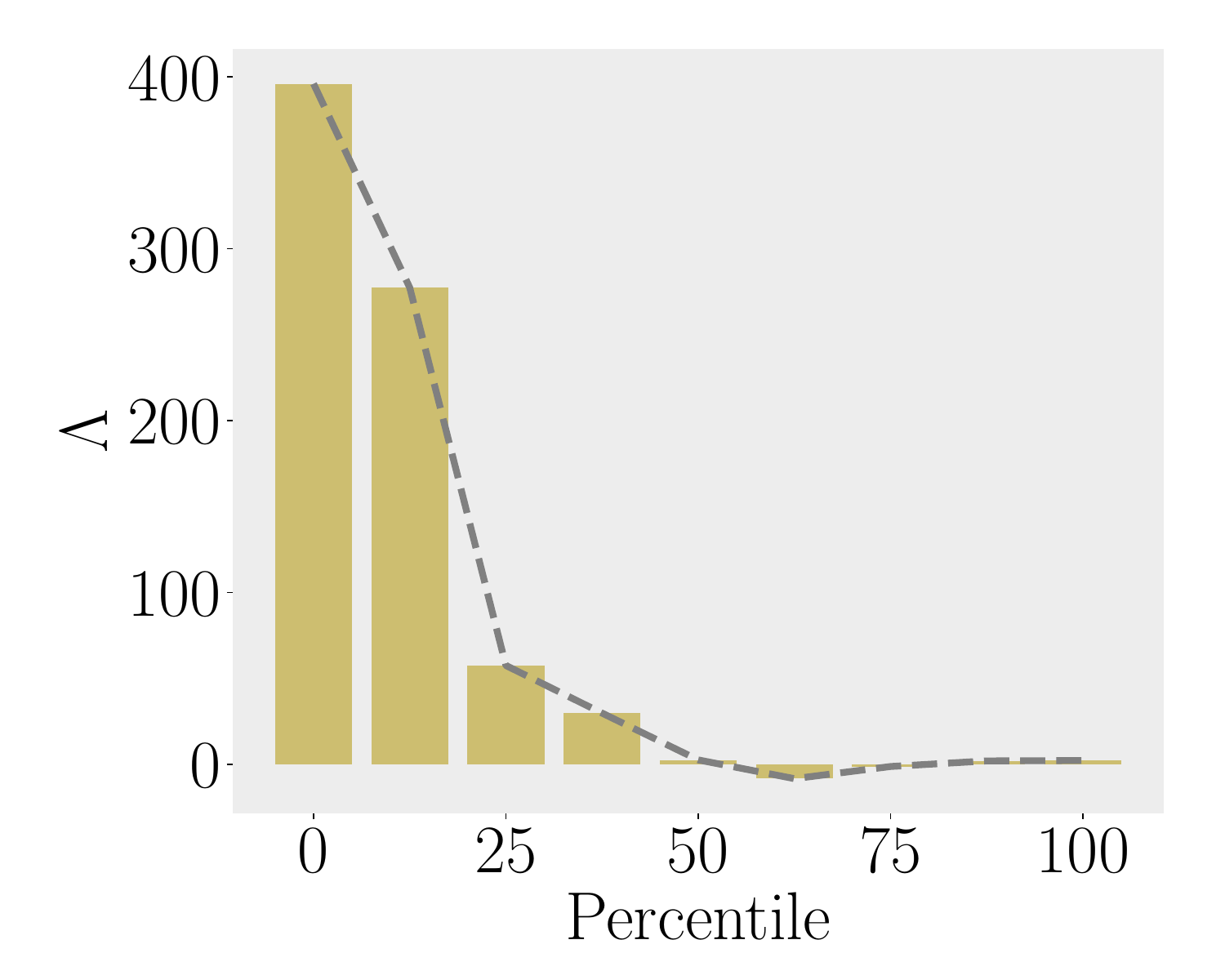}
        \label{fig:loc_num_user_4sq}}
        \subfigure [\# of Nearby Check-ins]{\centering \includegraphics[width=0.21\textwidth]
        {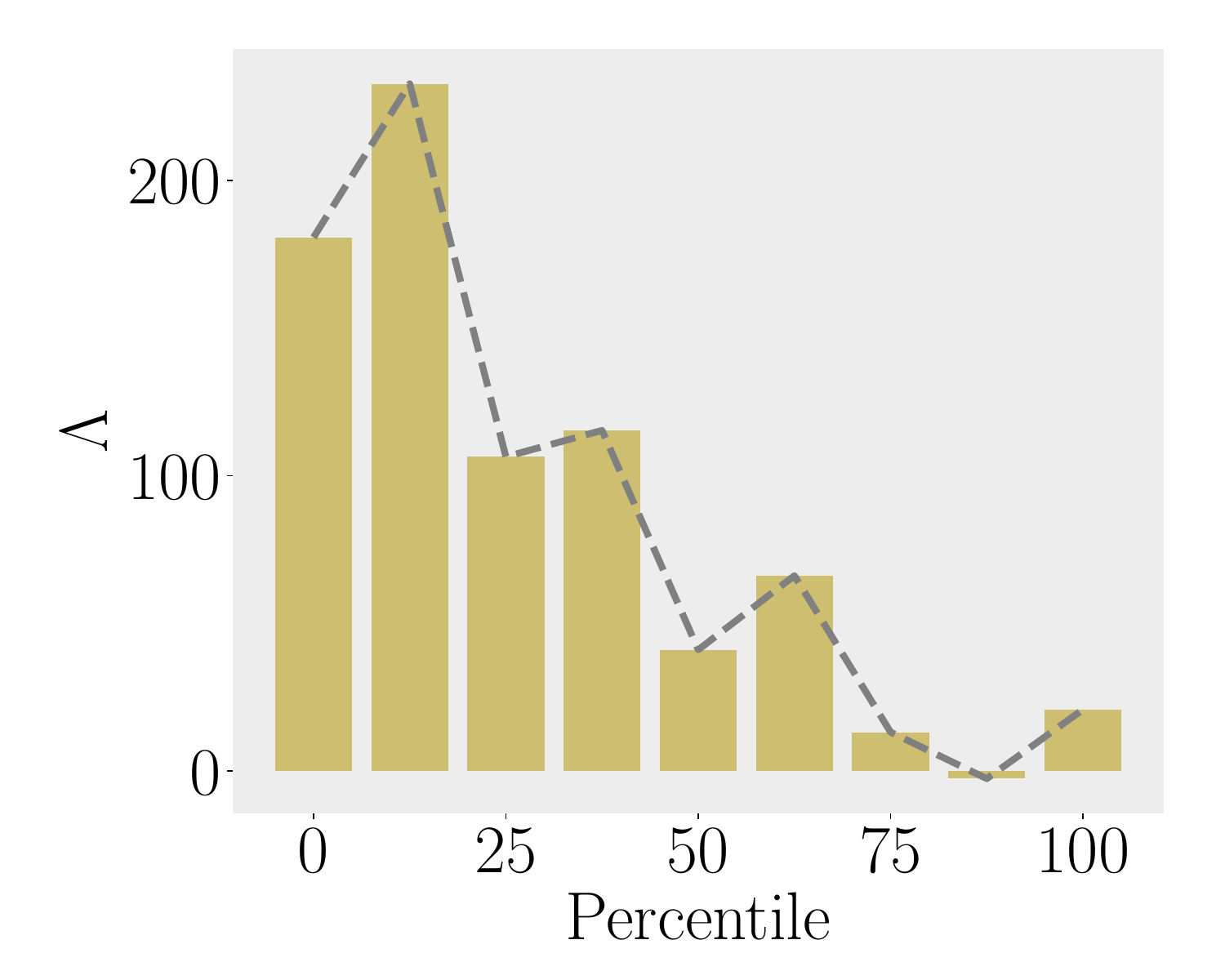}
        \label{fig:loc_num_loc_4sq}}
        \vspace{-4mm}
        \caption{How location-level aggregate statistics are related to \textsc{LocMIA}. Locations visited by fewer different users or have fewer surrounding check-ins are more vulnerable to \textsc{LocMIA}. 
        }
        \vspace{-2mm}
\end{figure}

\begin{figure*}[h!]
    \subfigure [\# of User Check-ins]{\includegraphics[width=0.22\textwidth]{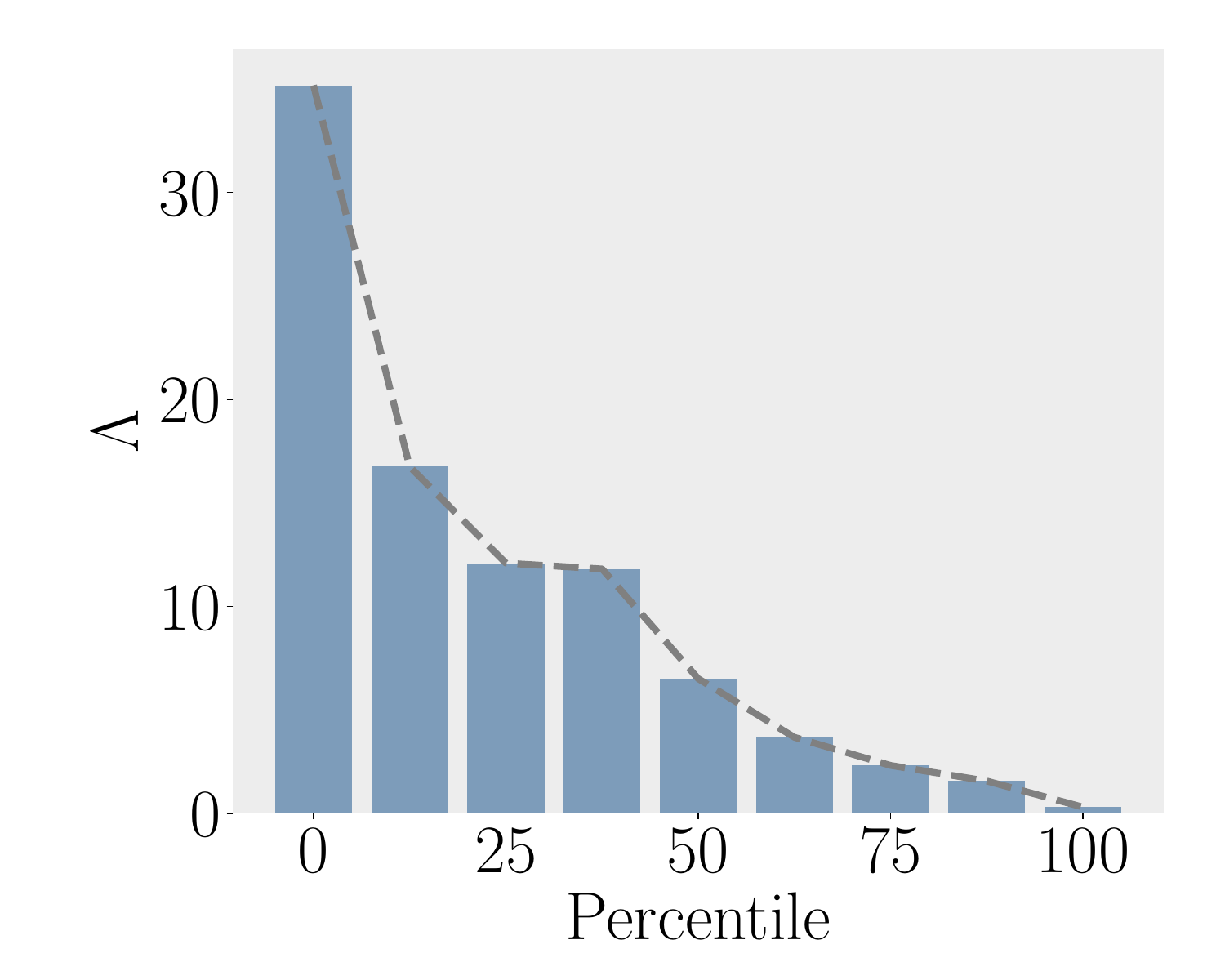}
    \label{fig:user_check_4sq}}
    \subfigure [\# of User POIs]{\includegraphics[width=0.22\textwidth]{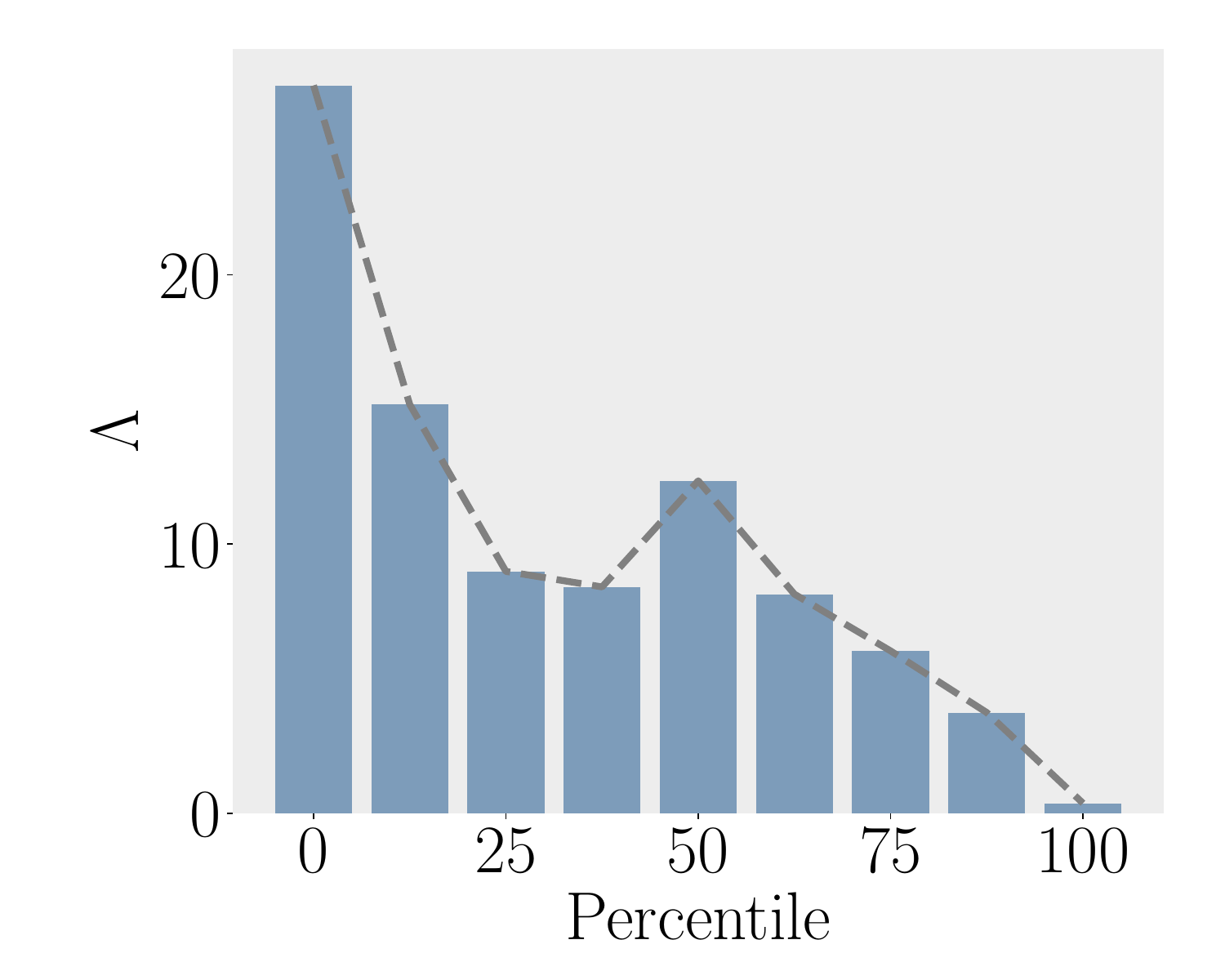}
     \label{fig:user_poi_4sq}}
    \subfigure [\# of User Trajectories]{\includegraphics[width=0.22\textwidth]{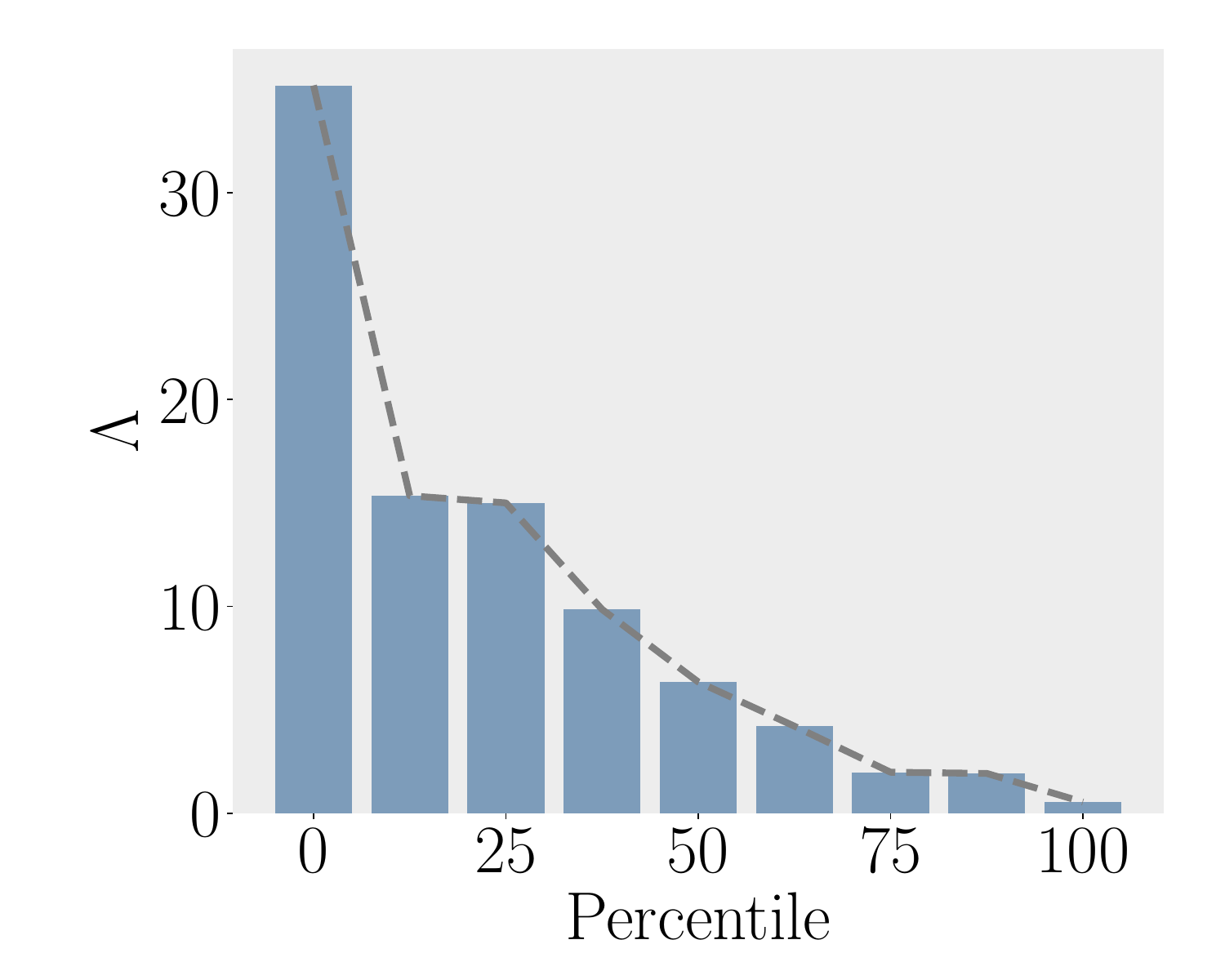}
    \label{fig:user_num_traj_4sq}}
    \subfigure [Avg User Traj Length]{\includegraphics[width=0.22\textwidth]{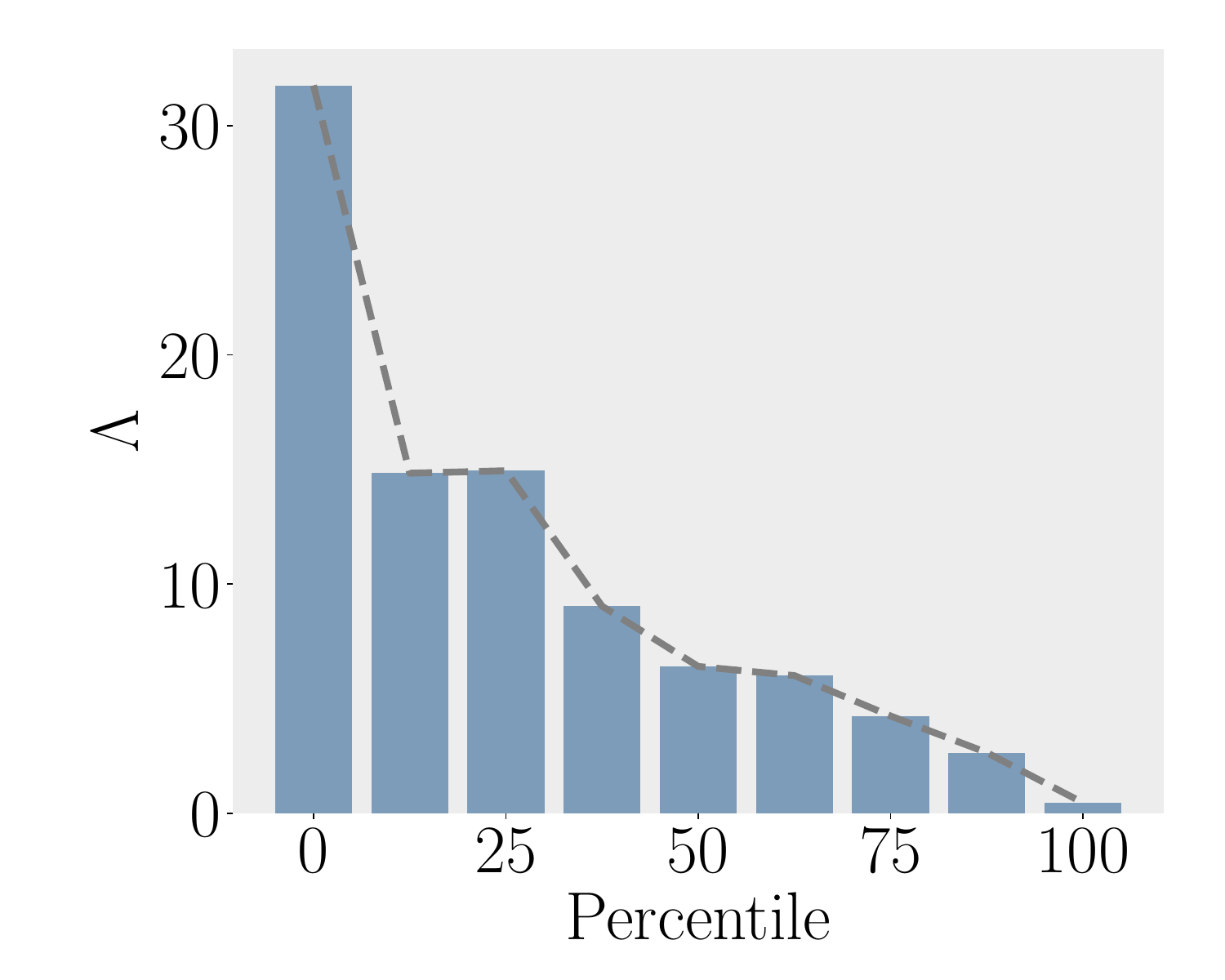}
    \label{fig:user_len_traj_4sq}}
    \vspace{-4mm}
    \caption{How user-level aggregate statistics are related to \textsc{TrajMIA}. \textit{x-axis:} Percentile categorizes users/locations/trajectories into different groups according to their feature values. \textit{y-axis:} $\Lambda$ indicates the (averaged) likelihood ratio of training trajectories/locations being the member over non-member from the hypothesis test for each group, with a higher value indicating the larger vulnerability. The users with fewer total check-ins, fewer unique POIs, and fewer or shorter trajectories are more vulnerable to \textsc{TrajMIA}. (\textsc{4sq})}
    \label{fig:user_poi_4sq_all}
    \vspace{-3mm}
\end{figure*}

\begin{figure}[h!]
        \subfigure [\# of Intercepting Trajs]{\includegraphics[width=0.21\textwidth]{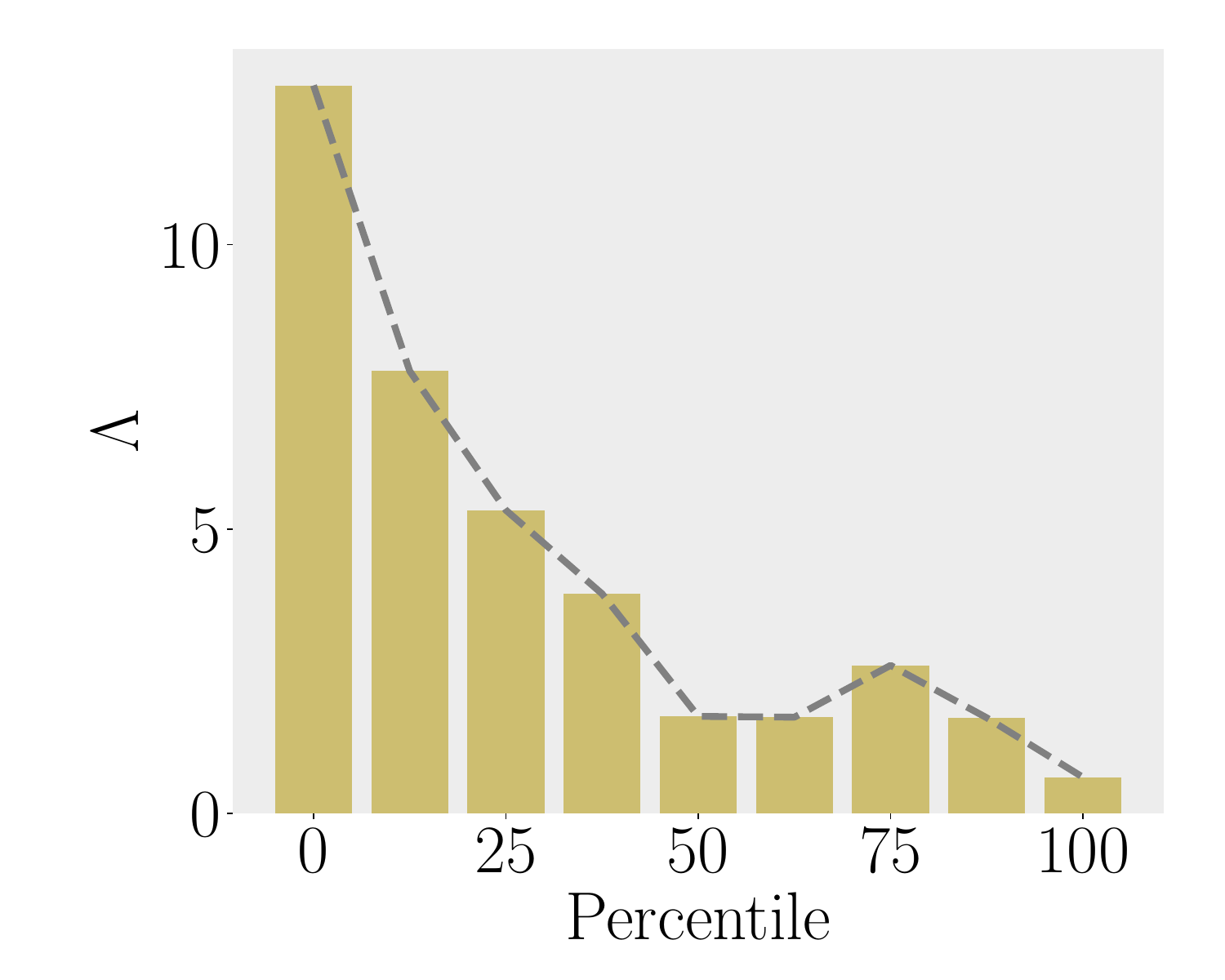}
        \label{ana:traj3}}
        \subfigure [\# of POIs in Traj]{\includegraphics[width=0.21\textwidth]{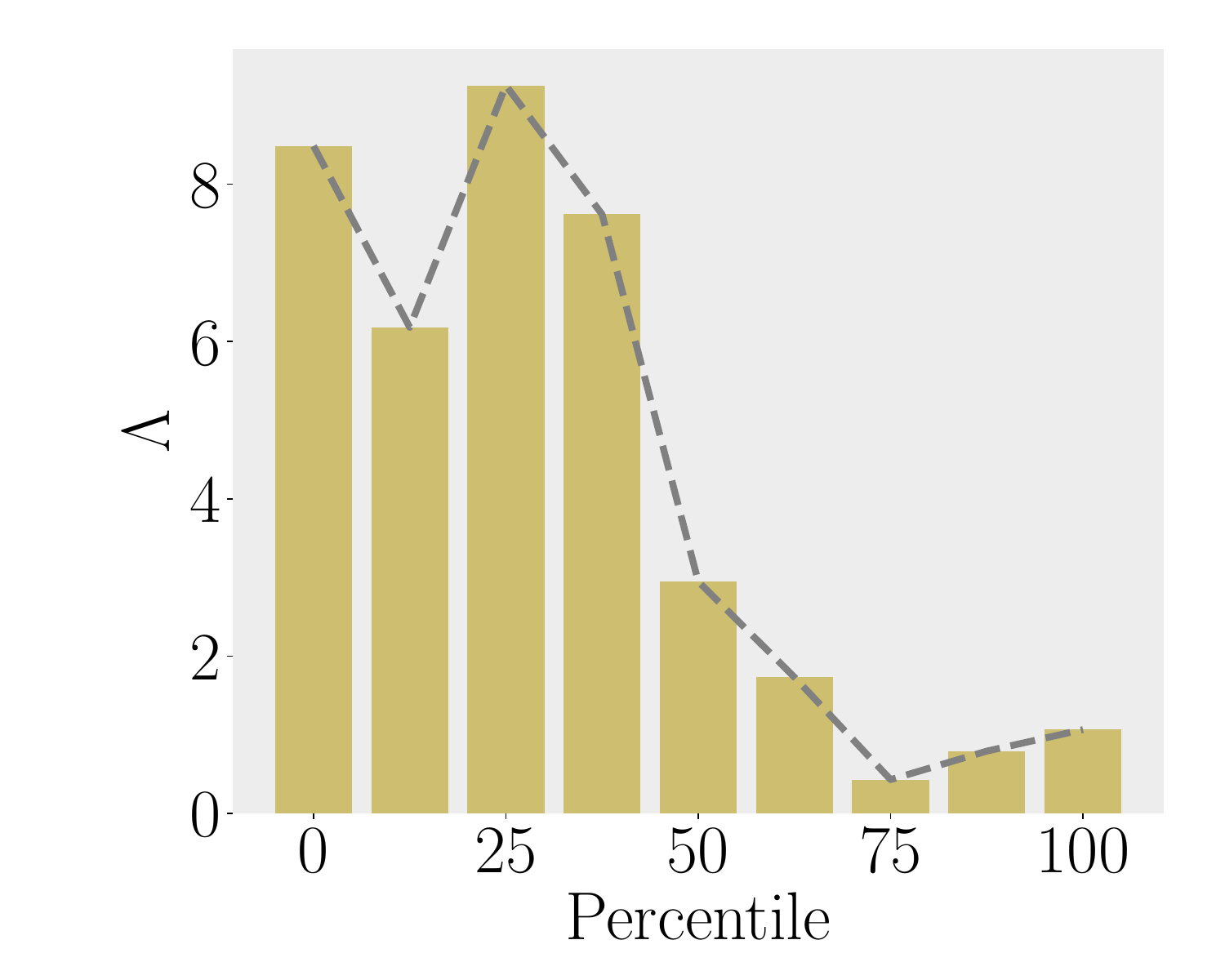}
        \label{ana:traj4}}
        \vspace{-4mm}
        \caption{How trajectory-level aggregate statistics are related to \textsc{TrajMIA}. The trajectories with fewer intercepting trajectories or fewer POIs are more vulnerable to \textsc{TrajMIA}.}
        \label{fig:locana}
        \vspace{-4mm}
\end{figure}

Prior research demonstrates that data outliers are the most vulnerable examples to privacy attacks~\citep{carlini2022membership, tramer2022truth} in image and text datasets. However, it is unclear whether the same conclusion holds in mobility data and what makes mobility data as data outliers. To this end, we investigate which factors of the mobility datasets influence the attack's efficacy. In particular, we collect aggregate statistics of mobility data from three perspectives: user, location, and trajectory. We analyze which factors in these three categories make mobility data vulnerable to our attacks. We defer the details of selecting the aggregate statistics and the list of selected aggregate statistics in our study in Appendix~\ref{ana:featureselect}. Our findings are as follows:

\begin{itemize}[leftmargin=*]
\item For \textsc{LocExtract}, we do not identify any meaningful pattern correlated with its attack performance. We speculate that a user's most common location is not directly related to the aggregate statistics we study. 

\item For \textsc{TrajExtract}, our findings indicate that \textit{users who have visited fewer unique POIs} are more vulnerable to this attack, as referenced in Figure~\ref{fig:user_att2} in Appendix~\ref{ana:user_ana}. 
This can be explained by the fact that when users have fewer POIs, the model is less uncertain in predicting the next location due to the reduced number of possible choices that the model memorizes.

\item For \textsc{LocMIA}, as shown in Figures~\ref{fig:loc_num_user_4sq} and~\ref{fig:loc_num_loc_4sq}, we find that \textit{locations visited by fewer users or have fewer surrounding check-ins} are more susceptible to \textsc{LocMIA}. 
We believe this is because those locations shared with fewer users or surrounding check-ins make them training data outliers.

\item For \textsc{TrajMIA}, \textit{users with fewer total check-ins} (Figure~\ref{fig:user_check_4sq}), \textit{unique POIs} (Figure~\ref{fig:user_poi_4sq}), and \textit{fewer or shorter trajectories} (Figures~\ref{fig:user_num_traj_4sq} and~\ref{fig:user_len_traj_4sq}) are more susceptible. In Figures~\ref{ana:traj3} and~\ref{ana:traj4}, we also see that \textit{trajectories intercepting less with others or with more check-ins} are more vulnerable to \textsc{TrajMIA}. We believe these user-level and trajectory-level aggregate statistics make the target examples data outliers.

\end{itemize}
In summary, we conclude that the effect of data outliers also exists in privacy attacks against POI recommendations. In the context of POI recommendation, the mobility data outliers could be characterized from the perspectives of user, location, and trajectory. Different attacks in our attack suite might be vulnerable to particular types of data outliers, which are more unique and are vulnerable against our attacks compared to other data.

\subsubsection{The impact of different attack designs (RQ3)}
\label{rq3}
\begin{figure}[h!]
    \centering
    \subfigure [Impact of $n_l$ on \textsc{LocMIA}] {\includegraphics[width=0.22\textwidth]{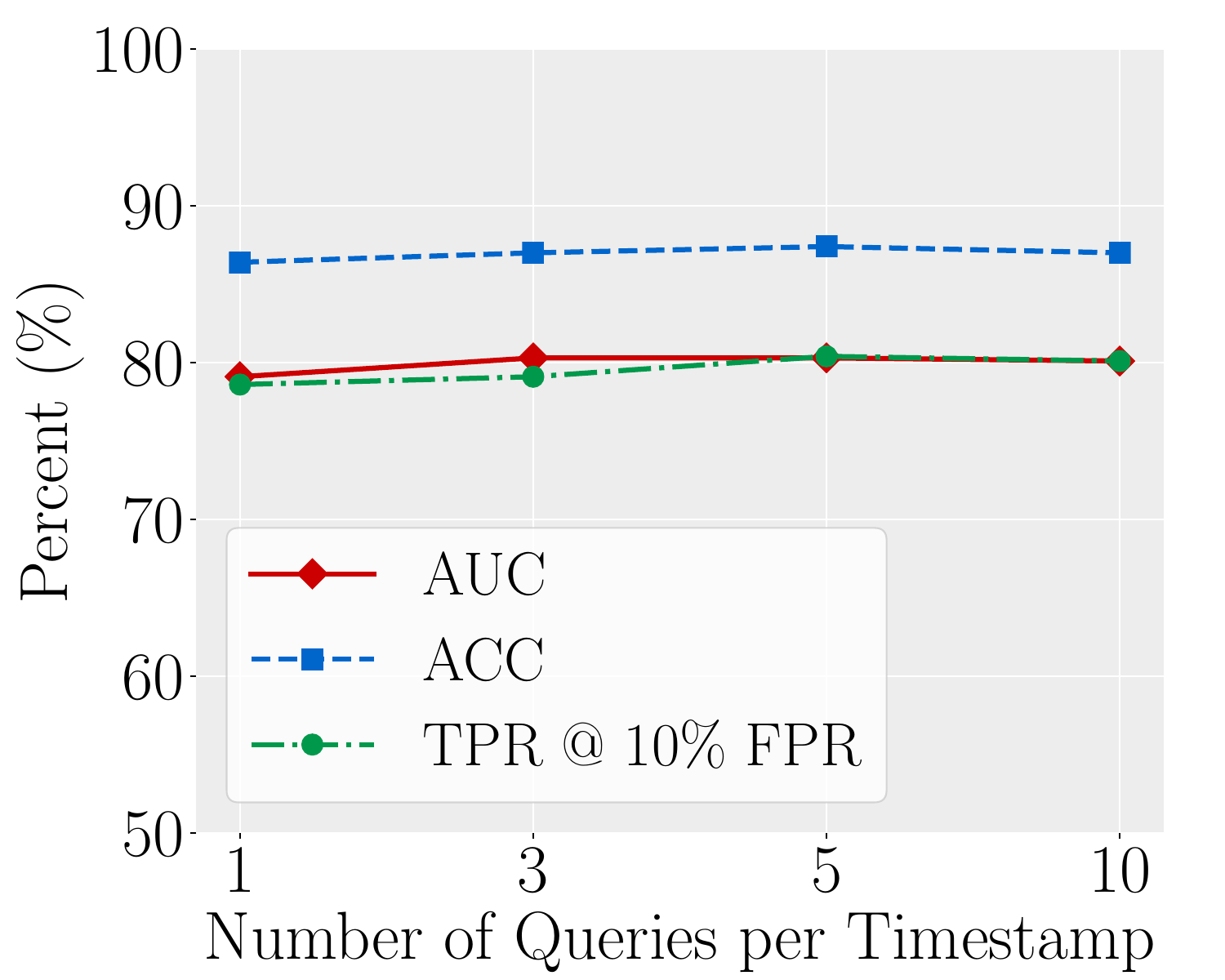}
    \label{fig:querylim_mia_loc}}
    \subfigure [Impact of $n_t$ on \textsc{LocMIA}] {\includegraphics[width=0.22\textwidth]{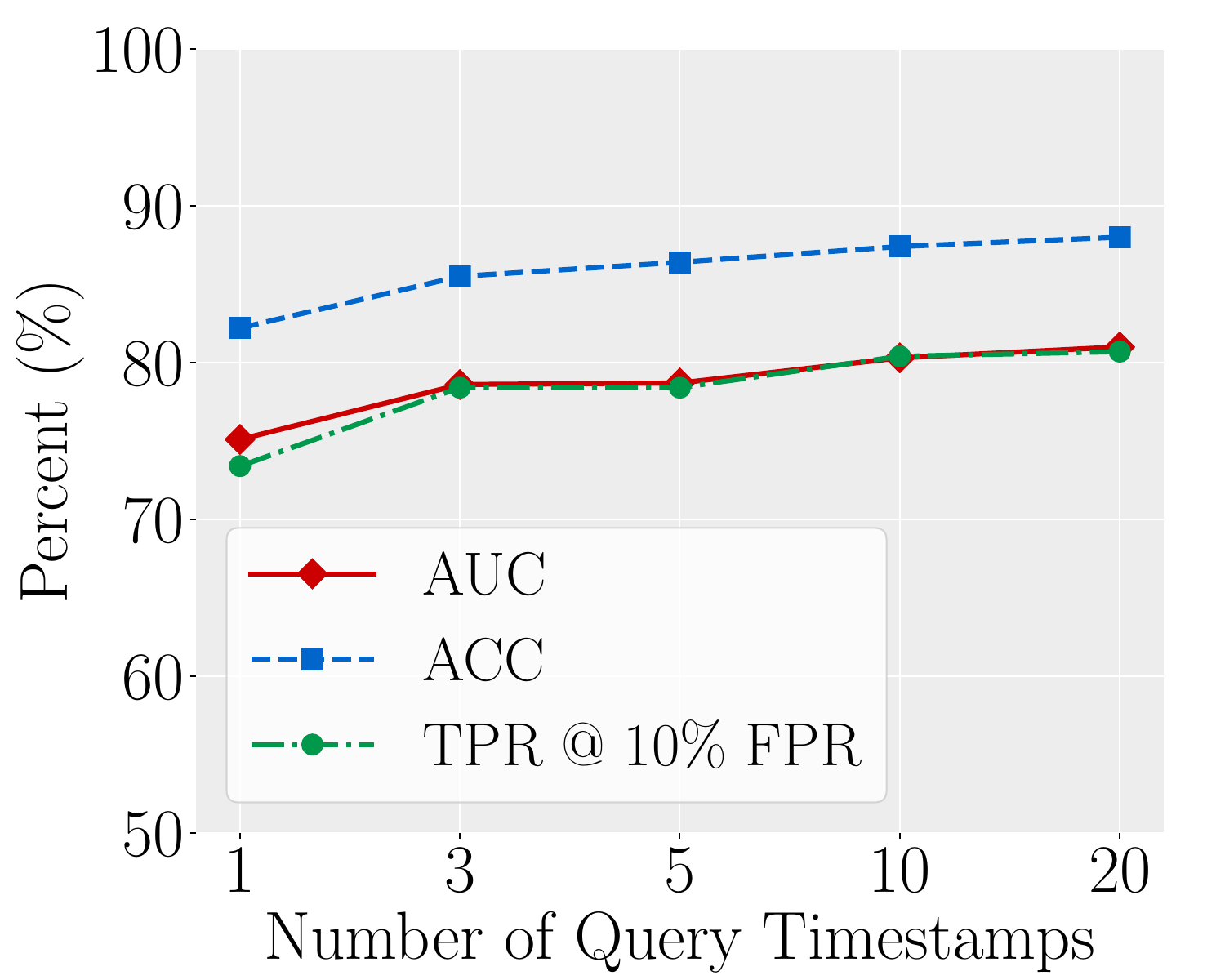}
    \label{fig:querylim_mia_time}}
    \vspace{-4mm}
    \caption{Both the number of query timestamps $n_t$ and the number of query locations $n_l$ affect the attack performance of \textsc{LocMIA}. We use the default $n_t=10$ and $n_l=10$ in (a) and (b), respectively. Given a few queries, \textsc{LocMIA} remains~effective.~(\textsc{4sq})}
    \label{fig:querylim_mia}
    \vspace{-5mm}
\end{figure}

We explore different attack designs that may affect the performance of each attack. In particular, we defer detailed results in Appendix~\ref{result:parameter} and summarize our key findings as follows:

\begin{itemize}[leftmargin=*]
\item For \textsc{LocExtract}, as shown in Figures~\ref{fig:dummytime}~and~\ref{fig:hard} in the Appendix, we find that employing \textit{appropriate query timestamp} and \textit{soft voting mechanism} leads to better attack performance. The reason is that the POI recommendation relies on temporal information to make more accurate predictions.
\item For \textsc{TrajExtract}, 
similar to \textsc{LocExtract}, we find that an appropriate query timestamp also promotes the attack performance. Moreover, Figure~\ref{fig:trajlen-att2} in the Appendix shows that our attack is more effective when extracting location sequences of \textit{shorter trajectories} as the influence of the starting location becomes weaker when the prediction moves forward.
\item For \textsc{LocMIA}, as shown in Figure~\ref{fig:querylim_mia}, \textit{utilizing more queries with different timestamps} in our spatial-temporal model query algorithm improves the results of inferring the membership of a target user-location pair $(u, l)$. Since the adversary lacks information about real input sequences that are followed by the target location $l$. Utilizing more queries helps to traverse the search space and promotes the attack performance.
\item For both \textsc{LocMIA} and \textsc{TrajMIA}, as shown in Figure~\ref{fig:shadownum} in the Appendix, we find that \textit{a larger number of shadow models} yields better attack performance since they provide more samples to simulate the loss distributions of in-samples and out-samples.
\end{itemize}
\noindent
Our analysis also reveals additional intriguing findings: (1) Temporal information in the attacker's queries can significantly affect the attack results. For extraction attacks, we find that the extraction ASRs first increase and then decrease as the query timestamp increases, which peaks at 0.5 (using the middle of the day as the query timestamp) according to Figure~\ref{fig:dummytime}. The reason is that most check-ins occur during the daytime rather than at night.

(2) Figure~\ref{fig:querylim_extraction} in the Appendix demonstrates that our data extraction attacks remain effective even with a limited number of queries. In other words, our attacks are efficient in real-world scenarios.

\begin{figure}[h!]
        \centering
        \subfigure [\textsc{LocExtract}] {\includegraphics[width=0.22\textwidth]{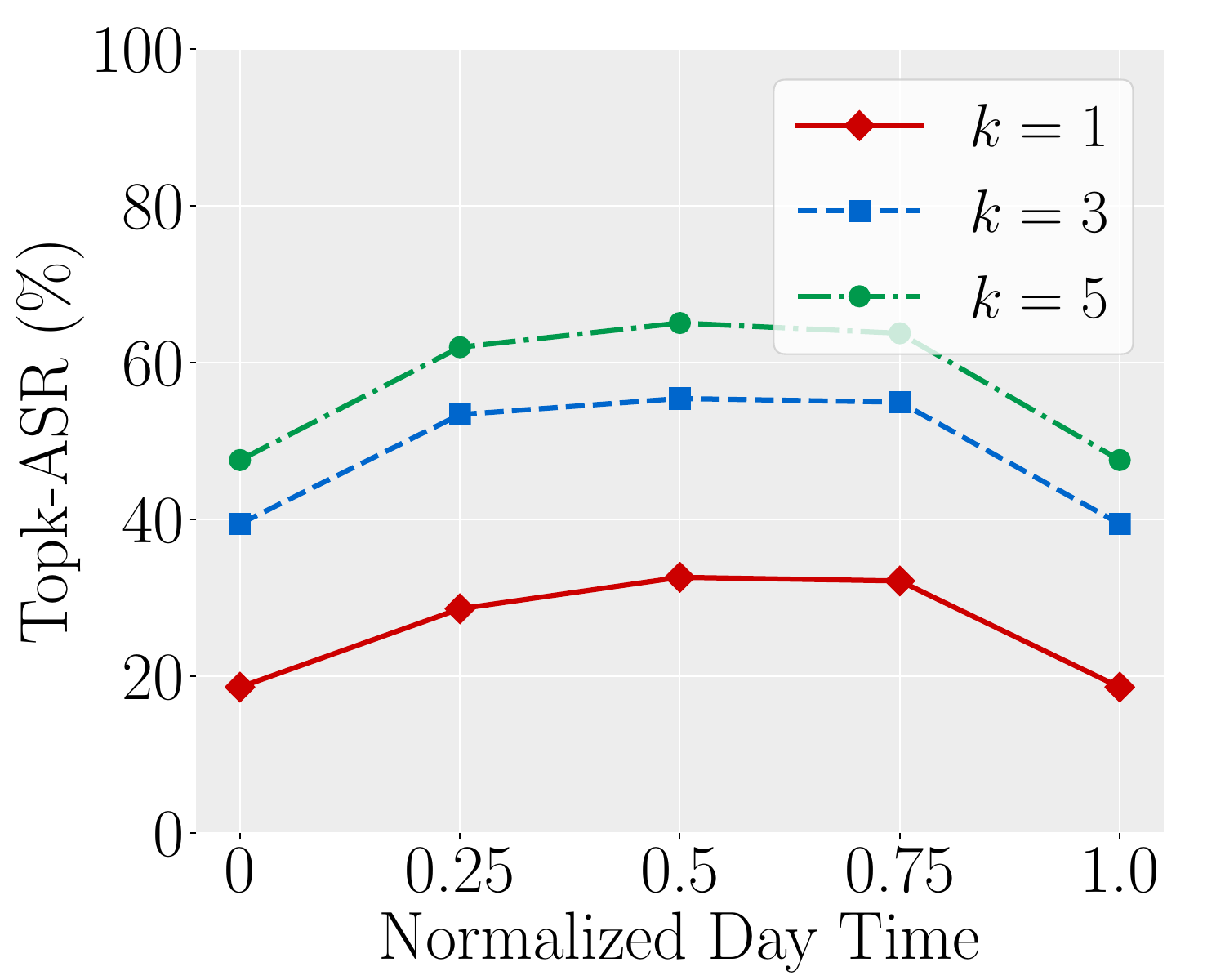}}
        \subfigure [\textsc{TrajExtract}] {\includegraphics[width=0.22\textwidth]{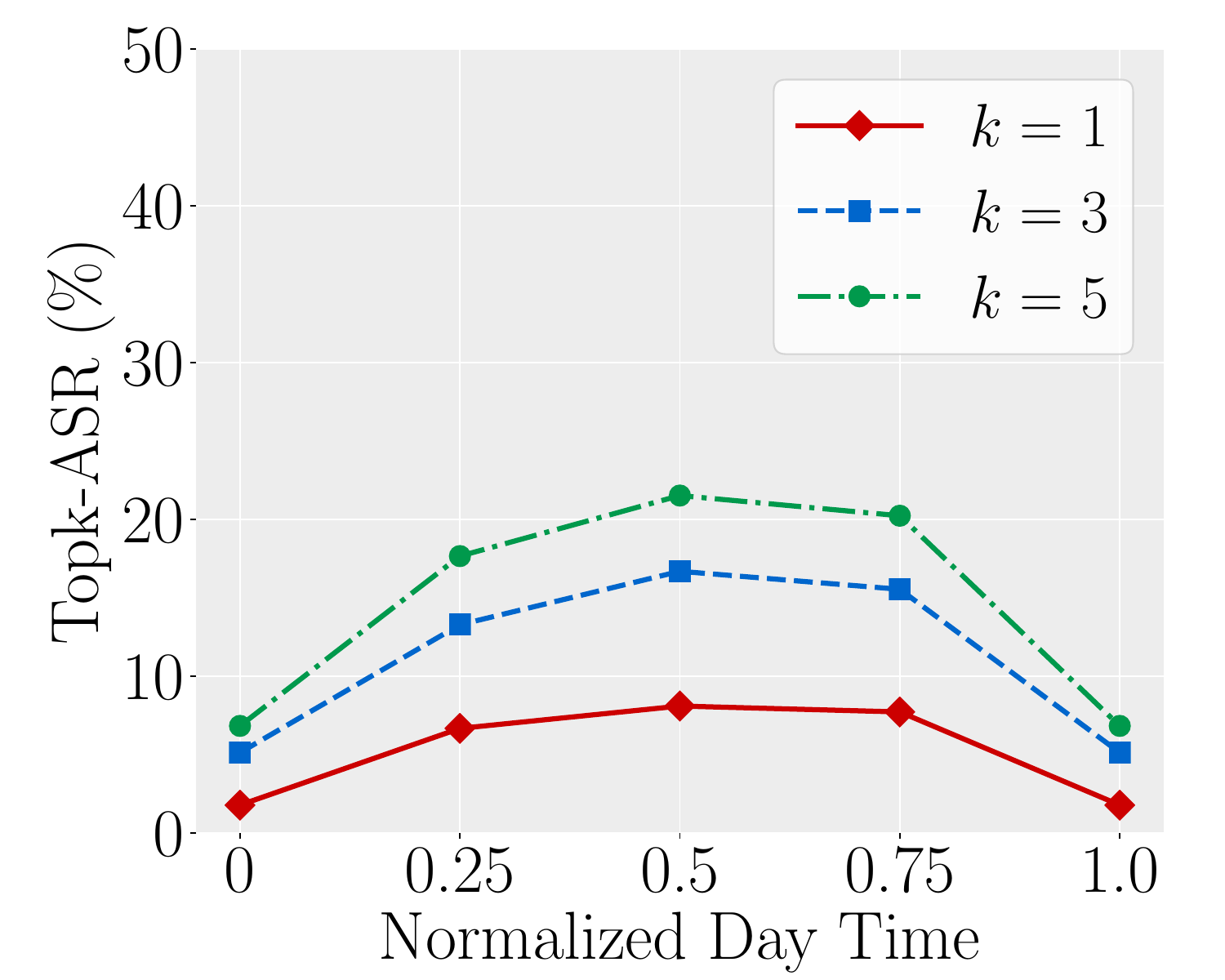}}
        \vspace{-4mm}
        \caption{The optimal query timestamp can significantly improve the performance of \textsc{LocExtract} and \textsc{TrajExtract}.(\textsc{4sq})}
        \label{fig:dummytime}
        \vspace{-3mm}
\end{figure}

\section{Defense}
\label{sec:defense}
We evaluate existing defenses against privacy attacks on machine learning models. Due to the limited space, we illustrate the defense mechanisms and the key findings in the main paper and defer experimental details to Appendix~\ref{sec:defenses}. 
\subsection{Defense Techniques}
In particular, we evaluate two streams of defense mechanisms on proposed attacks, including (1) standard techniques to reduce overfitting (e.g., early stopping, $l_2$ regularization) and (2) differential privacy-based defenses (e.g., DP-SGD~\citep{ACGMM16}) for provable risk mitigation. The standard techniques reduce the victim model's memorization to some degree, but they are insufficient due to the lack of statistical guarantees. To fill this gap, differential privacy~\citep{DWO14} is also used to defend against our attacks, which can theoretically limit the impact of a single data point on the model's performance.

Specifically, we first experiment with DP-SGD~\citep{ACGMM16}, the most representative DP-based defense, to train differentially-private POI recommendation models. The key idea of DP-SGD is to add Gaussian noises $\mathcal{N}(0, \sigma^2 C^2 I)$ to the clipped gradients $g$ of the model during its training process. $C$ is a clipping threshold that bounds the sensitivity of $g$ by ensuring $\lVert g \rVert \leq C$. To achieve $(\epsilon,\delta)$-DP, we have $\sigma =  \sqrt{2 \ln \frac{1.25}{\delta}}/\epsilon$. Despite that DP-SGD provides promising defense performance on language tasks~\cite{fernandes2019generalised}, we find that it can substantially sacrifice the model's utility on the POI recommendation task. Specifically, the top-10 accuracy is only 4.97\% when the mechanism satisfies $(5,0.001)$-DP, while the original top-10 accuracy without DP is 71\%. The reason for this performance decrease is that POI recommendation aims to make accurate user-level predictions within a large output space (i.e., $>4,000$ possible POIs). For different users, even the same location sequence may lead to a different result, which means that the model needs to capture user-specific behavior patterns from a relatively small user dataset. As a result, the training is quite sensitive to the noises introduced by DP-SGD, making it not applicable to POI recommendations.

However, we argue that DP-SGD provides undifferentiated protection for all the mobility data, while for POI recommendation, protecting more tailored sensitive information is more important. For example, a defender may only care about whether a list of check-ins about home addresses is protected or not. To this end, we introduce the notion of selective DP~\citep{SCLJY22} to relax DP and improve the model's utility-privacy trade-offs. Specifically, we apply the state-of-the-art selective DP method JFT~\citep{shi2022just} to protect different levels of sensitive information for each attack. 
The key idea of JFT is to adopt a two-phase training process: 
in the phase-I training, JFT redacts the sensitive information in the training dataset and optimizes the model with a standard optimizer; in the phase-II training, JFT applies DP-SGD to finetune the model on the original dataset in a privacy-preserving manner.
Due to the phase-I training, we observe that the model's utility is significantly promoted. In addition to JFT, we also apply Geo-Indistinguishability (Geo-Ind)~\citep{andres2013geo} to protect common locations in \textsc{LocExtract}. We note that Geo-Ind is only applicable to \textsc{LocExtract} (but not \textsc{LocMIA}) because it requires modifying the training data and is incompatible with the notion of membership inference.

\subsection{Takeaway Messages from the Defense}

We evaluate different defense mechanisms in terms of their performance in preventing each attack from stealing the corresponding sensitive information. Table~\ref{sensitive_info} summarizes the exposure of sensitive information in each attack. Besides, Appendix~\ref{sec:defenses} illustrates our evaluation metrics, experimental setup, and the results. Recall that the check-ins within a mobility dataset are not equally important. Therefore, in our experiments, we comprehensively evaluate the defense mechanisms from two perspectives. Specifically, we measure their performance in \emph{(1)} protecting all the sensitive information and \emph{(2)} protecting a targeted subset of sensitive information from being attacked.
Tables~\ref{defense}~and~\ref{defense2} in the Appendix show that existing defenses provide a certain degree of guarantee in mitigating privacy risks of ML-based POI recommendations, especially for the targeted subset of sensitive information. However, there is no such unified defense that can successfully defend against all the proposed attacks within a small utility drop. In other words, our exploration highlights the need for more advanced defenses.

{\tiny
\begin{table}[t!]\renewcommand{\arraystretch}{1.2}
\fontsize{8.5}{8.5}\selectfont
\centering
\caption{The exposure of sensitive information in each attack.}
\vspace{-2mm}
\begin{tabular}{c|c}
\toprule
Attack & Sensitive Information  \\ \midrule
\textsc{LocExtract} & Most common location of each user \\ 
\textsc{TrajExtract} & Each location sequence/sub-sequence $(x_L)$ \\
\textsc{LocMIA} & Each user-location pair $(u,l)$ \\ 
\textsc{TrajMIA} & Each trajectory sequence/sub-sequence $(x_T)$ \\ 
\bottomrule
\end{tabular}
\vspace{-3mm}

\label{sensitive_info}
\end{table}
}

\section{Discussion}
\subsection{Broader Impacts}
\label{broader}

\begin{itemize}[leftmargin=*]
\item Our attack suite unveils the vulnerability of POI recommendation models to privacy attacks. Our attack mechanisms (e.g., the Spatial-Temporal Query, which better utilizes spatial-temporal information to boost attack performance) and findings (e.g., identifying the types of data more likely to be memorized by the model) also benefit the measurement of privacy leakage for a broader range of learning-based models that involve spatio-temporal data, e.g., pre-trained trajectory models~\cite{poi_embedding_privacy_1} and spatio-temporal LLMs~\cite{llm_traj_liangzhao}.

\item The defense results indicate that selectively protecting sensitive information and using approaches like fine-tuning pre-trained models on a smaller private dataset might improve the utility-privacy trade-off. In other words, our study reveals that selective defense serves as a promising direction to reduce the privacy risks of these models.
\end{itemize}

\subsection{Related Work}

Mobility data contain rich information that can reveal individual privacy such as user identity. Previous work utilizes side-channel attacks to extract sensitive information about mobility data from LBS, including social relationships~\citep{poi_social_communication, powar2023sok}, aggregated trajectories~\citep{pyrgelis2017knock, pyrgelis2020measuring, zhang2020locmia}, trajectory history~\citep{poi_linkage_percom, poi_linkage_percom2}, network packets~\citep{vratonjic2014location,jiang2007preserving} and location embeddings~\cite{poi_embedding_privacy_1}. Despite the focus of previous work, deep neural networks (DNN) built on large volumes of mobility data have recently become state-of-the-art backbones for LBS, opening a new surface for privacy attacks. 
To the best of our knowledge, our work is the first of its kind to investigate the vulnerabilities of DNN models in leaking sensitive information about mobility data using inference attacks.

\noindent
\textbf{Privacy Attacks.}~~
Various types of privacy attacks, such as membership inference 
attacks~\citep{shokri2017membership, salem2018ml, carlini2022membership, gomrokchi2023membership}, training data extraction attacks~\citep{carlini2019secret, carlini2023extracting}, and model inversion attacks~\citep{fredrikson2015model} have been proposed to infer sensitive information from model training data. Our attack suite contains membership inference and data extraction attacks. Existing data extraction and membership inference attacks~\citep{carlini2019secret, carlini2022membership} are insufficient for POI recommendation models due to the spatio-temporal nature of the data. Our work takes the first step to extracting sensitive location and trajectory patterns from POI recommendation models and solving unique challenges to infer the membership of both user-location pairs and user trajectories.
As a final remark, our attacks differ from previous MIAs in mobility data \citep{pyrgelis2017knock,zhang2020locmia}, which focus on the privacy risks of data aggregation.

\noindent
\textbf{Mobility Data Privacy}\label{rel:mobility}~~
Mobility data contain rich information that can reveal individual privacy such as user identity. Previous work utilizes side-channel attacks to extract sensitive information about mobility data from LBS, including social relationships~\citep{poi_social_communication, poi_social_internet_computing, poi_social_mm}, aggregated trajectories~\citep{pyrgelis2017knock, pyrgelis2020measuring, zhang2020locmia}, trajectory history~\citep{poi_linkage_percom, poi_linkage_percom2, poi_linkage_percom3}, network packets~\citep{vratonjic2014location,jiang2007preserving} and location embeddings~\cite{poi_embedding_privacy_1}. Despite the focus of previous work, deep neural networks (DNN) built on large volumes of mobility data have recently become state-of-the-art backbones for LBS, opening a new surface for privacy attacks. 
To the best of our knowledge, our work is the first of its kind to investigate the vulnerabilities of DNN models in leaking sensitive information about mobility data using inference attacks.

\noindent
\textbf{More Related Works}~~Due to the limited space, we have deferred more related work of defense %
to Appendix section~\ref{morerelated}.

\section{Conclusion} 
In this work, we take the first step to evaluate the privacy risks of the POI recommendation models. In particular, we introduce an attack suite containing data extraction attacks and membership inference attacks to extract and infer sensitive information about location and trajectory in mobility data. We conduct extensive experiments to demonstrate the effectiveness of our attacks. 
Additionally, we analyze what types of mobility data are vulnerable to the proposed attacks. 
To mitigate our attacks, we further adapt two mainstream defense mechanisms to the task of POI recommendation. Our results show that there is no single solid defense that can simultaneously defend against proposed attacks. 
Our findings underscore the urgent need for better privacy-preserving approaches for POI recommendation models. Interesting future works include: (1) Generalize the attack suite to measure privacy risks of real-world location-based services (e.g., Google Maps) in a more challenging setting (e.g., label-only setting). (2) Develop more advanced defense mechanisms against our attacks.

\noindent
\textbf{Ethics Statement.}~~
This work introduces a novel attack suite on POI recommendation models trained on public anonymized mobility datasets with no personally identifiable information, aimed at bringing potential vulnerabilities in POI models to public attention. The success of our attacks offers insights into future privacy leakage measurement in learning-based models involving spatio-temporal data. Moreover, it underscores the need for improved defense solutions for POI models with better utility-privacy trade-offs.
We hope our study fosters further research in protecting the privacy of mobility data.

\section{Author Contributions}
\begin{itemize}[leftmargin=*]
    \item Jianfeng, Yuan, and Kunlin decided on the goal and research question of the paper.
    \item Jianfeng, Yuan, Kunlin, and Jinghuai defined the threat model.
    \item Will, Yuan, Zhiqing, Guang, and Desheng advised on the formalization of the threat model.
    \item Kunlin and Jianfeng studied and collected literature related to attacks.
    \item Jianfeng, Jinghuai, and Will studied and collected literature related to defense mechanisms.
    \item Kunlin and Zhiqing studied and collected literature related to POI recommendations.
    \item Kunlin, Jianfeng, Jinghuai, and Yuan designed the attack mechanism.
    \item Zhiqing, Guang, and Desheng guided data and model selection.
    \item Kunlin and Jinghuai developed and executed the attacks.
    \item Kunlin, Jianfeng, and Jinghuai devised the analysis to evaluate attack performance.
    \item Kunlin and Jinghuai conducted the analysis experiments and produced result visualizations.
    \item Jianfeng and Jinghuai proposed defense strategies.
    \item Jinghuai implemented the defense mechanisms.
    \item Kunlin, Jianfeng, and Jinghuai wrote and edited the paper.
    \item Jianfeng, Yuan, Desheng, and Guang provided suggestions for paper writing.
\end{itemize}

\begin{acks}
We sincerely thank the reviewers for their valuable feedback on the paper.
This work is supported in part by the National Science
Foundation (NSF) Awards 1951890, 1952096, 2003874, 2047822, 2317184, 2325369, 2411151, 2411152, 2411153,  UCLA ITLP and Okawa foundation.
Any opinions, findings, conclusions, or recommendations expressed in this publication are those of the authors and do not necessarily reflect the views of sponsors.
\end{acks}

\bibliographystyle{ACM-Reference-Format}
\bibliography{new}


\begin{thebibliography}{77}


\ifx \showCODEN    \undefined \def \showCODEN     #1{\unskip}     \fi
\ifx \showDOI      \undefined \def \showDOI       #1{#1}\fi
\ifx \showISBNx    \undefined \def \showISBNx     #1{\unskip}     \fi
\ifx \showISBNxiii \undefined \def \showISBNxiii  #1{\unskip}     \fi
\ifx \showISSN     \undefined \def \showISSN      #1{\unskip}     \fi
\ifx \showLCCN     \undefined \def \showLCCN      #1{\unskip}     \fi
\ifx \shownote     \undefined \def \shownote      #1{#1}          \fi
\ifx \showarticletitle \undefined \def \showarticletitle #1{#1}   \fi
\ifx \showURL      \undefined \def \showURL       {\relax}        \fi
\providecommand\bibfield[2]{#2}
\providecommand\bibinfo[2]{#2}
\providecommand\natexlab[1]{#1}
\providecommand\showeprint[2][]{arXiv:#2}

\bibitem[\protect\citeauthoryear{Abadi, Chu, Goodfellow, McMahan, Mironov, Talwar, and Zhang}{Abadi et~al\mbox{.}}{2016}]%
        {ACGMM16}
\bibfield{author}{\bibinfo{person}{Martin Abadi}, \bibinfo{person}{Andy Chu}, \bibinfo{person}{Ian Goodfellow}, \bibinfo{person}{H~Brendan McMahan}, \bibinfo{person}{Ilya Mironov}, \bibinfo{person}{Kunal Talwar}, {and} \bibinfo{person}{Li Zhang}.} \bibinfo{year}{2016}\natexlab{}.
\newblock \showarticletitle{Deep learning with differential privacy}. \bibinfo{publisher}{ACM SIGSAC Conference on Computer and Communications Security}.
\newblock


\bibitem[\protect\citeauthoryear{Andr{\'e}s, Bordenabe, Chatzikokolakis, and Palamidessi}{Andr{\'e}s et~al\mbox{.}}{2013}]%
        {andres2013geo}
\bibfield{author}{\bibinfo{person}{Miguel~E Andr{\'e}s}, \bibinfo{person}{Nicol{\'a}s~E Bordenabe}, \bibinfo{person}{Konstantinos Chatzikokolakis}, {and} \bibinfo{person}{Catuscia Palamidessi}.} \bibinfo{year}{2013}\natexlab{}.
\newblock \showarticletitle{Geo-indistinguishability: Differential privacy for location-based systems}. \bibinfo{publisher}{ACM SIGSAC Conference on Computer and Communications Security}.
\newblock


\bibitem[\protect\citeauthoryear{Bai, Yao, Li, Wang, and Wang}{Bai et~al\mbox{.}}{2020}]%
        {traffic_nips2020}
\bibfield{author}{\bibinfo{person}{Lei Bai}, \bibinfo{person}{Lina Yao}, \bibinfo{person}{Can Li}, \bibinfo{person}{Xianzhi Wang}, {and} \bibinfo{person}{Can Wang}.} \bibinfo{year}{2020}\natexlab{}.
\newblock \showarticletitle{Adaptive graph convolutional recurrent network for traffic forecasting}. \bibinfo{publisher}{International Conference on Neural Information Processing Systems}.
\newblock


\bibitem[\protect\citeauthoryear{Bao, Xu, Zhu, Wang, and Li}{Bao et~al\mbox{.}}{2021}]%
        {bao2021successive}
\bibfield{author}{\bibinfo{person}{Ting Bao}, \bibinfo{person}{Lei Xu}, \bibinfo{person}{Liehuang Zhu}, \bibinfo{person}{Lihong Wang}, {and} \bibinfo{person}{Tielei Li}.} \bibinfo{year}{2021}\natexlab{}.
\newblock \showarticletitle{Successive point-of-interest recommendation with personalized local differential privacy}.
\newblock  \bibinfo{volume}{70}, \bibinfo{number}{10} (\bibinfo{year}{2021}), \bibinfo{pages}{10477--10488}.
\newblock


\bibitem[\protect\citeauthoryear{Blumberg and Eckersley}{Blumberg and Eckersley}{2009}]%
        {blumberg2009locational}
\bibfield{author}{\bibinfo{person}{Andrew~J Blumberg} {and} \bibinfo{person}{Peter Eckersley}.} \bibinfo{year}{2009}\natexlab{}.
\newblock \showarticletitle{On locational privacy, and how to avoid losing it forever}.
\newblock  (\bibinfo{year}{2009}).
\newblock


\bibitem[\protect\citeauthoryear{Bordenabe, Chatzikokolakis, and Palamidessi}{Bordenabe et~al\mbox{.}}{2014}]%
        {bordenabe2014optimal}
\bibfield{author}{\bibinfo{person}{Nicol{\'a}s~E Bordenabe}, \bibinfo{person}{Konstantinos Chatzikokolakis}, {and} \bibinfo{person}{Catuscia Palamidessi}.} \bibinfo{year}{2014}\natexlab{}.
\newblock \showarticletitle{Optimal geo-indistinguishable mechanisms for location privacy}. \bibinfo{publisher}{ACM SIGSAC Conference on Computer and Communications Security}.
\newblock


\bibitem[\protect\citeauthoryear{Carlini, Chien, Nasr, Song, Terzis, and Tramer}{Carlini et~al\mbox{.}}{2022}]%
        {carlini2022membership}
\bibfield{author}{\bibinfo{person}{Nicholas Carlini}, \bibinfo{person}{Steve Chien}, \bibinfo{person}{Milad Nasr}, \bibinfo{person}{Shuang Song}, \bibinfo{person}{Andreas Terzis}, {and} \bibinfo{person}{Florian Tramer}.} \bibinfo{year}{2022}\natexlab{}.
\newblock \showarticletitle{Membership inference attacks from first principles}. \bibinfo{publisher}{IEEE Symposium on Security and Privacy}.
\newblock


\bibitem[\protect\citeauthoryear{Carlini, Hayes, Nasr, Jagielski, Sehwag, Tramer, Balle, Ippolito, and Wallace}{Carlini et~al\mbox{.}}{2023}]%
        {carlini2023extracting}
\bibfield{author}{\bibinfo{person}{Nicolas Carlini}, \bibinfo{person}{Jamie Hayes}, \bibinfo{person}{Milad Nasr}, \bibinfo{person}{Matthew Jagielski}, \bibinfo{person}{Vikash Sehwag}, \bibinfo{person}{Florian Tramer}, \bibinfo{person}{Borja Balle}, \bibinfo{person}{Daphne Ippolito}, {and} \bibinfo{person}{Eric Wallace}.} \bibinfo{year}{2023}\natexlab{}.
\newblock \showarticletitle{Extracting training data from diffusion models}. \bibinfo{publisher}{32nd USENIX Security Symposium (USENIX Security 23)}, \bibinfo{pages}{5253--5270}.
\newblock


\bibitem[\protect\citeauthoryear{Carlini, Liu, Erlingsson, Kos, and Song}{Carlini et~al\mbox{.}}{2019}]%
        {carlini2019secret}
\bibfield{author}{\bibinfo{person}{Nicholas Carlini}, \bibinfo{person}{Chang Liu}, \bibinfo{person}{{\'U}lfar Erlingsson}, \bibinfo{person}{Jernej Kos}, {and} \bibinfo{person}{Dawn Song}.} \bibinfo{year}{2019}\natexlab{}.
\newblock \showarticletitle{The Secret Sharer: Evaluating and testing unintended memorization in neural networks}. \bibinfo{publisher}{USENIX Security Symposium}.
\newblock


\bibitem[\protect\citeauthoryear{Carlini, Tramer, Wallace, Jagielski, Herbert-Voss, Lee, Roberts, Brown, Song, Erlingsson, et~al\mbox{.}}{Carlini et~al\mbox{.}}{2021}]%
        {carlini2021extracting}
\bibfield{author}{\bibinfo{person}{Nicholas Carlini}, \bibinfo{person}{Florian Tramer}, \bibinfo{person}{Eric Wallace}, \bibinfo{person}{Matthew Jagielski}, \bibinfo{person}{Ariel Herbert-Voss}, \bibinfo{person}{Katherine Lee}, \bibinfo{person}{Adam Roberts}, \bibinfo{person}{Tom Brown}, \bibinfo{person}{Dawn Song}, \bibinfo{person}{Ulfar Erlingsson}, {et~al\mbox{.}}} \bibinfo{year}{2021}\natexlab{}.
\newblock \showarticletitle{Extracting training data from large language models}. \bibinfo{publisher}{30th USENIX Security Symposium (USENIX Security 21)}, \bibinfo{pages}{2633--2650}.
\newblock


\bibitem[\protect\citeauthoryear{Chen, Liu, and Yu}{Chen et~al\mbox{.}}{2014}]%
        {chen2014nlpmm}
\bibfield{author}{\bibinfo{person}{Meng Chen}, \bibinfo{person}{Yang Liu}, {and} \bibinfo{person}{Xiaohui Yu}.} \bibinfo{year}{2014}\natexlab{}.
\newblock \showarticletitle{Nlpmm: A next location predictor with markov modeling}. \bibinfo{publisher}{Advances in Knowledge Discovery and Data Mining: Pacific-Asia Conference}.
\newblock


\bibitem[\protect\citeauthoryear{Cheng, Yang, Lyu, and King}{Cheng et~al\mbox{.}}{2013}]%
        {cheng2013you}
\bibfield{author}{\bibinfo{person}{Chen Cheng}, \bibinfo{person}{Haiqin Yang}, \bibinfo{person}{Michael~R Lyu}, {and} \bibinfo{person}{Irwin King}.} \bibinfo{year}{2013}\natexlab{}.
\newblock \showarticletitle{Where you like to go next: Successive point-of-interest recommendation}. \bibinfo{publisher}{International Joint Conference on Artificial Intelligence}.
\newblock


\bibitem[\protect\citeauthoryear{Cho, Myers, and Leskovec}{Cho et~al\mbox{.}}{2011}]%
        {cho2011friendship}
\bibfield{author}{\bibinfo{person}{Eunjoon Cho}, \bibinfo{person}{Seth~A Myers}, {and} \bibinfo{person}{Jure Leskovec}.} \bibinfo{year}{2011}\natexlab{}.
\newblock \showarticletitle{Friendship and mobility: user movement in location-based social networks}. \bibinfo{publisher}{ACM SIGKDD International Conference on Knowledge Discovery and Data Mining}.
\newblock


\bibitem[\protect\citeauthoryear{Ding, Xi, Wu, Liu, Wang, and Zhou}{Ding et~al\mbox{.}}{2022}]%
        {poi_embedding_privacy_1}
\bibfield{author}{\bibinfo{person}{Jiaxin Ding}, \bibinfo{person}{Shichuan Xi}, \bibinfo{person}{Kailong Wu}, \bibinfo{person}{Pan Liu}, \bibinfo{person}{Xinbing Wang}, {and} \bibinfo{person}{Chenghu Zhou}.} \bibinfo{year}{2022}\natexlab{}.
\newblock \showarticletitle{Analyzing sensitive information leakage in trajectory embedding models}. \bibinfo{publisher}{International Conference on Advances in Geographic Information Systems}.
\newblock


\bibitem[\protect\citeauthoryear{Dwork, Roth, et~al\mbox{.}}{Dwork et~al\mbox{.}}{2014}]%
        {DWO14}
\bibfield{author}{\bibinfo{person}{Cynthia Dwork}, \bibinfo{person}{Aaron Roth}, {et~al\mbox{.}}} \bibinfo{year}{2014}\natexlab{}.
\newblock \showarticletitle{The algorithmic foundations of differential privacy}.
\newblock  (\bibinfo{year}{2014}).
\newblock


\bibitem[\protect\citeauthoryear{{EU}}{{EU}}{2018}]%
        {gdpr}
\bibfield{author}{\bibinfo{person}{{EU}}.} \bibinfo{year}{2018}\natexlab{}.
\newblock \bibinfo{title}{{General data protection regulation}}.
\newblock \bibinfo{howpublished}{\url{https://en.wikipedia.org/wiki/General_Data_Protection_Regulation}}.
\newblock


\bibitem[\protect\citeauthoryear{Fan, Lewis, and Dauphin}{Fan et~al\mbox{.}}{2018}]%
        {fan2018hierarchical}
\bibfield{author}{\bibinfo{person}{Angela Fan}, \bibinfo{person}{Mike Lewis}, {and} \bibinfo{person}{Yann Dauphin}.} \bibinfo{year}{2018}\natexlab{}.
\newblock \showarticletitle{Hierarchical neural story generation}.
\newblock  (\bibinfo{year}{2018}).
\newblock


\bibitem[\protect\citeauthoryear{Fernandes, Dras, and McIver}{Fernandes et~al\mbox{.}}{2019}]%
        {fernandes2019generalised}
\bibfield{author}{\bibinfo{person}{Natasha Fernandes}, \bibinfo{person}{Mark Dras}, {and} \bibinfo{person}{Annabelle McIver}.} \bibinfo{year}{2019}\natexlab{}.
\newblock \showarticletitle{Generalised differential privacy for text document processing}. Springer International Publishing, \bibinfo{publisher}{Principles of Security and Trust: 8th International Conference, POST 2019, Held as Part of the European Joint Conferences on Theory and Practice of Software, ETAPS 2019, Prague, Czech Republic, April 6--11, 2019, Proceedings 8}, \bibinfo{pages}{123--148}.
\newblock


\bibitem[\protect\citeauthoryear{Fredrikson, Jha, and Ristenpart}{Fredrikson et~al\mbox{.}}{2015a}]%
        {10.1145/2810103.2813677}
\bibfield{author}{\bibinfo{person}{Matt Fredrikson}, \bibinfo{person}{Somesh Jha}, {and} \bibinfo{person}{Thomas Ristenpart}.} \bibinfo{year}{2015}\natexlab{a}.
\newblock \showarticletitle{Model inversion attacks that exploit confidence information and basic countermeasures}. \bibinfo{publisher}{ACM SIGSAC Conference on Computer and Communications Security}.
\newblock


\bibitem[\protect\citeauthoryear{Fredrikson, Jha, and Ristenpart}{Fredrikson et~al\mbox{.}}{2015b}]%
        {fredrikson2015model}
\bibfield{author}{\bibinfo{person}{Matt Fredrikson}, \bibinfo{person}{Somesh Jha}, {and} \bibinfo{person}{Thomas Ristenpart}.} \bibinfo{year}{2015}\natexlab{b}.
\newblock \showarticletitle{Model Inversion Attacks That Exploit Confidence Information and Basic Countermeasures}. \bibinfo{publisher}{Proceedings of the 22nd ACM SIGSAC Conference on Computer and Communications Security}, \bibinfo{address}{New York, NY, USA}, \bibinfo{pages}{1322–1333}.
\newblock
\showISBNx{9781450338325}
\urldef\tempurl%
\url{https://doi.org/10.1145/2810103.2813677}
\showDOI{\tempurl}


\bibitem[\protect\citeauthoryear{Gambs, Killijian, and del Prado~Cortez}{Gambs et~al\mbox{.}}{2014}]%
        {gambs2014anonymization}
\bibfield{author}{\bibinfo{person}{S{\'e}bastien Gambs}, \bibinfo{person}{Marc-Olivier Killijian}, {and} \bibinfo{person}{Miguel~N{\'u}{\~n}ez del Prado~Cortez}.} \bibinfo{year}{2014}\natexlab{}.
\newblock \showarticletitle{De-anonymization attack on geolocated data}.
\newblock  (\bibinfo{year}{2014}).
\newblock


\bibitem[\protect\citeauthoryear{Gedik and Liu}{Gedik and Liu}{2005}]%
        {gedik2005location}
\bibfield{author}{\bibinfo{person}{Bugra Gedik} {and} \bibinfo{person}{Ling Liu}.} \bibinfo{year}{2005}\natexlab{}.
\newblock \showarticletitle{Location privacy in mobile systems: A personalized anonymization model}. \bibinfo{publisher}{IEEE International Conference on Distributed Computing Systems}.
\newblock


\bibitem[\protect\citeauthoryear{Golle and Partridge}{Golle and Partridge}{2009}]%
        {poi_linkage_percom2}
\bibfield{author}{\bibinfo{person}{Philippe Golle} {and} \bibinfo{person}{Kurt Partridge}.} \bibinfo{year}{2009}\natexlab{}.
\newblock \showarticletitle{On the anonymity of home/work location pairs}. Springer, \bibinfo{publisher}{Pervasive Computing: 7th International Conference, Pervasive 2009, Nara, Japan, May 11-14, 2009. Proceedings 7}, \bibinfo{pages}{390--397}.
\newblock


\bibitem[\protect\citeauthoryear{Gomrokchi, Amin, Aboutalebi, Wong, and Precup}{Gomrokchi et~al\mbox{.}}{2023}]%
        {gomrokchi2023membership}
\bibfield{author}{\bibinfo{person}{Maziar Gomrokchi}, \bibinfo{person}{Susan Amin}, \bibinfo{person}{Hossein Aboutalebi}, \bibinfo{person}{Alexander Wong}, {and} \bibinfo{person}{Doina Precup}.} \bibinfo{year}{2023}\natexlab{}.
\newblock \showarticletitle{Membership Inference Attacks Against Temporally Correlated Data in Deep Reinforcement Learning}.
\newblock \bibinfo{journal}{\emph{IEEE Access}} (\bibinfo{year}{2023}).
\newblock


\bibitem[\protect\citeauthoryear{Goodfellow, Bengio, and Courville}{Goodfellow et~al\mbox{.}}{2016}]%
        {GBC16}
\bibfield{author}{\bibinfo{person}{Ian Goodfellow}, \bibinfo{person}{Yoshua Bengio}, {and} \bibinfo{person}{Aaron Courville}.} \bibinfo{year}{2016}\natexlab{}.
\newblock \showarticletitle{Deep Learning}. \bibinfo{publisher}{MIT Press}, \bibinfo{pages}{221--265}.
\newblock


\bibitem[\protect\citeauthoryear{Gruteser and Grunwald}{Gruteser and Grunwald}{2003}]%
        {gruteser2003anonymous}
\bibfield{author}{\bibinfo{person}{Marco Gruteser} {and} \bibinfo{person}{Dirk Grunwald}.} \bibinfo{year}{2003}\natexlab{}.
\newblock \showarticletitle{Anonymous usage of location-based services through spatial and temporal cloaking}. \bibinfo{publisher}{International Conference on Mobile Systems, Applications and Services}.
\newblock


\bibitem[\protect\citeauthoryear{Hara, Suzuki, Iwata, Arase, and Xie}{Hara et~al\mbox{.}}{2016}]%
        {hara2016dummy}
\bibfield{author}{\bibinfo{person}{Takahiro Hara}, \bibinfo{person}{Akiyoshi Suzuki}, \bibinfo{person}{Mayu Iwata}, \bibinfo{person}{Yuki Arase}, {and} \bibinfo{person}{Xing Xie}.} \bibinfo{year}{2016}\natexlab{}.
\newblock \showarticletitle{Dummy-based user location anonymization under real-world constraints}.
\newblock  (\bibinfo{year}{2016}).
\newblock


\bibitem[\protect\citeauthoryear{Hassan, Hussain, and Bates}{Hassan et~al\mbox{.}}{2018}]%
        {hassan2018analysis}
\bibfield{author}{\bibinfo{person}{Wajih~Ul Hassan}, \bibinfo{person}{Saad Hussain}, {and} \bibinfo{person}{Adam Bates}.} \bibinfo{year}{2018}\natexlab{}.
\newblock \showarticletitle{Analysis of privacy protections in fitness tracking social networks-or-you can run, but can you hide?} \bibinfo{publisher}{USENIX Security Symposium}.
\newblock


\bibitem[\protect\citeauthoryear{He, Li, and Liao}{He et~al\mbox{.}}{2017}]%
        {he2017category}
\bibfield{author}{\bibinfo{person}{Jing He}, \bibinfo{person}{Xin Li}, {and} \bibinfo{person}{Lejian Liao}.} \bibinfo{year}{2017}\natexlab{}.
\newblock \showarticletitle{Category-aware next point-of-interest recommendation via listwise bayesian personalized ranking.} \bibinfo{publisher}{International Joint Conference on Artificial Intelligence}.
\newblock


\bibitem[\protect\citeauthoryear{Henne, Szongott, and Smith}{Henne et~al\mbox{.}}{2013}]%
        {henne2013snapme}
\bibfield{author}{\bibinfo{person}{Benjamin Henne}, \bibinfo{person}{Christian Szongott}, {and} \bibinfo{person}{Matthew Smith}.} \bibinfo{year}{2013}\natexlab{}.
\newblock \showarticletitle{SnapMe if you can: Privacy threats of other peoples' geo-tagged media and what we can do about it}. \bibinfo{publisher}{ACM Conference on Security and Privacy in Wireless and Mobile Networks}.
\newblock


\bibitem[\protect\citeauthoryear{Hoh, Gruteser, Xiong, and Alrabady}{Hoh et~al\mbox{.}}{2006}]%
        {poi_linkage_percom3}
\bibfield{author}{\bibinfo{person}{Baik Hoh}, \bibinfo{person}{Marco Gruteser}, \bibinfo{person}{Hui Xiong}, {and} \bibinfo{person}{Ansaf Alrabady}.} \bibinfo{year}{2006}\natexlab{}.
\newblock \showarticletitle{Enhancing security and privacy in traffic-monitoring systems}.
\newblock  (\bibinfo{year}{2006}).
\newblock


\bibitem[\protect\citeauthoryear{Islam, Mohammad, Das, and Ali}{Islam et~al\mbox{.}}{2020}]%
        {islam2020survey}
\bibfield{author}{\bibinfo{person}{Md.~Ashraful Islam}, \bibinfo{person}{Mir~Mahathir Mohammad}, \bibinfo{person}{Sarkar Snigdha~Sarathi Das}, {and} \bibinfo{person}{Mohammed~Eunus Ali}.} \bibinfo{year}{2020}\natexlab{}.
\newblock \bibinfo{title}{A survey on deep learning based Point-Of-Interest (POI) recommendations}.
\newblock
\newblock
\showeprint[arxiv]{cs.IR/2011.10187}


\bibitem[\protect\citeauthoryear{Jagielski, Ullman, and Oprea}{Jagielski et~al\mbox{.}}{2020}]%
        {jagielski2020auditing}
\bibfield{author}{\bibinfo{person}{Matthew Jagielski}, \bibinfo{person}{Jonathan Ullman}, {and} \bibinfo{person}{Alina Oprea}.} \bibinfo{year}{2020}\natexlab{}.
\newblock \showarticletitle{Auditing differentially private machine learning: How private is private SGD?}
\newblock  (\bibinfo{year}{2020}).
\newblock


\bibitem[\protect\citeauthoryear{Jiang, Wang, and Hu}{Jiang et~al\mbox{.}}{2007}]%
        {jiang2007preserving}
\bibfield{author}{\bibinfo{person}{Tao Jiang}, \bibinfo{person}{Helen~J Wang}, {and} \bibinfo{person}{Yih-Chun Hu}.} \bibinfo{year}{2007}\natexlab{}.
\newblock \showarticletitle{Preserving location privacy in wireless LANs}. \bibinfo{publisher}{International Conference on Mobile Systems, Applications and Services}.
\newblock


\bibitem[\protect\citeauthoryear{Kong and Wu}{Kong and Wu}{2018}]%
        {kong2018hst}
\bibfield{author}{\bibinfo{person}{Dejiang Kong} {and} \bibinfo{person}{Fei Wu}.} \bibinfo{year}{2018}\natexlab{}.
\newblock \showarticletitle{HST-LSTM: A hierarchical spatial-temporal long-short term memory network for location prediction.} \bibinfo{publisher}{International Joint Conference on Artificial Intelligence}.
\newblock


\bibitem[\protect\citeauthoryear{Krumm}{Krumm}{2007a}]%
        {krumm2007inference}
\bibfield{author}{\bibinfo{person}{John Krumm}.} \bibinfo{year}{2007}\natexlab{a}.
\newblock \showarticletitle{Inference attacks on location tracks}. Springer, \bibinfo{publisher}{Pervasive Computing: 5th International Conference, PERVASIVE 2007, Toronto, Canada, May 13-16, 2007. Proceedings 5}, \bibinfo{pages}{127--143}.
\newblock


\bibitem[\protect\citeauthoryear{Krumm}{Krumm}{2007b}]%
        {poi_linkage_percom}
\bibfield{author}{\bibinfo{person}{John Krumm}.} \bibinfo{year}{2007}\natexlab{b}.
\newblock \showarticletitle{Inference attacks on location tracks}. Springer, \bibinfo{publisher}{Pervasive Computing: 5th International Conference, PERVASIVE 2007, Toronto, Canada, May 13-16, 2007. Proceedings 5}, \bibinfo{pages}{127--143}.
\newblock


\bibitem[\protect\citeauthoryear{Lan, Ma, Huang, Wang, Yang, and Li}{Lan et~al\mbox{.}}{2022}]%
        {lan2022dstagnn_traffic}
\bibfield{author}{\bibinfo{person}{Shiyong Lan}, \bibinfo{person}{Yitong Ma}, \bibinfo{person}{Weikang Huang}, \bibinfo{person}{Wenwu Wang}, \bibinfo{person}{Hongyu Yang}, {and} \bibinfo{person}{Pyang Li}.} \bibinfo{year}{2022}\natexlab{}.
\newblock \showarticletitle{Dstagnn: Dynamic spatial-temporal aware graph neural network for traffic flow forecasting}. \bibinfo{publisher}{International Conference on Machine Learning}.
\newblock


\bibitem[\protect\citeauthoryear{Lian, Wu, Ge, Xie, and Chen}{Lian et~al\mbox{.}}{2020}]%
        {lian2020geography}
\bibfield{author}{\bibinfo{person}{Defu Lian}, \bibinfo{person}{Yongji Wu}, \bibinfo{person}{Yong Ge}, \bibinfo{person}{Xing Xie}, {and} \bibinfo{person}{Enhong Chen}.} \bibinfo{year}{2020}\natexlab{}.
\newblock \showarticletitle{Geography-aware sequential location recommendation}. \bibinfo{publisher}{ACM SIGKDD International Conference on Knowledge Discovery and Data Mining}.
\newblock


\bibitem[\protect\citeauthoryear{Lian, Zhao, Xie, Sun, Chen, and Rui}{Lian et~al\mbox{.}}{2014}]%
        {lian2014geomf}
\bibfield{author}{\bibinfo{person}{Defu Lian}, \bibinfo{person}{Cong Zhao}, \bibinfo{person}{Xing Xie}, \bibinfo{person}{Guangzhong Sun}, \bibinfo{person}{Enhong Chen}, {and} \bibinfo{person}{Yong Rui}.} \bibinfo{year}{2014}\natexlab{}.
\newblock \showarticletitle{GeoMF: Joint geographical modeling and matrix factorization for point-of-interest recommendation}. \bibinfo{publisher}{ACM SIGKDD International Conference on Knowledge Discovery and Data Mining}.
\newblock


\bibitem[\protect\citeauthoryear{Liu, Wu, Wang, and Tan}{Liu et~al\mbox{.}}{2016}]%
        {Liu2016PredictingTN}
\bibfield{author}{\bibinfo{person}{Q. Liu}, \bibinfo{person}{Shu Wu}, \bibinfo{person}{Liang Wang}, {and} \bibinfo{person}{Tieniu Tan}.} \bibinfo{year}{2016}\natexlab{}.
\newblock \showarticletitle{Predicting the next location: A recurrent model with spatial and temporal contexts}. \bibinfo{publisher}{AAAI Conference on Artificial Intelligence}.
\newblock


\bibitem[\protect\citeauthoryear{Liu, Liu, Lu, Teng, Zhu, and Xiong}{Liu et~al\mbox{.}}{2017}]%
        {site_rec_kdd17}
\bibfield{author}{\bibinfo{person}{Yanchi Liu}, \bibinfo{person}{Chuanren Liu}, \bibinfo{person}{Xinjiang Lu}, \bibinfo{person}{Mingfei Teng}, \bibinfo{person}{Hengshu Zhu}, {and} \bibinfo{person}{Hui Xiong}.} \bibinfo{year}{2017}\natexlab{}.
\newblock \showarticletitle{Point-of-Interest demand modeling with human mobility patterns}. \bibinfo{publisher}{ACM SIGKDD International Conference on Knowledge Discovery and Data Mining}.
\newblock


\bibitem[\protect\citeauthoryear{Luo, Liu, and Liu}{Luo et~al\mbox{.}}{2021}]%
        {luo2021stan}
\bibfield{author}{\bibinfo{person}{Yingtao Luo}, \bibinfo{person}{Qiang Liu}, {and} \bibinfo{person}{Zhaocheng Liu}.} \bibinfo{year}{2021}\natexlab{}.
\newblock \showarticletitle{Stan: Spatio-temporal attention network for next location recommendation}. \bibinfo{publisher}{Web Conference}.
\newblock


\bibitem[\protect\citeauthoryear{Meng, Cui, He, Su, and Gao}{Meng et~al\mbox{.}}{2017}]%
        {meng2017travel}
\bibfield{author}{\bibinfo{person}{Chuishi Meng}, \bibinfo{person}{Yu Cui}, \bibinfo{person}{Qing He}, \bibinfo{person}{Lu Su}, {and} \bibinfo{person}{Jing Gao}.} \bibinfo{year}{2017}\natexlab{}.
\newblock \showarticletitle{Travel purpose inference with GPS trajectories, POIs, and geo-tagged social media data}. \bibinfo{publisher}{IEEE International Conference on Big Data}.
\newblock


\bibitem[\protect\citeauthoryear{Meyerowitz and Roy~Choudhury}{Meyerowitz and Roy~Choudhury}{2009}]%
        {meyerowitz2009hiding}
\bibfield{author}{\bibinfo{person}{Joseph Meyerowitz} {and} \bibinfo{person}{Romit Roy~Choudhury}.} \bibinfo{year}{2009}\natexlab{}.
\newblock \showarticletitle{Hiding stars with fireworks: location privacy through camouflage}. \bibinfo{publisher}{Annual International Conference on Mobile Computing and Networking}.
\newblock


\bibitem[\protect\citeauthoryear{{Mike Boland}}{{Mike Boland}}{2021}]%
        {4sq_place}
\bibfield{author}{\bibinfo{person}{{Mike Boland}}.} \bibinfo{year}{2021}\natexlab{}.
\newblock \bibinfo{title}{Foursquare’s power play continues with relaunched places and new API}.
\newblock
\newblock
\urldef\tempurl%
\url{https://www.localogy.com/2021/03/foursquares-power-play-continues-with-relaunched-places-and-new-api/}
\showURL{%
\tempurl}


\bibitem[\protect\citeauthoryear{Powar and Beresford}{Powar and Beresford}{2023}]%
        {powar2023sok}
\bibfield{author}{\bibinfo{person}{Jovan Powar} {and} \bibinfo{person}{Alastair~R Beresford}.} \bibinfo{year}{2023}\natexlab{}.
\newblock \showarticletitle{SoK: Managing risks of linkage attacks on data privacy}.
\newblock  (\bibinfo{year}{2023}).
\newblock


\bibitem[\protect\citeauthoryear{Pyrgelis, Troncoso, and De~Cristofaro}{Pyrgelis et~al\mbox{.}}{2017}]%
        {pyrgelis2017knock}
\bibfield{author}{\bibinfo{person}{Apostolos Pyrgelis}, \bibinfo{person}{Carmela Troncoso}, {and} \bibinfo{person}{Emiliano De~Cristofaro}.} \bibinfo{year}{2017}\natexlab{}.
\newblock \showarticletitle{Knock knock, who's there? Membership inference on aggregate location data}.
\newblock  (\bibinfo{year}{2017}).
\newblock


\bibitem[\protect\citeauthoryear{Pyrgelis, Troncoso, and De~Cristofaro}{Pyrgelis et~al\mbox{.}}{2020}]%
        {pyrgelis2020measuring}
\bibfield{author}{\bibinfo{person}{Apostolos Pyrgelis}, \bibinfo{person}{Carmela Troncoso}, {and} \bibinfo{person}{Emiliano De~Cristofaro}.} \bibinfo{year}{2020}\natexlab{}.
\newblock \showarticletitle{Measuring membership privacy on aggregate location time-series}.
\newblock \bibinfo{journal}{\emph{Proceedings of the ACM on Measurement and Analysis of Computing Systems}} \bibinfo{volume}{4}, \bibinfo{number}{2} (\bibinfo{year}{2020}), \bibinfo{pages}{1--28}.
\newblock


\bibitem[\protect\citeauthoryear{Ren, Hu, Li, Wang, Wu, Li, and Hu}{Ren et~al\mbox{.}}{2023}]%
        {ren2023your}
\bibfield{author}{\bibinfo{person}{Lingfei Ren}, \bibinfo{person}{Ruimin Hu}, \bibinfo{person}{Dengshi Li}, \bibinfo{person}{Zheng Wang}, \bibinfo{person}{Junhang Wu}, \bibinfo{person}{Xixi Li}, {and} \bibinfo{person}{Wenyi Hu}.} \bibinfo{year}{2023}\natexlab{}.
\newblock \showarticletitle{Who is your friend: inferring cross-regional friendship from mobility profiles}.
\newblock  (\bibinfo{year}{2023}).
\newblock


\bibitem[\protect\citeauthoryear{Salem, Zhang, Humbert, Berrang, Fritz, and Backes}{Salem et~al\mbox{.}}{2018}]%
        {salem2018ml}
\bibfield{author}{\bibinfo{person}{Ahmed Salem}, \bibinfo{person}{Yang Zhang}, \bibinfo{person}{Mathias Humbert}, \bibinfo{person}{Pascal Berrang}, \bibinfo{person}{Mario Fritz}, {and} \bibinfo{person}{Michael Backes}.} \bibinfo{year}{2018}\natexlab{}.
\newblock \showarticletitle{Ml-leaks: Model and data independent membership inference attacks and defenses on machine learning models}.
\newblock  (\bibinfo{year}{2018}).
\newblock


\bibitem[\protect\citeauthoryear{Shi, Chen, Zhang, Jia, and Yu}{Shi et~al\mbox{.}}{2022a}]%
        {shi2022just}
\bibfield{author}{\bibinfo{person}{Weiyan Shi}, \bibinfo{person}{Si Chen}, \bibinfo{person}{Chiyuan Zhang}, \bibinfo{person}{Ruoxi Jia}, {and} \bibinfo{person}{Zhou Yu}.} \bibinfo{year}{2022}\natexlab{a}.
\newblock \showarticletitle{Just fine-tune twice: Selective differential privacy for large language models}.
\newblock  (\bibinfo{year}{2022}).
\newblock


\bibitem[\protect\citeauthoryear{Shi, Cui, Li, Jia, and Yu}{Shi et~al\mbox{.}}{2022b}]%
        {SCLJY22}
\bibfield{author}{\bibinfo{person}{Weiyan Shi}, \bibinfo{person}{Aiqi Cui}, \bibinfo{person}{Evan Li}, \bibinfo{person}{Ruoxi Jia}, {and} \bibinfo{person}{Zhou Yu}.} \bibinfo{year}{2022}\natexlab{b}.
\newblock \showarticletitle{Selective Differential Privacy for Language Modeling}. \bibinfo{publisher}{Conference of the North American Chapter of the Association for Computational Linguistics: Human Language Technologies}.
\newblock


\bibitem[\protect\citeauthoryear{Shokri, Stronati, Song, and Shmatikov}{Shokri et~al\mbox{.}}{2017a}]%
        {shokri2017membership}
\bibfield{author}{\bibinfo{person}{Reza Shokri}, \bibinfo{person}{Marco Stronati}, \bibinfo{person}{Congzheng Song}, {and} \bibinfo{person}{Vitaly Shmatikov}.} \bibinfo{year}{2017}\natexlab{a}.
\newblock \showarticletitle{Membership inference attacks against machine learning models}. \bibinfo{publisher}{IEEE Symposium on Security and Privacy}.
\newblock


\bibitem[\protect\citeauthoryear{Shokri, Stronati, Song, and Shmatikov}{Shokri et~al\mbox{.}}{2017b}]%
        {shokri_2017_MIA}
\bibfield{author}{\bibinfo{person}{R. Shokri}, \bibinfo{person}{M. Stronati}, \bibinfo{person}{C. Song}, {and} \bibinfo{person}{V. Shmatikov}.} \bibinfo{year}{2017}\natexlab{b}.
\newblock \showarticletitle{Membership inference attacks against machine learning models}. \bibinfo{publisher}{IEEE Symposium on Security and Privacy}.
\newblock


\bibitem[\protect\citeauthoryear{Shokri, Theodorakopoulos, Papadimitratos, Kazemi, and Hubaux}{Shokri et~al\mbox{.}}{2013}]%
        {shokri2013hiding}
\bibfield{author}{\bibinfo{person}{Reza Shokri}, \bibinfo{person}{George Theodorakopoulos}, \bibinfo{person}{Panos Papadimitratos}, \bibinfo{person}{Ehsan Kazemi}, {and} \bibinfo{person}{Jean-Pierre Hubaux}.} \bibinfo{year}{2013}\natexlab{}.
\newblock \showarticletitle{Hiding in the mobile crowd: Location privacy through collaboration}.
\newblock  (\bibinfo{year}{2013}).
\newblock


\bibitem[\protect\citeauthoryear{{Shubham Sharma}}{{Shubham Sharma}}{2022}]%
        {4sq_news2}
\bibfield{author}{\bibinfo{person}{{Shubham Sharma}}.} \bibinfo{year}{2022}\natexlab{}.
\newblock \bibinfo{title}{How Foursquare helps enterprises drive positive results with geospatial technology}.
\newblock
\newblock
\urldef\tempurl%
\url{https://venturebeat.com/data-infrastructure/how-foursquare-helps-enterprises/}
\showURL{%
\tempurl}


\bibitem[\protect\citeauthoryear{Srivatsa and Hicks}{Srivatsa and Hicks}{2012a}]%
        {srivatsa2012deanonymizing}
\bibfield{author}{\bibinfo{person}{Mudhakar Srivatsa} {and} \bibinfo{person}{Mike Hicks}.} \bibinfo{year}{2012}\natexlab{a}.
\newblock \showarticletitle{Deanonymizing mobility traces: Using social network as a side-channel}. \bibinfo{publisher}{ACM Conference on Computer and Communications Security}.
\newblock


\bibitem[\protect\citeauthoryear{Srivatsa and Hicks}{Srivatsa and Hicks}{2012b}]%
        {poi_social_communication}
\bibfield{author}{\bibinfo{person}{Mudhakar Srivatsa} {and} \bibinfo{person}{Mike Hicks}.} \bibinfo{year}{2012}\natexlab{b}.
\newblock \showarticletitle{Deanonymizing mobility traces: Using social network as a side-channel}. \bibinfo{publisher}{ACM Conference on Computer and Communications Security}.
\newblock


\bibitem[\protect\citeauthoryear{Sun, Qian, Chen, Liang, Nguyen, and Yin}{Sun et~al\mbox{.}}{2020}]%
        {sun2020go}
\bibfield{author}{\bibinfo{person}{Ke Sun}, \bibinfo{person}{Tieyun Qian}, \bibinfo{person}{Tong Chen}, \bibinfo{person}{Yile Liang}, \bibinfo{person}{Quoc Viet~Hung Nguyen}, {and} \bibinfo{person}{Hongzhi Yin}.} \bibinfo{year}{2020}\natexlab{}.
\newblock \showarticletitle{Where to go next: Modeling long-and short-term user preferences for point-of-interest recommendation}. \bibinfo{publisher}{AAAI Conference on Artificial Intelligence}.
\newblock


\bibitem[\protect\citeauthoryear{Tram{\`e}r, Shokri, San~Joaquin, Le, Jagielski, Hong, and Carlini}{Tram{\`e}r et~al\mbox{.}}{2022}]%
        {tramer2022truth}
\bibfield{author}{\bibinfo{person}{Florian Tram{\`e}r}, \bibinfo{person}{Reza Shokri}, \bibinfo{person}{Ayrton San~Joaquin}, \bibinfo{person}{Hoang Le}, \bibinfo{person}{Matthew Jagielski}, \bibinfo{person}{Sanghyun Hong}, {and} \bibinfo{person}{Nicholas Carlini}.} \bibinfo{year}{2022}\natexlab{}.
\newblock \showarticletitle{Truth serum: Poisoning machine learning models to reveal their secrets}. \bibinfo{publisher}{ACM SIGSAC Conference on Computer and Communications Security}.
\newblock


\bibitem[\protect\citeauthoryear{Vicente, Freni, Bettini, and Jensen}{Vicente et~al\mbox{.}}{2011a}]%
        {vicente2011location}
\bibfield{author}{\bibinfo{person}{Carmen~Ruiz Vicente}, \bibinfo{person}{Dario Freni}, \bibinfo{person}{Claudio Bettini}, {and} \bibinfo{person}{Christian~S Jensen}.} \bibinfo{year}{2011}\natexlab{a}.
\newblock \showarticletitle{Location-related privacy in geo-social networks}.
\newblock  (\bibinfo{year}{2011}).
\newblock


\bibitem[\protect\citeauthoryear{Vicente, Freni, Bettini, and Jensen}{Vicente et~al\mbox{.}}{2011b}]%
        {poi_social_internet_computing}
\bibfield{author}{\bibinfo{person}{Carmen~Ruiz Vicente}, \bibinfo{person}{Dario Freni}, \bibinfo{person}{Claudio Bettini}, {and} \bibinfo{person}{Christian~S Jensen}.} \bibinfo{year}{2011}\natexlab{b}.
\newblock \showarticletitle{Location-related privacy in geo-social networks}.
\newblock  (\bibinfo{year}{2011}).
\newblock


\bibitem[\protect\citeauthoryear{Vratonjic, Huguenin, Bindschaedler, and Hubaux}{Vratonjic et~al\mbox{.}}{2014}]%
        {vratonjic2014location}
\bibfield{author}{\bibinfo{person}{Nevena Vratonjic}, \bibinfo{person}{K{\'e}vin Huguenin}, \bibinfo{person}{Vincent Bindschaedler}, {and} \bibinfo{person}{Jean-Pierre Hubaux}.} \bibinfo{year}{2014}\natexlab{}.
\newblock \showarticletitle{A location-privacy threat stemming from the use of shared public IP addresses}.
\newblock  (\bibinfo{year}{2014}).
\newblock


\bibitem[\protect\citeauthoryear{Wang, Jiang, Jiang, Li, and Zhao}{Wang et~al\mbox{.}}{2021}]%
        {libcity}
\bibfield{author}{\bibinfo{person}{Jingyuan Wang}, \bibinfo{person}{Jiawei Jiang}, \bibinfo{person}{Wenjun Jiang}, \bibinfo{person}{Chao Li}, {and} \bibinfo{person}{Wayne~Xin Zhao}.} \bibinfo{year}{2021}\natexlab{}.
\newblock \showarticletitle{LibCity: An open library for traffic prediction}. \bibinfo{publisher}{International Conference on Advances in Geographic Information Systems}.
\newblock


\bibitem[\protect\citeauthoryear{Wang, Höpken, and Jannach}{Wang et~al\mbox{.}}{2023}]%
        {wang2023survey}
\bibfield{author}{\bibinfo{person}{Zehui Wang}, \bibinfo{person}{Wolfram Höpken}, {and} \bibinfo{person}{Dietmar Jannach}.} \bibinfo{year}{2023}\natexlab{}.
\newblock \bibinfo{title}{A survey on Point-of-Interest recommendations leveraging heterogeneous data}.
\newblock
\newblock
\showeprint[arxiv]{cs.IR/2308.07426}


\bibitem[\protect\citeauthoryear{Xin, Lu, Xu, Liu, Gu, Dou, and Xiong}{Xin et~al\mbox{.}}{2021}]%
        {xin2021outoftown}
\bibfield{author}{\bibinfo{person}{Haoran Xin}, \bibinfo{person}{Xinjiang Lu}, \bibinfo{person}{Tong Xu}, \bibinfo{person}{Hao Liu}, \bibinfo{person}{Jingjing Gu}, \bibinfo{person}{Dejing Dou}, {and} \bibinfo{person}{Hui Xiong}.} \bibinfo{year}{2021}\natexlab{}.
\newblock \bibinfo{title}{Out-of-town recommendation with travel intention modeling}.
\newblock
\newblock
\showeprint[arxiv]{cs.IR/2101.12555}


\bibitem[\protect\citeauthoryear{Xu, Mei, Liu, Zhao, and Ding}{Xu et~al\mbox{.}}{2023}]%
        {xu2023efficient}
\bibfield{author}{\bibinfo{person}{Chonghuan Xu}, \bibinfo{person}{Xinyao Mei}, \bibinfo{person}{Dongsheng Liu}, \bibinfo{person}{Kaidi Zhao}, {and} \bibinfo{person}{Austin~Shijun Ding}.} \bibinfo{year}{2023}\natexlab{}.
\newblock \showarticletitle{An efficient privacy-preserving point-of-interest recommendation model based on local differential privacy}.
\newblock  \bibinfo{volume}{9}, \bibinfo{number}{3} (\bibinfo{year}{2023}), \bibinfo{pages}{3277--3300}.
\newblock


\bibitem[\protect\citeauthoryear{Xu, Cui, Zhu, and Yang}{Xu et~al\mbox{.}}{2014}]%
        {poi_social_mm}
\bibfield{author}{\bibinfo{person}{Dan Xu}, \bibinfo{person}{Peng Cui}, \bibinfo{person}{Wenwu Zhu}, {and} \bibinfo{person}{Shiqiang Yang}.} \bibinfo{year}{2014}\natexlab{}.
\newblock \showarticletitle{Graph-based residence location inference for social media users}.
\newblock  (\bibinfo{year}{2014}).
\newblock


\bibitem[\protect\citeauthoryear{Yan, Xu, Mahmood, Dong, and Sheng}{Yan et~al\mbox{.}}{2022}]%
        {yan2022perturb}
\bibfield{author}{\bibinfo{person}{Yan Yan}, \bibinfo{person}{Fei Xu}, \bibinfo{person}{Adnan Mahmood}, \bibinfo{person}{Zhuoyue Dong}, {and} \bibinfo{person}{Quan~Z Sheng}.} \bibinfo{year}{2022}\natexlab{}.
\newblock \showarticletitle{Perturb and optimize users’ location privacy using geo-indistinguishability and location semantics}.
\newblock  (\bibinfo{year}{2022}).
\newblock


\bibitem[\protect\citeauthoryear{Yang, Fankhauser, Rosso, and Cudre-Mauroux}{Yang et~al\mbox{.}}{2020}]%
        {yang2020location}
\bibfield{author}{\bibinfo{person}{Dingqi Yang}, \bibinfo{person}{Benjamin Fankhauser}, \bibinfo{person}{Paolo Rosso}, {and} \bibinfo{person}{Philippe Cudre-Mauroux}.} \bibinfo{year}{2020}\natexlab{}.
\newblock \showarticletitle{Location prediction over sparse user mobility traces using rnns}. \bibinfo{publisher}{International Joint Conference on Artificial Intelligence}.
\newblock


\bibitem[\protect\citeauthoryear{Yang, Zhang, Zheng, and Yu}{Yang et~al\mbox{.}}{2014}]%
        {yang2014modeling}
\bibfield{author}{\bibinfo{person}{Dingqi Yang}, \bibinfo{person}{Daqing Zhang}, \bibinfo{person}{Vincent~W Zheng}, {and} \bibinfo{person}{Zhiyong Yu}.} \bibinfo{year}{2014}\natexlab{}.
\newblock \showarticletitle{Modeling user activity preference by leveraging user spatial temporal characteristics in LBSNs}.
\newblock  (\bibinfo{year}{2014}).
\newblock


\bibitem[\protect\citeauthoryear{Yang, Liu, and Zhao}{Yang et~al\mbox{.}}{2022}]%
        {10.1145/3477495.3531983}
\bibfield{author}{\bibinfo{person}{Song Yang}, \bibinfo{person}{Jiamou Liu}, {and} \bibinfo{person}{Kaiqi Zhao}.} \bibinfo{year}{2022}\natexlab{}.
\newblock \showarticletitle{GETNext: Trajectory flow map enhanced transformer for next POI recommendation}. \bibinfo{publisher}{ACM SIGIR Conference on Research and Development in Information Retrieval}.
\newblock


\bibitem[\protect\citeauthoryear{Zhang, Zhang, and Zhao}{Zhang et~al\mbox{.}}{2020}]%
        {zhang2020locmia}
\bibfield{author}{\bibinfo{person}{Guanglin Zhang}, \bibinfo{person}{Anqi Zhang}, {and} \bibinfo{person}{Ping Zhao}.} \bibinfo{year}{2020}\natexlab{}.
\newblock \showarticletitle{Locmia: Membership inference attacks against aggregated location data}.
\newblock  (\bibinfo{year}{2020}).
\newblock


\bibitem[\protect\citeauthoryear{Zhang, Chow, and Li}{Zhang et~al\mbox{.}}{2014}]%
        {zhang2014lore}
\bibfield{author}{\bibinfo{person}{Jia-Dong Zhang}, \bibinfo{person}{Chi-Yin Chow}, {and} \bibinfo{person}{Yanhua Li}.} \bibinfo{year}{2014}\natexlab{}.
\newblock \showarticletitle{Lore: Exploiting sequential influence for location recommendations}. \bibinfo{publisher}{ACM SIGSPATIAL International Conference on Advances in Geographic Information Systems}.
\newblock


\bibitem[\protect\citeauthoryear{Zhang, Amiri, Liu, Z{\"u}fle, and Zhao}{Zhang et~al\mbox{.}}{2023}]%
        {llm_traj_liangzhao}
\bibfield{author}{\bibinfo{person}{Zheng Zhang}, \bibinfo{person}{Hossein Amiri}, \bibinfo{person}{Zhenke Liu}, \bibinfo{person}{Andreas Z{\"u}fle}, {and} \bibinfo{person}{Liang Zhao}.} \bibinfo{year}{2023}\natexlab{}.
\newblock \showarticletitle{Large Language Models for Spatial Trajectory Patterns Mining}.
\newblock  (\bibinfo{year}{2023}).
\newblock


\bibitem[\protect\citeauthoryear{Zhao, Zhao, Yang, Lyu, and King}{Zhao et~al\mbox{.}}{2016}]%
        {zhao2016stellar}
\bibfield{author}{\bibinfo{person}{Shenglin Zhao}, \bibinfo{person}{Tong Zhao}, \bibinfo{person}{Haiqin Yang}, \bibinfo{person}{Michael Lyu}, {and} \bibinfo{person}{Irwin King}.} \bibinfo{year}{2016}\natexlab{}.
\newblock \showarticletitle{STELLAR: Spatial-temporal latent ranking for successive point-of-interest recommendation}. \bibinfo{publisher}{AAAI Conference on Artificial Intelligence}.
\newblock


\end{thebibliography}

\appendix
\newpage
\begin{table*}[h]
  \renewcommand{\arraystretch}{2}
  \fontsize{8}{8}\selectfont
  \centering
  \caption{A summary of the threat model.}
  \begin{tabular}{ccc}
    \toprule
    \textbf{Attack} & \textbf{Adversary Objective} & \textbf{Adversary Knowledge}\\
    \midrule
    \textsc{LocExtract} & \parbox{6cm}{Extract the most frequently visited location \(l\) of a target user \(u\)} & -- \\
    \midrule
    \textsc{TrajExtract} & \parbox{6cm}{Extract the location sequence of a target user \(u\) with length \(n\): \(x_L=\{l_0, \dots, l_{n-1}\}\)} & Starting location \(l_0\) \\
    \midrule
    \textsc{LocMIA} & \parbox{6cm}{Infer the membership of a user-location pair~(\(u\),\(l\))} & Shadow dataset \(D_{\mathrm{s}}\) \\
    \midrule
    \textsc{TrajMIA} & \parbox{6cm}{Infer the membership of a trajectory sequence \(x_T=\{(l_0,t_0),\dots,(l_n,t_{n})\}\)} & Shadow dataset \(D_{\mathrm{s}}\) \\
    \bottomrule
  \end{tabular}
  \label{tab:attobj}
\end{table*}

\section{Attack Algorithms}
\label{app:attack}
\noindent
\textbf{\textsc{TrajExtract}} To start with, we initialize $\beta$ candidate trajectories with the same starting location $l$ and the query time $t$ given a user (lines 1-3). Next, we iteratively extend the candidate trajectories: to extract the $i$-th ($i \geq 1$) locations of $\hat{x}_T^{0:i-1}$ in $\beta$ candidate trajectories, we query the model using $\hat{x}_T^{0:i-1}$ and compute the log perplexity for $\beta \times \mathcal{L}$ possible new trajectories with a length of $i+1$. We then choose the $\beta$ trajectories with the lowest log perplexity as the new candidate trajectories in the next iteration (line 6). The iterations end until the length of the candidate trajectories reaches $n$. Lastly, we take the location sequences from the final trajectories. Note that both \textsc{LocExtract} and \textsc{TrajExtract} need the timestamp $t$ to query the victim model, and we will show the effects of timestamp $t$ in our experiments.

\noindent
\textbf{LiRA}
The key idea of LiRA~\cite{carlini2022membership} is that models trained with or without $X_{target}$ (i.e., $f_{in}$ and $f_{out}$) would produce different loss distributions for $X_{target}$.
Specifically, LiRA consists of four main steps: 

(1) Train an equal number of shadow models \textit{with} and \textit{without} $X_{target}$ using the shadow dataset owned by the attacker to simulate the behavior of the black-box victim model.

(2) Use $X_{target}$ to query each shadow model to obtain loss values. The obtained loss values will be used to calculate two loss distributions, depending on whether the queried shadow model was trained on $X_{target}$.

(3) Query the victim model with $X_{target}$ to obtain the output logits.

(4) Conduct a $\Lambda$ hypothesis test to infer the membership of the $X_{target}$. The output score $\Lambda$ indicates whether the output from the victim model is closer to one of the two loss distributions.

LiRA originally trained 2$N$ shadow models for each target example. However, this approach suffers from computational inefficiency when the number of target examples is large. To address this issue, we employ the parallelized approach described in~\citep{carlini2022membership}, which reuses the same set of $2N$ shadow models for inferring the membership of multiple $X_{target}$. 
Our attacks further extend the key concept of LiRA and make it feasible for POI recommendation models by incorporating spatial-temporal information through unique designs, such as our spatial-temporal model query algorithm.

\begin{algorithm}[h]
\caption{Common Location Extraction Attack} %
\label{alg:common_point}
\begin{algorithmic}[1]
\Require Victim model: $f_{\theta}$, target user: $u$, query budget: $q$, query timestamp: $t$, output size: $k$
\Ensure Top-$k$ predictions: $[\hat{l}_{top1}$,\dots,$\hat{l}_{topk}]$
\State $\mathrm{logits} \leftarrow$ \{\}
\For{$q$ times}
    \State $l \leftarrow$ $\textsc{RandomSample}$($\mathcal{L}$) \Comment{Randomly generate a location from the location space}
    \State $\mathrm{logits} \cup f_{\theta} \big(u, \{(l, t)\} \big)$ 
\EndFor
\State $\mathrm{logits_{agg}} = \textsc{Aggregate}(\mathrm{logits})$ \Comment{Aggregate confidence for all locations}
\State \Return $\hat{l}_{top1}$,\dots,$\hat{l}_{topk} \gets \textsc{Argmax}_k(\mathrm{logits_{agg}})$
\end{algorithmic}
\end{algorithm}

\begin{algorithm}[h]
\caption{Training Trajectory Extraction Attack}%
\label{alg:TrainingTraj}
\begin{algorithmic}[1]
\Require Victim model: $f_\theta$, target user: $u$, starting location: $l_0$, target extraction length: $n$, query timestamp: $t$, beam width:~$\beta$%
\Ensure Top-$\beta$ possible extraction results: $\hat{x}_{L_0}^{0:n},\dots,\hat{x}_{L_\beta}^{0:n}$
\For{$b \leftarrow 0 \text{ to } \beta - 1$}
    \State $\hat{x}_{T_b}^{0:0} \gets (u,(l_0,t))$ \Comment{Initialize the beam with $l_0$ and $t$}
\EndFor
\For{$i \leftarrow 1 \text{ to } n-1$}
    \For {$\hat{x}_T^{0:i-1}$ $\text{ in } \{\hat{x}_{T_0}^{0:i-1},\dots,\hat{x}_{T_\beta}^{0:i-1}$\}}
        \State $\{\hat{x}_{T_0}^{0:i},\dots,\hat{x}_{T_\beta}^{0:i}\}\gets~\textsc{UpdateBeam}_\beta (f_{\theta}(u,\hat{x}_T^{0:i-1}))$ \Comment{Update the beam by keeping $\beta$ trajectory with the smallest PPL from the query output and current beam}
    \EndFor
\EndFor
\State $\hat{x}_{L_0}^{0:n-1},\dots,\hat{x}_{L_\beta}^{0:n-1} \gets \textsc{Getloc} (\hat{x}_{T_0}^{0:n-1},\dots,\hat{x}_{T_\beta}^{0:n-1})$ \Comment{Take the location sequence from $\hat{x}_T^{0:n-1}$ as result $\hat{x}_L^{0:n-1}$}
\State \Return $\hat{x}_{L_0}^{0:n-1},\dots,\hat{x}_{L_\beta}^{0:n-1}$
\end{algorithmic}
\end{algorithm}

\begin{algorithm}[h]
\caption{\textsc{SpaTemQuery}: Spatial-Temporal Model Query Algorithm for \textsc{LocMIA}}
\label{alg:LocQuery}
\begin{algorithmic}[1]
\Require Target model: $f_{target}$, number of query timestamps: $n_t$, number of query locations: $n_l$, target example: $X_{target}$
\Ensure Membership confidence score: $conf$
\State $u,l \gets X_{target}$
\State $conf_{all} \gets \{\}$
\For{$i \gets 0$ to $n_t-1$}
    \State $conf_{t} \gets \{\}$
    \For{$j \gets 0$ to $n_l-1$}
        \State $t_i \gets i/n_t$\;
        \State $l_j \gets \textsc{RandomSample}(\mathcal{L})$\;
        \State $conf_{t} \gets conf_{t} \cup f_{target}(u,(l_j, t_i))$ \Comment{Query the model with random location and a synthetic timestamp}
    \EndFor
    \State $conf_{all} \gets conf_{all} \cup \text{mean}(conf_{t})$ \Comment{Calculate average confidence from all queries for this timestamp}
\EndFor
\State \Return $conf \gets \text{max}(conf_{all})$ \Comment{Take the confidence scores with largest confidence at position $l$ as output}
\end{algorithmic}
\end{algorithm}

\begin{algorithm}[htbp] 
\caption{Membership Inference Attack} 
\label{alg:lira_loc}
Below, we demonstrate our location-level MIA and trajectory-level MIA algorithms. The lines marked in \textcolor{red}{red} are specific to \textsc{LocMIA}, while the lines marked in \textcolor{purple}{purple} are specific to \textsc{TrajMIA}. Both attacks share the remaining lines.
\begin{algorithmic}[1]
\Require Victim model: $f_\theta$, shadow data: $\dshadow$, number of shadow models: $N$, inference target: $X_{target}$, \textcolor{red}{number of query timestamps: $n_t$, number of query locations: $n_l$} %
\Ensure The likelihood ratio to determine if we should reject the hypothesis that $X_{target}$ is a member of $f_\theta$: $\Lambda$
\State $conf_{in},conf_{out}\leftarrow $ \{\},\{\}
\State \textcolor{red}{$X_S \gets$ \textsc{RandomSample}($\{X_{S}: X_{target} \in X_{S}\}$)} \Comment{Sample a location sequence and includes $X_{target}$}
\State \textcolor{purple}{$X_S \gets$ $X_{target}$} 

\For{$i \gets 0$ to $N$}
    \State $D_{in} \gets \textsc{RandomSample}(\dshadow) \cup X_S$ 
    \State $D_{out} \gets \textsc{RandomSample}(\dshadow)$ \textbackslash~$X_S$ 
    \State $f_{in}, f_{out} \gets$ \textsc{Train}($D_{in}$), \textsc{Train}($D_{out}$) \Comment{Train $f_{in}$ and $f_{out}$}
    \State \textcolor{red}{$conf_{in}~\gets~conf_{in}~\cup~\phi(\textsc{SpaTemQuery}(f_{in},n_t,n_l,X_{target}))$}
    \State \textcolor{red}{$conf_{out}~\gets~conf_{out}~\cup$\\ \hspace{19mm}$\phi(\textsc{SpaTemQuery}(f_{out},n_t,n_l,X_{target}))$} 
    \State \textcolor{purple}{$conf_{in}\gets conf_{in} \cup $\\ \hspace{19mm}$\phi\big(\text{mean}(\{f_{in}(X_S)^{0:0}, \dots, f_{in}(X_S)^{0:n-1}\})\big)$}
    \State \textcolor{purple}{$conf_{out}\gets conf_{out}~\cup $\\ \hspace{19mm}$ \phi\big(\text{mean}(\{f_{out}(X_S)^{0:0}, \dots, f_{out}(X_S)^{0:n-1}\})\big)$}
\EndFor
\State $\mu_{\text{in}}, \mu_{\text{out}}\leftarrow$ mean($conf_{\text{in}}$), mean($conf_{\text{out}}$)
\State $\sigma_{\text{in}}^2,\sigma_{\text{out}}^2 \leftarrow$ var($conf_{\text{in}}$),var($conf_{\text{out}}$)
\State \textcolor{red}{$conf_{\text{obs}} \leftarrow \phi(\textsc{SpaTemQuery}(f_\theta,n_t,n_l,X_{target}))$} 
\State \textcolor{purple}{$conf_{\text{obs}}\gets \phi\big(\text{mean}(\{f_\theta(X_S)^{0:0}, \dots, f_\theta(X_S)^{0:n-1}\})\big)$}
\State\Return $\Lambda = \frac{p(conf_{\text{obs}} | \mathcal{N}(\mu_{\text{in}},\sigma_{\text{in}}^2))}{p(conf_{\text{obs}} | \mathcal{N}(\mu_{\text{out}},\sigma_{\text{out}}^2))}$  \Comment{Hypothesis test}
\end{algorithmic}
\end{algorithm}

\label{app:detail_exp}
{\tiny
\begin{table}[!ht]
\renewcommand{\arraystretch}{1.2}
\fontsize{8}{8}\selectfont
\centering
\caption{Statistics of POI Recommendation Datasets.}
\begin{tabular}{cccccc}
\toprule
 & \#POIs & \#Check-ins & \#Users & \#Trajectories & Avg. Len. \\\midrule
\textsc{4sq}     & 4,556  & 63,648 & 1,070 &17,700 &3.63 \\
\textsc{Gowalla} & 2,559  & 32,633 & 1,419 &7,256 &4.46 \\
\bottomrule
\end{tabular}
\label{data_feature}
\end{table}
}

\section{Detailed Experimental Setup}

\subsection{Datasets}We conduct experiments on two POI recommendation benchmark datasets -- FourSquare (\textsc{4sq})~\citep{yang2014modeling} and \textsc{Gowalla}~\citep{cho2011friendship} datasets. 
Following the literature~\citep{10.1145/3477495.3531983,kong2018hst}, we use the check-ins collected in NYC for both sources.
The \textsc{4sq} dataset consists of 76,481 check-ins during ten months (from April 12, 2012, to February 16, 2013). The \textsc{Gowalla} dataset comprises 35,674 check-ins collected over a duration of 20 months (from February 2009 to October 2010). In both datasets, a check-in record can be represented as [user ID, check-in time, latitude, longitude, location ID].

\subsection{Data Preprocessing}
We preprocess each dataset following the literature~\citep{10.1145/3477495.3531983}: (1) We first filter out unpopular POIs and users that appear less than ten times to reduce noises introduced by uncommon check-ins. (2) To construct trajectories of different users in a daily manner, the entire check-in sequence of each user is divided into trajectories with 24-hour intervals. Then, we filter out the trajectories with only a single check-in. (3) We further normalize the timestamp (from 0:00 AM to 11:59 PM) in each check-in record into $[0, 1]$. After the aforementioned steps, the key statistics of the \textsc{4sq} and \textsc{Gowalla} datasets are shown in Table~\ref{data_feature}. (4) Lastly, we split the datasets into the training, validation, and test sets using the ratio of 8:1:1.

\noindent
\textbf{Victim Model Training Settings:}~~ 
We use the official implementation of \textsc{GETNext}\footnote{\texttt{https://github.com/songyangme/GETNext}} and \textsc{LSTPM}\footnote{\texttt{https://github.com/NLPWM-WHU/LSTPM}} to train victim models. In particular, we train each model with a batch size 32 for 200 epochs by default. We use five random seeds in all experiments and report the average results.

\subsection{Attack Settings}

\noindent
\textbf{\textsc{LocExtract}}~~
Given a target user $u$, we extract the most visited location $l_{top1}$ from the victim model $f_{\theta}$ with a query number $q=50$. 
We set the query timestamp $t=0.5$ (i.e., the middle of the day) by default, and we will present how the change of the query timestamp affects the attack performance in the ablation study. 

\noindent
\textbf{\textsc{TrajExtract}}~~
In this attack, we experiment with $n=4$ by default, though the attacker can potentially extract location (sub-)sequences with arbitrary length. We set the beam size $\beta=50$ in the beam search to query the victim model and update candidate trajectories. For each query, we also have the default query timestamp $t=0.5$. 

\noindent
\textbf{\textsc{LocMIA}}~~
In our experiments, since we randomly sample 80\% of trajectories as the training dataset $D_{tr}$ to build a victim model for MIA, we treat the remaining 20\% data as non-members. 
For each target user $u$ and the POI location $l$ pair, we generate $N=64$ synthesis trajectories using $\textsc{TrajSynthesis}$ with the query timestamp $t_s=0.5$. With the synthesis trajectories, we can also have 64 in-models ($f_{in}$) and 64 out-models ($f_{out}$). We also set $n_t=10$ and $n_l=10$.
For evaluation, we conduct a hypothesis test on a balanced number of members and non-members.

\noindent
\textbf{\textsc{TrajMIA}}~~
We extract the membership information of some trajectory sequences with arbitrary lengths from the victim model. 
We also build $N=64$ in-models ($f_{in}$) and $N=64$ out-models ($f_{out}$) for a target trajectory sequence. For evaluation, we conduct a hypothesis test on a balanced number of members and non-members.

\section{More Details of Analyzing Factors in Mobility Data that Make it Vulnerable to the Attacks}
\label{ana:user_ana}

\subsection{How to Select Aggregate Statistics}
\label{ana:featureselect}

This section outlines the basic principles and details for selecting representative aggregate statistics for analysis. For user-level aggregate statistics, we target the basic statistical information quantifying properties of locations and trajectories of a user. For location-level and trajectory-level aggregate statistics, we study their users, ``neighboring'' check-ins and trajectories, and the check-in time information. In summary, we select the following aggregate statistics:

\begin{itemize}
    \item User-level aggregate statistics:
    \begin{enumerate}
        \item Total number of check-ins;
        \item Number of unique visited POIs;
        \item Number of trajectories;
        \item Average trajectory length;
    \end{enumerate}
    \item Location-level aggregate statistics:
    \begin{enumerate}
        \item Number of users who have visited this POI;
        \item Number of check-ins surrounding ($\leq$ 1km) this POI;
        \item Number of trajectories sharing this POI;
        \item Average time in a day for the visits to
the POI;
    \end{enumerate}
    \item Trajectory-level aggregate statistics:
    \begin{enumerate}
        \item Number of users who have the same trajectories;
        \item Number of check-ins surrounding ($\leq$ 1km) all POI in the trajectory;
        \item Number of intercepting trajectories;
        \item Average check-in time of the trajectory.
    \end{enumerate}
\end{itemize}

\begin{figure}[t]
  \centering
  \subfigure [\# Unique POIs (\textsc{4sq})]{\includegraphics[width=0.23\textwidth]{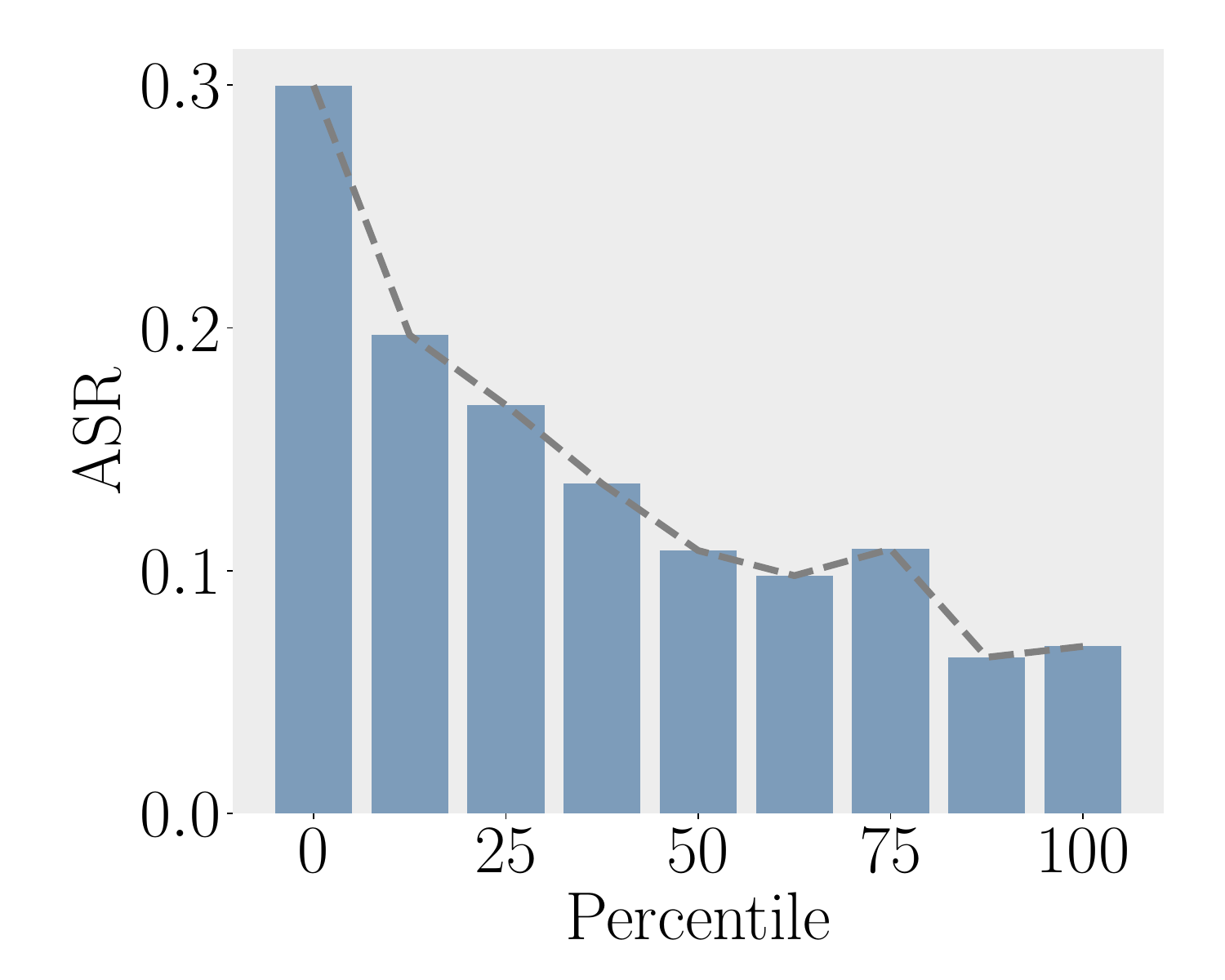}}
  \subfigure [\# Unique POIs (\textsc{Gowalla})]{\includegraphics[width=0.23\textwidth]{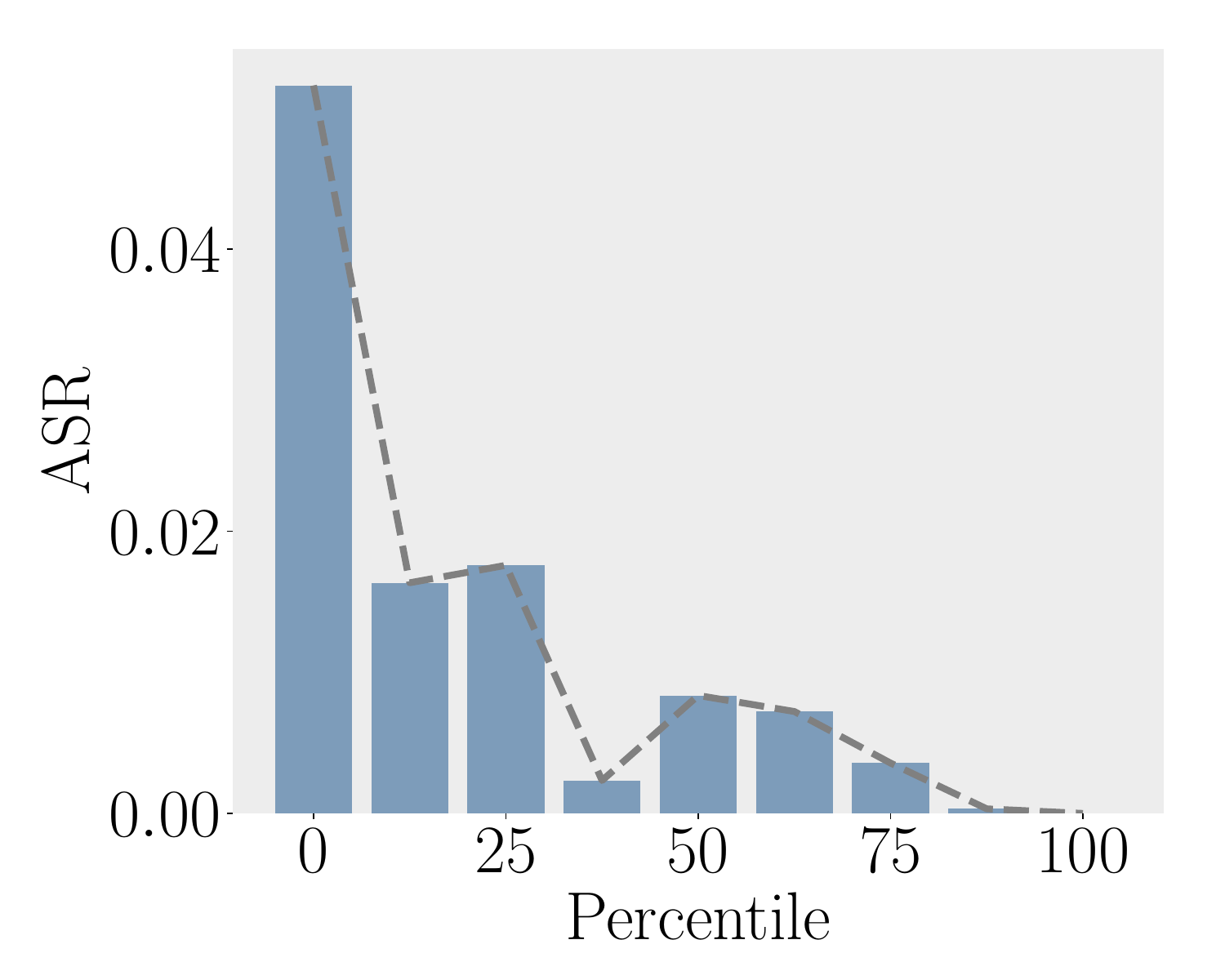}}
  \vspace{-2mm}
  \caption{How user-level aggregate statistics are related to \textsc{TrajExtract}. The users who have fewer unique POIs are more vulnerable to \textsc{TrajExtract}.}
  \label{fig:user_att2} 
\end{figure}

\begin{figure*}[t!]
    \subfigure [\# of User Check-ins]{\includegraphics[width=0.23\textwidth]{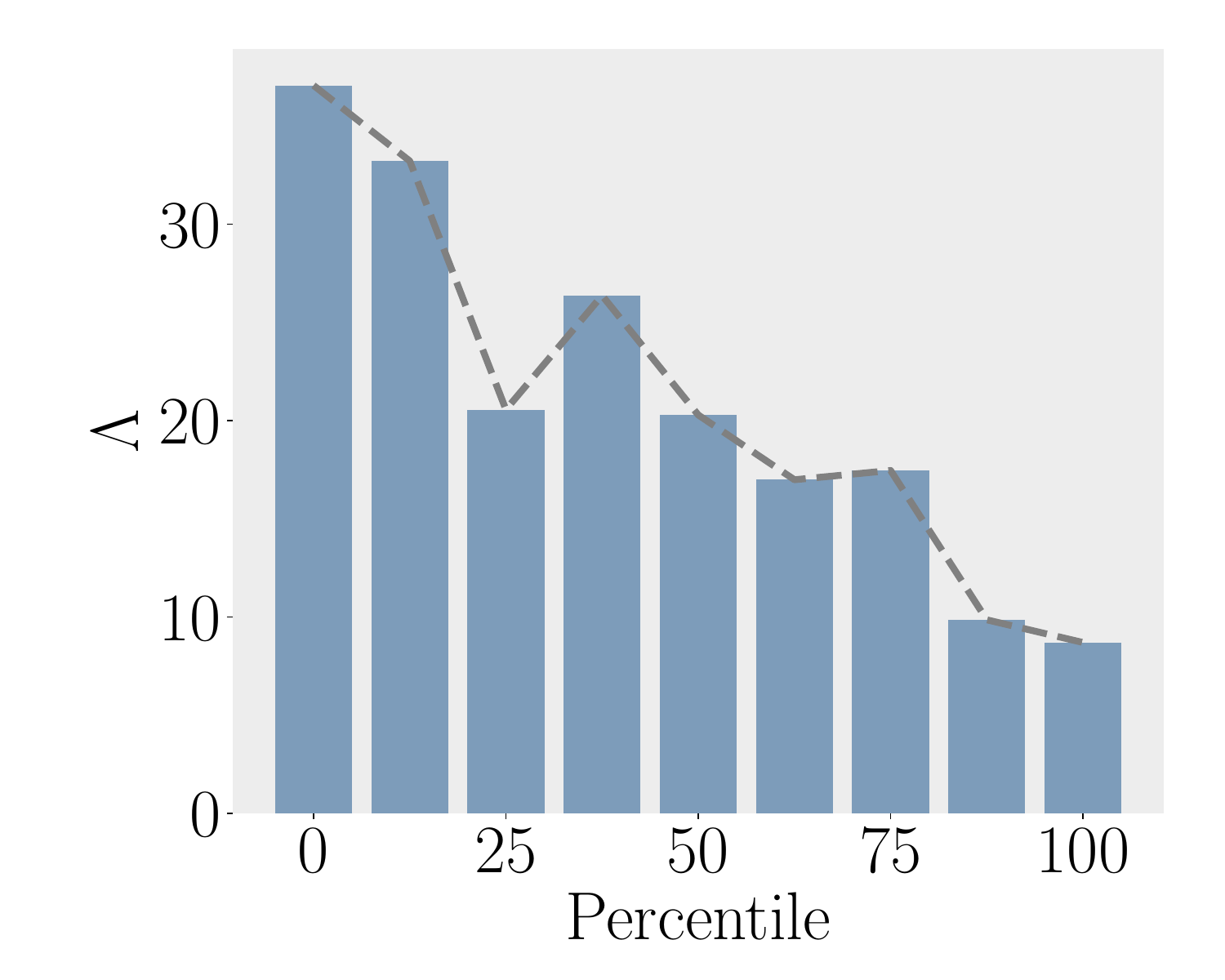}
    \label{fig:user_check_gow}}
    \subfigure [\# of User POIs]{\includegraphics[width=0.23\textwidth]{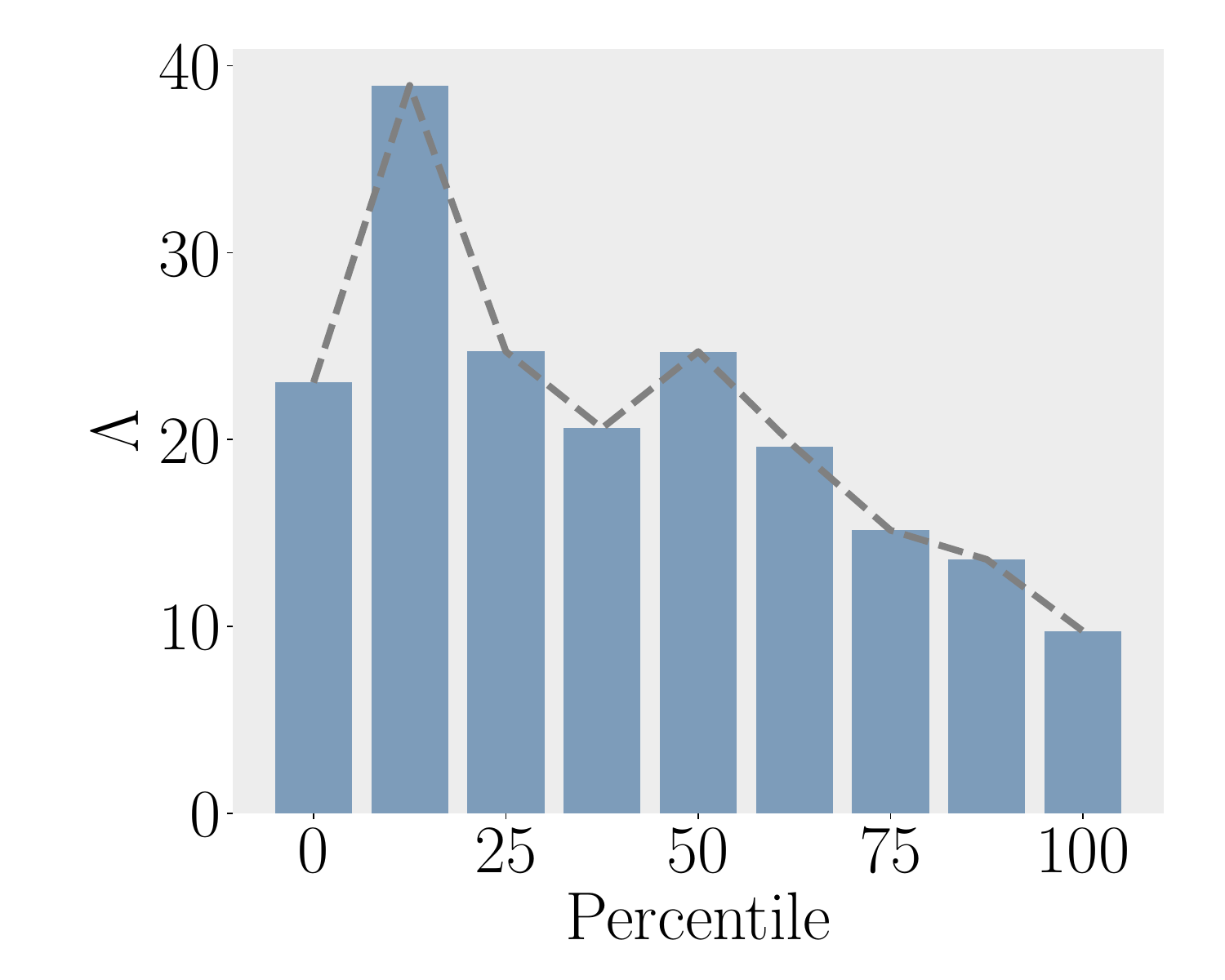}
     \label{fig:user_poi_gow}}
    \subfigure [\# of User Trajectories]{\includegraphics[width=0.23\textwidth]{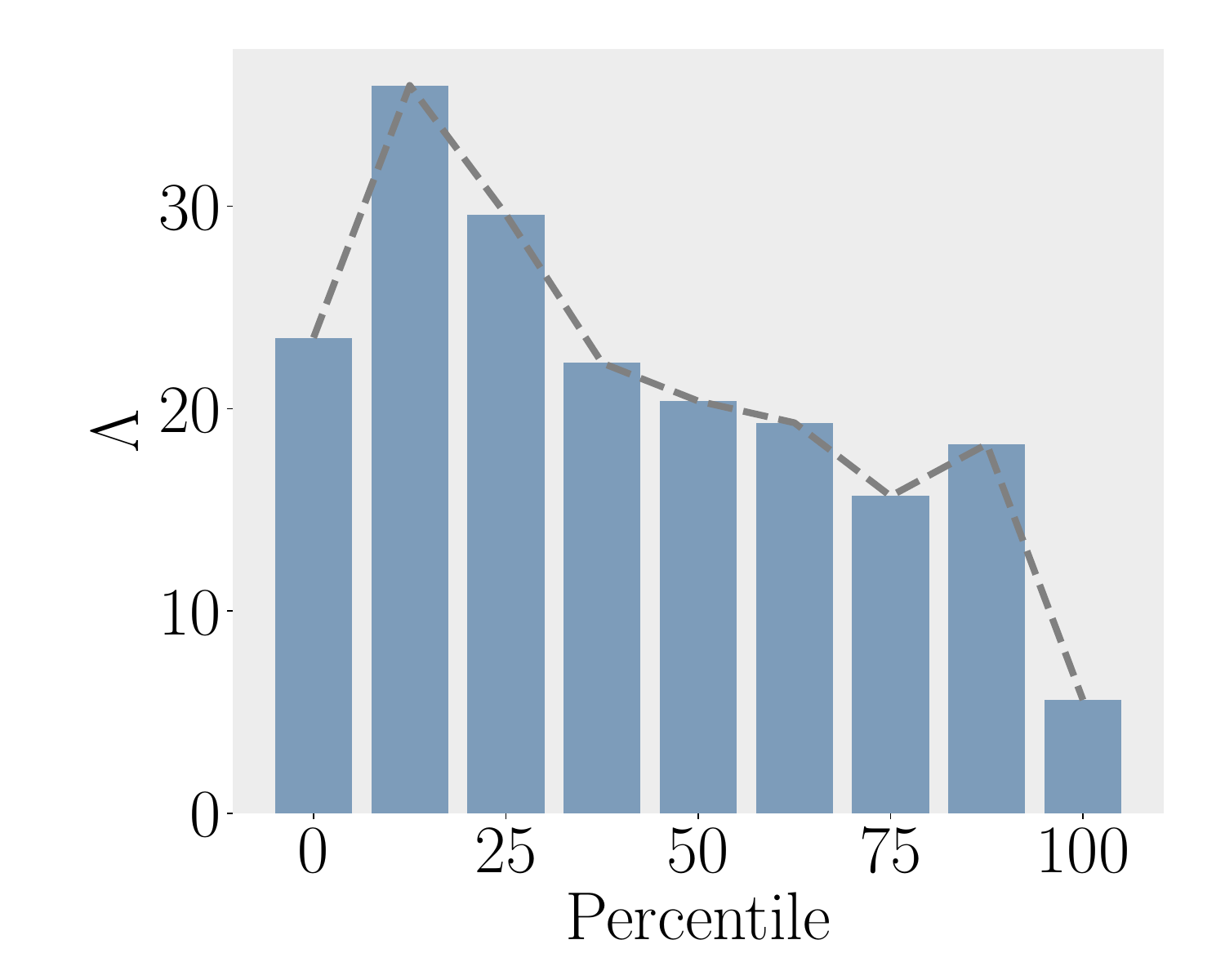}
    \label{fig:user_num_traj_gow}}
    \subfigure [Avg User Traj Length]{\includegraphics[width=0.23\textwidth]{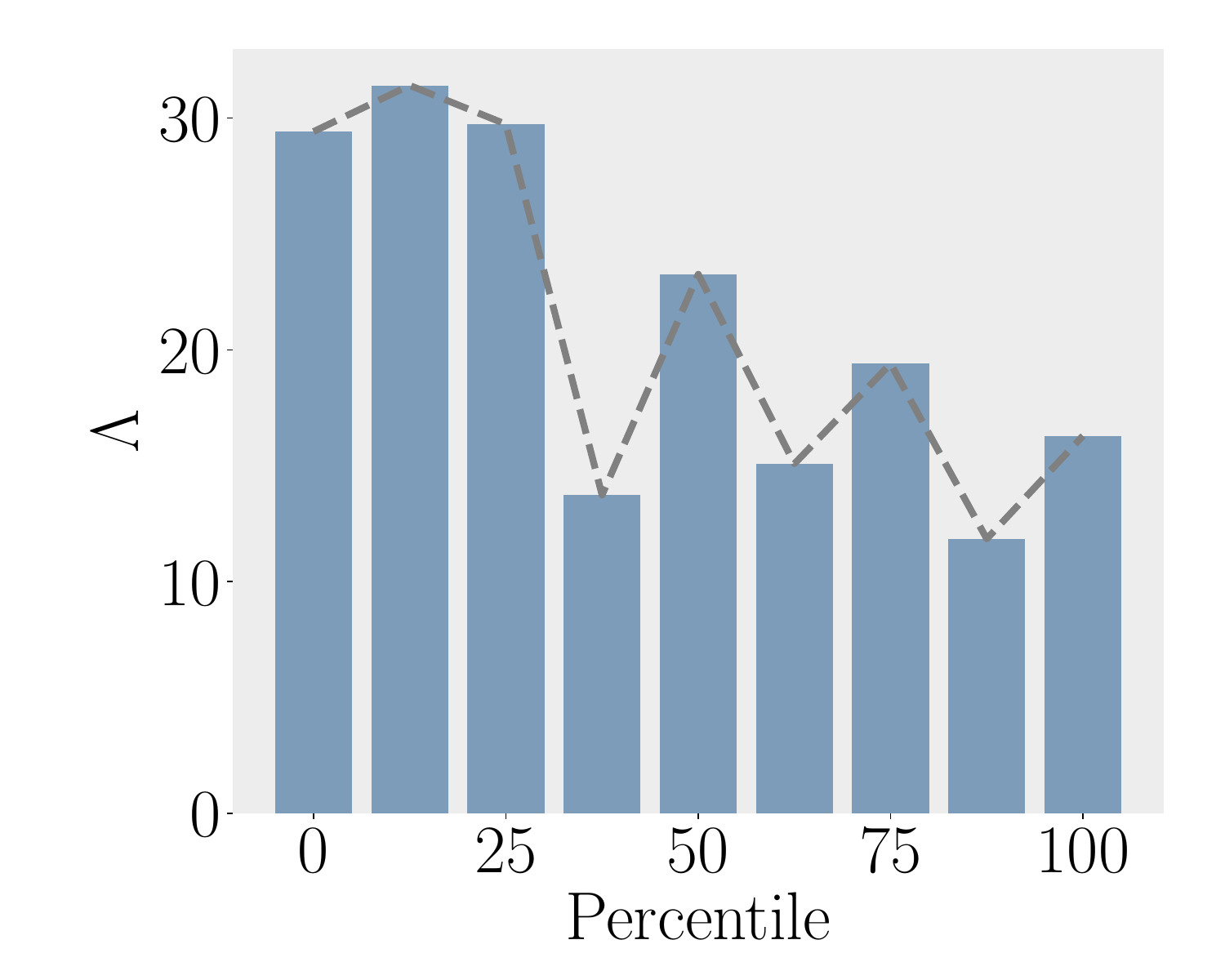}
    \label{fig:user_len_traj_gow}}
    \vspace{-2mm}
    \caption{How user-level aggregate statistics are related to \textsc{TrajMIA}. The users with fewer total check-ins, fewer unique POIs, and fewer or shorter trajectories are more vulnerable to \textsc{TrajMIA}. (\textsc{Gowalla})}
    \label{fig:user_gow}
\end{figure*}

\begin{figure}[t!]
    \begin{minipage}[t]{0.48\textwidth}
        \subfigure [\# of Users Visited] {\includegraphics[width=0.46\textwidth]
        {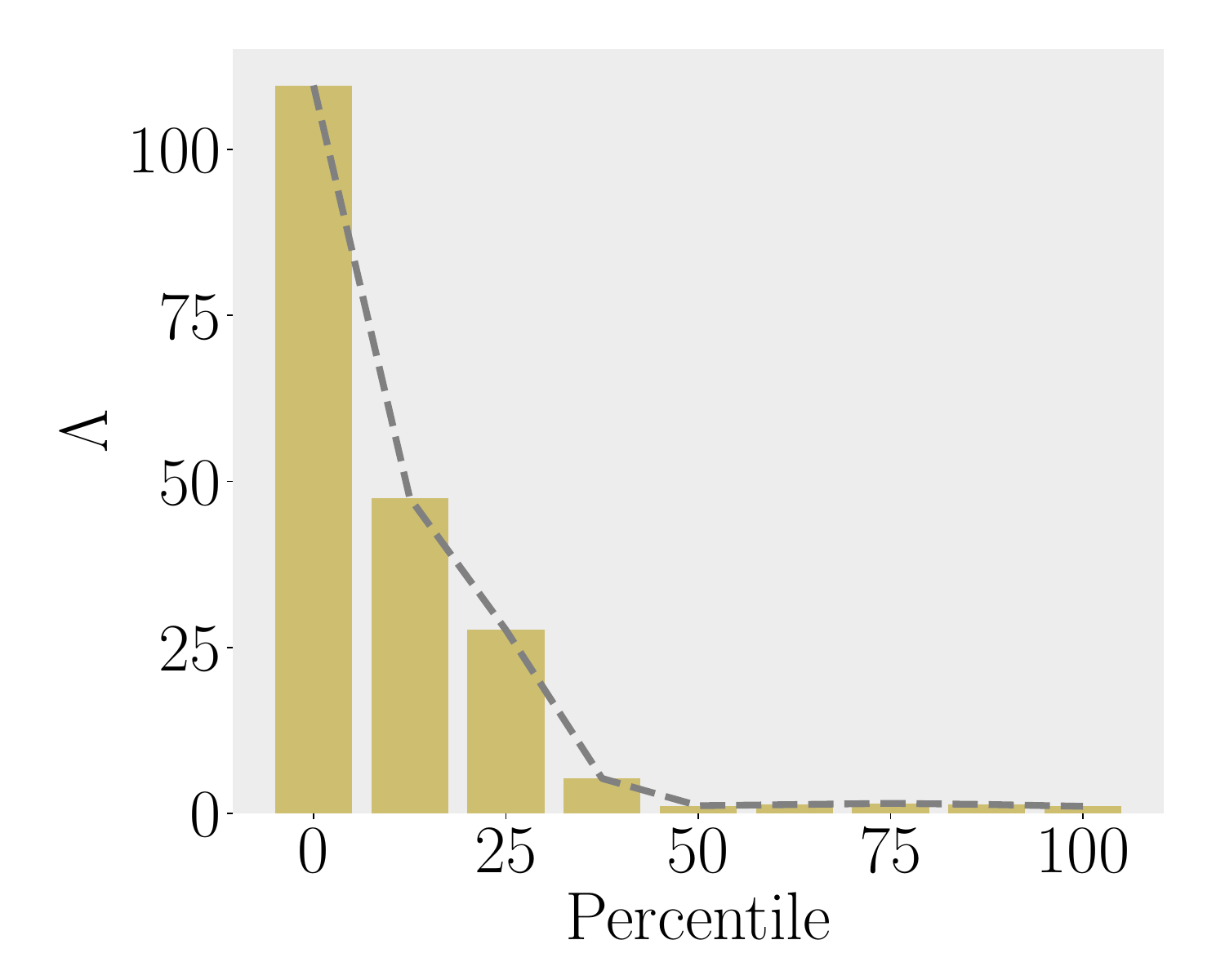}
        \label{fig:loc_num_user_gow}}
        \subfigure [\# of Nearby Check-ins] {\includegraphics[width=0.46\textwidth]
        {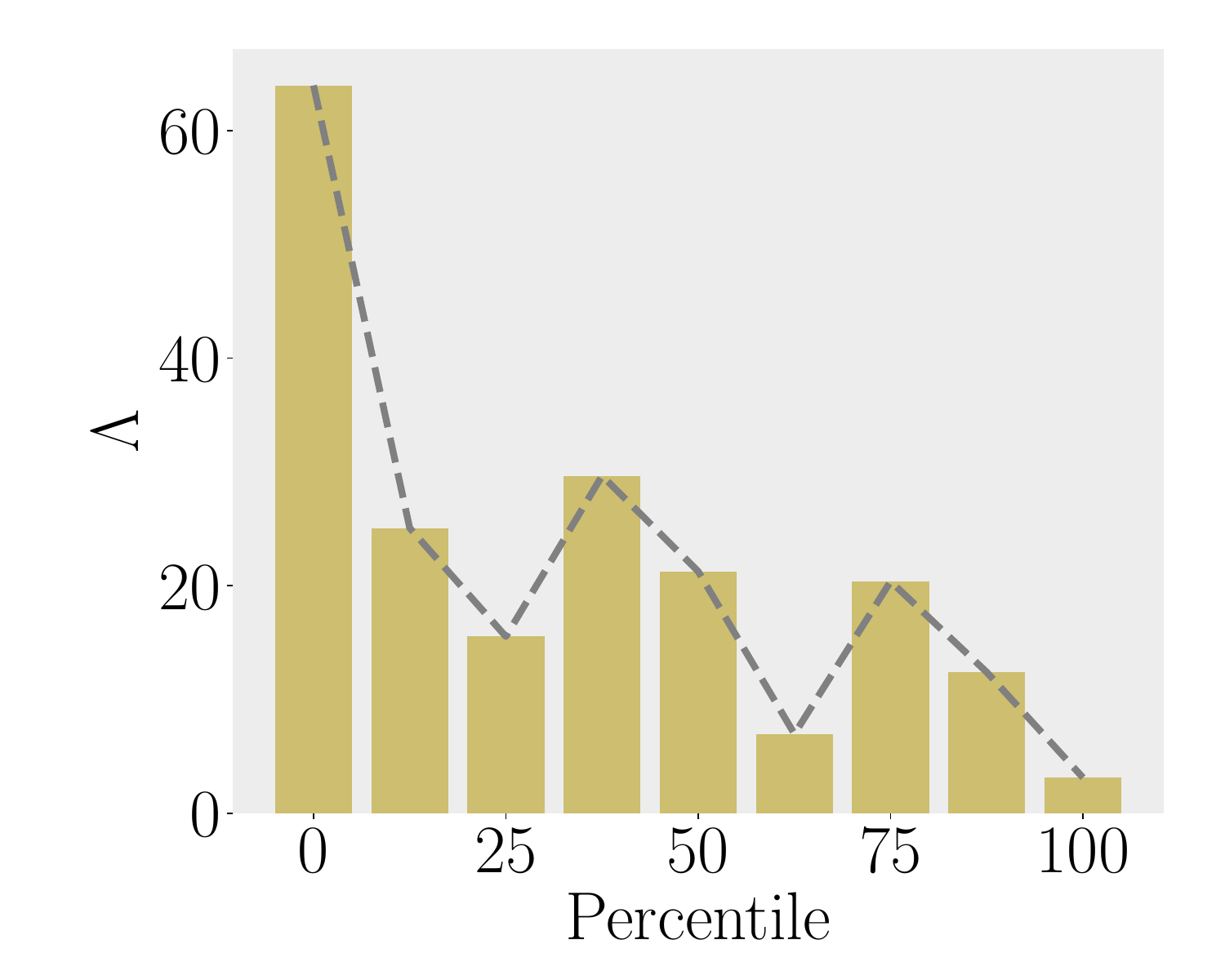}
        \label{fig:loc_num_loc_gow}}
        \vspace{-2mm}
        \caption{How location-level aggregate statistics are related to \textsc{LocMIA}. The locations shared by fewer users or have fewer surrounding check-ins are more vulnerable to \textsc{LocMIA}. (\textsc{Gowalla})}
    \end{minipage}
    \hfill
    \begin{minipage}[t]{0.48\textwidth}
    \subfigure [\# of Intercepting Trajs]{\includegraphics[width=0.46\textwidth]{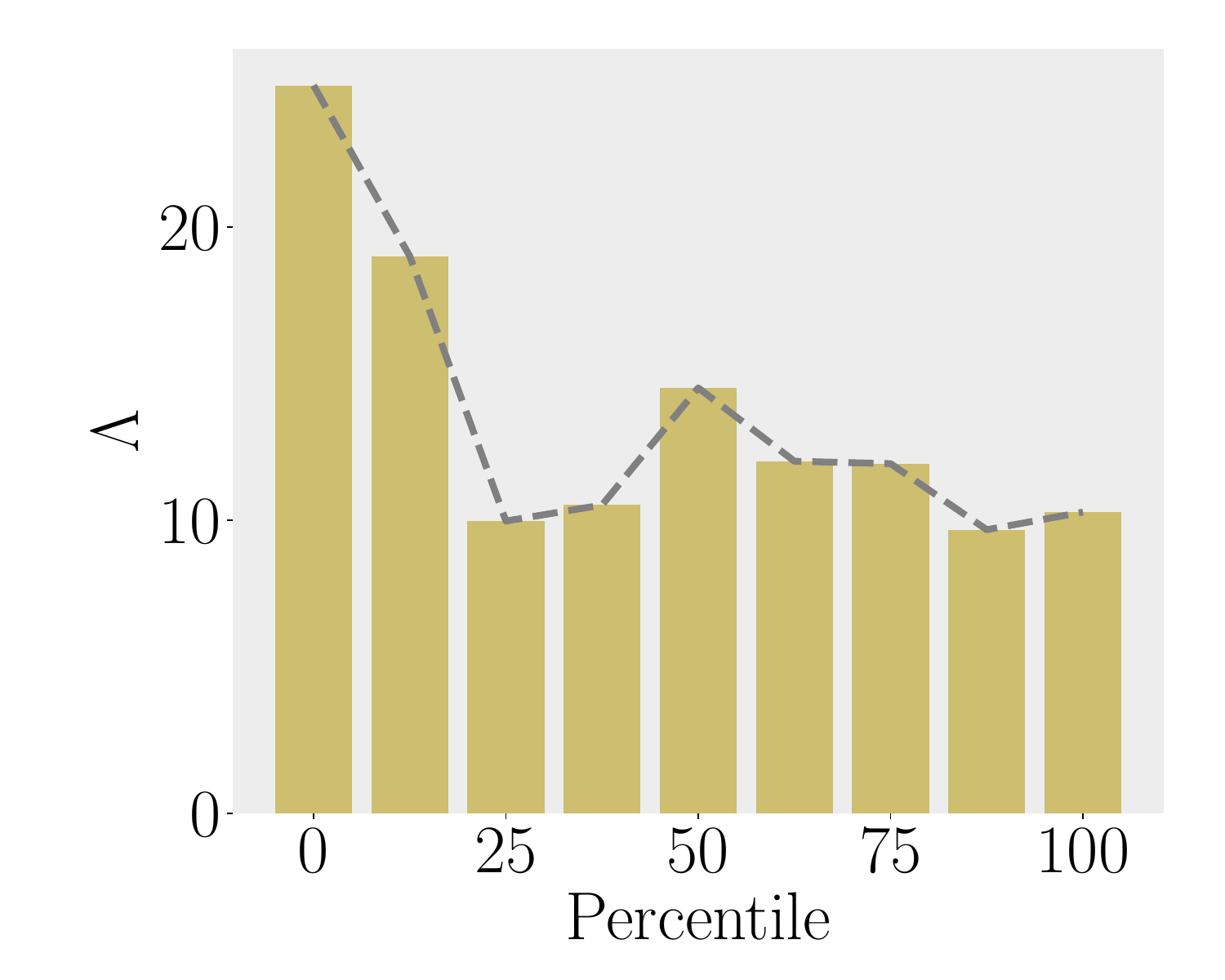}
    \label{ana:traj1}}
    \subfigure [\# of POIs in Traj ]{\includegraphics[width=0.46\textwidth]{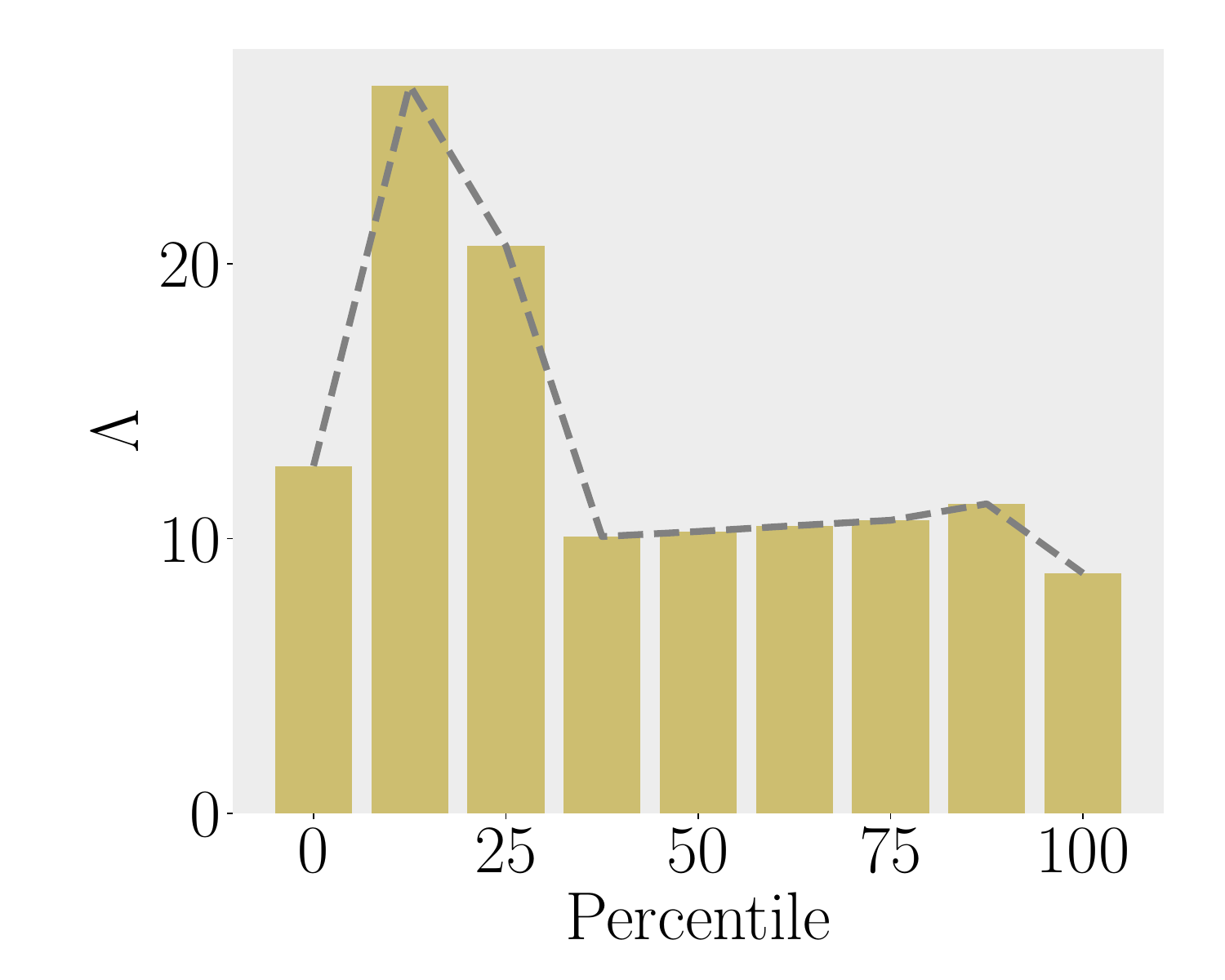}
    \label{ana:traj2}}
    \vspace{-2mm}
    \caption{How trajectory-level aggregate statistics are related to \textsc{TrajMIA}. The trajectories with fewer intercepting trajectories or fewer POIs in the trajectory are more vulnerable to \textsc{TrajMIA}. (\textsc{Gowalla})}
    \end{minipage}
    \label{fig:locana_gow}
\end{figure}

\section{The Impact of Training and Attack Parameters}
\label{result:parameter}

In this section, we analyze how training parameters (Sec.~\ref{Ablation:training}) and attack parameters affect the attack performance of data extraction attack (Sec.~\ref{Ablation:Extraction}) and membership inference attack (Sec.~\ref{Ablation:MIA}).

\subsection{The Impact of Training Parameters}
\label{Ablation:training}

Our primary objective of this analysis is to understand whether the occurrence of overfitting, commonly associated with excessive training epochs, leads to heightened information leakage based on our proposed attacks. By showcasing the ASR at various model training stages, we aim to gain insights into the relationship between overfitting and ASR.

\begin{figure*}[t!]
    \centering
    \subfigure [\textsc{LocExtract}] {\includegraphics[width=0.23\textwidth]{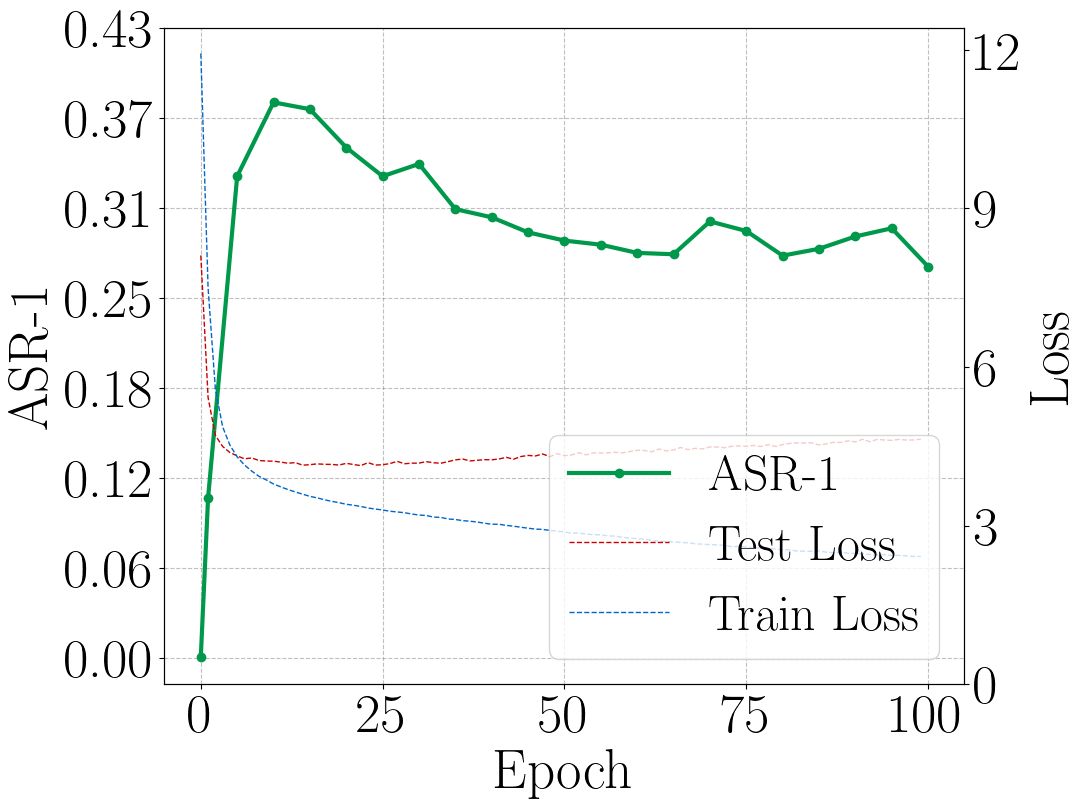}}
    \subfigure [\textsc{TrajExtract}] {\includegraphics[width=0.23\textwidth]{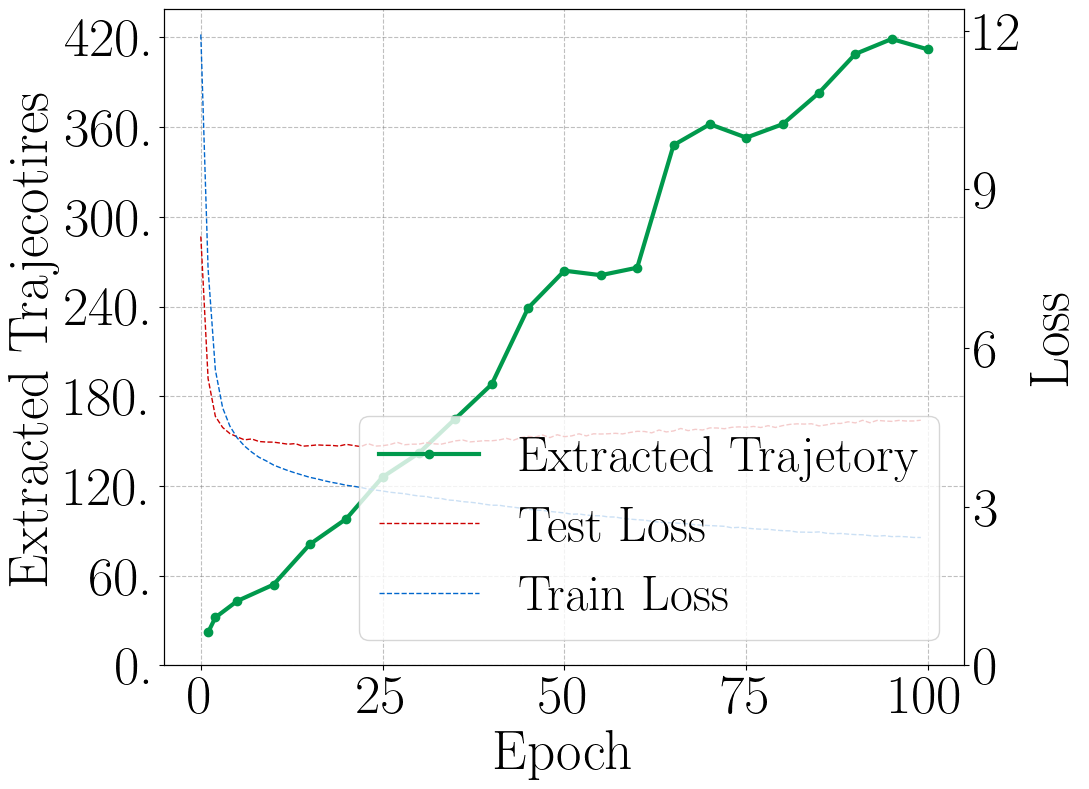}}
    \subfigure [\textsc{LocMIA}] {\includegraphics[width=0.23\textwidth]{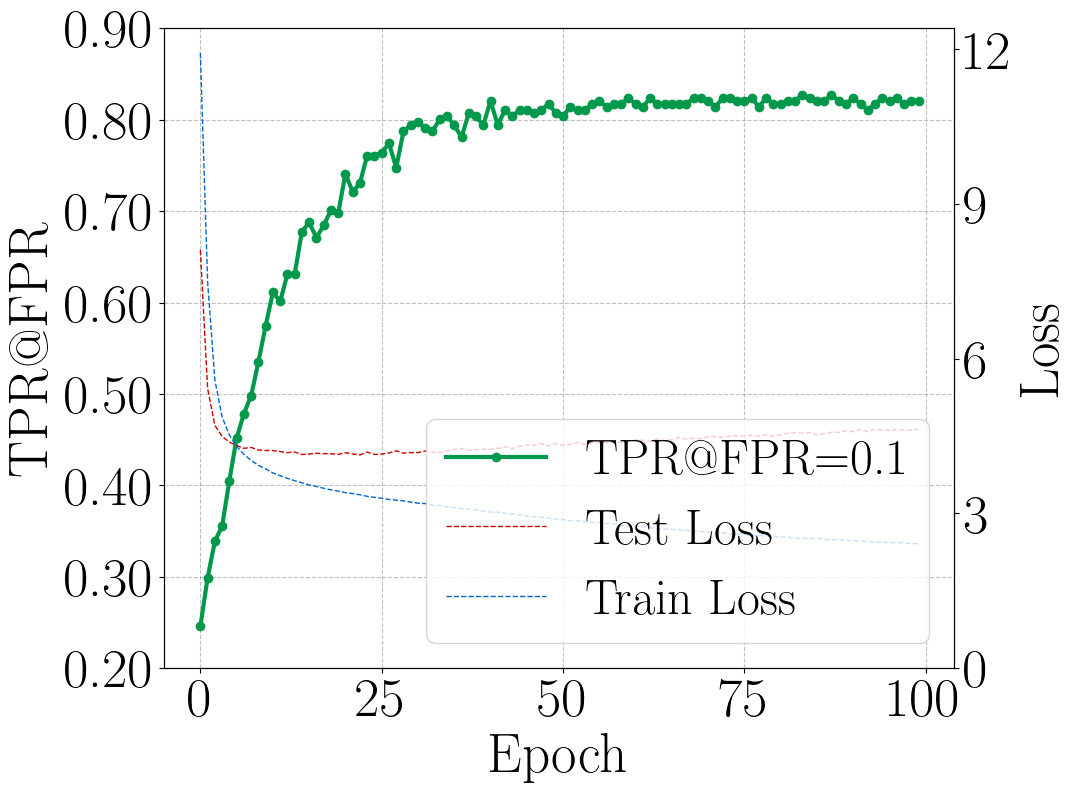}}
    \subfigure [\textsc{TrajMIA}] {\includegraphics[width=0.23\textwidth]{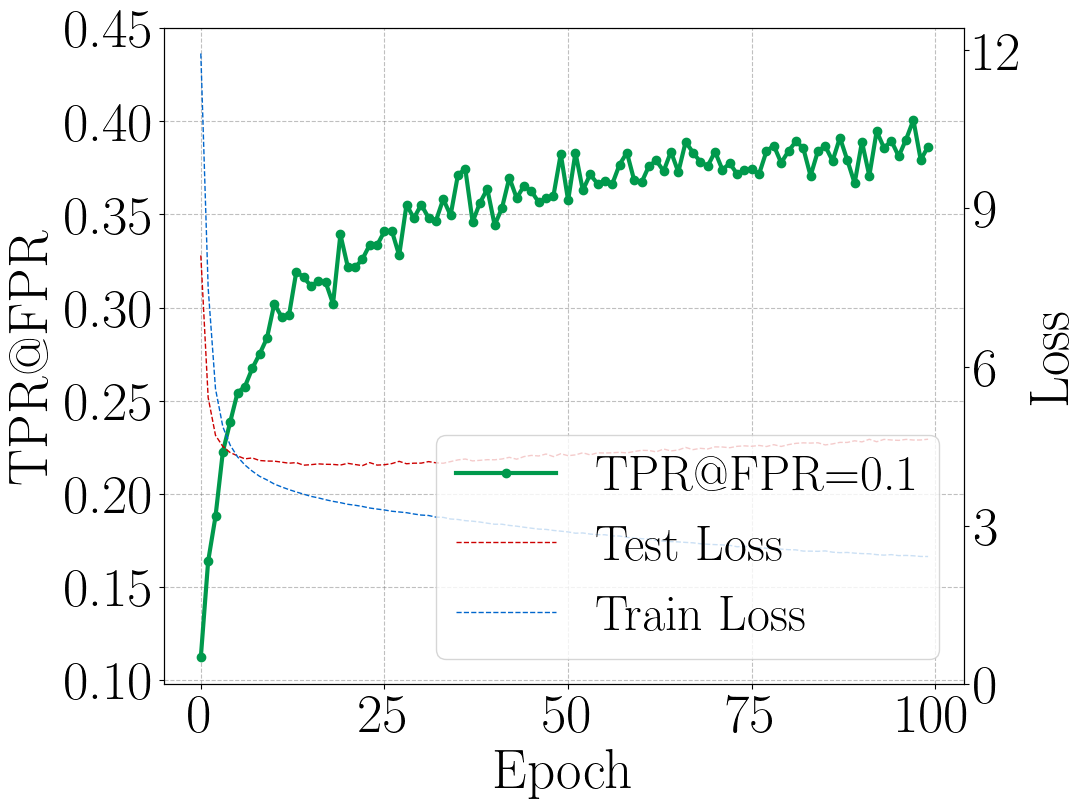}}
    \vspace{-2mm}
    \caption{
    The impact of the model generalization on the performance of four privacy attacks.}
    \label{fig:attackfitting}
\end{figure*}

Based on the results presented in Figure \ref{fig:attackfitting}, we observe that \textsc{LocExtract} achieves the best performance when the model is in the convergence stage. 
We speculate that continuing training beyond convergence leads to overfitting, causing a loss of generalization. Specifically, when the model is overfitted, it tends to assign higher confidence to the training data while disregarding the general rules present in the dataset. In contrast, our attack employs random queries to extract the general rules learned by the model from the dataset, resulting in better performance when applied to the optimally fitted model.

The attack performance of \textsc{TrajExtract} improves as the training process progresses, which can be attributed to the model becoming increasingly overfitted. The overfit model is more likely to output the exact training trajectory and generates more precise training trajectories than the best-fitted model when given the same number of queries.
Similarly, the results of our membership inference attacks reveal a trend of attack performance consistently improving with the progression of the training process. This observation aligns with our expectations, as when the model undergoes more training iterations, the effects of training data are more emphasized. Consequently, the distribution of query results in our attack on the seen training data diverging further from the distribution derived from the unseen data. This growing disparity between the two distributions facilitates the membership inference task, particularly on overfitted models. 
The analysis of these three attacks indicates a consistent trend, highlighting the increased risk of privacy leakage due to overfitting with respect to the original training data.

\subsection{Ablation Study on Our Data Extraction Attacks}
\label{Ablation:Extraction}

\begin{figure*}[t!]
    \centering
        \subfigure [\textsc{LocExtract} (\textsc{4sq})] {\includegraphics[width=0.22\textwidth]{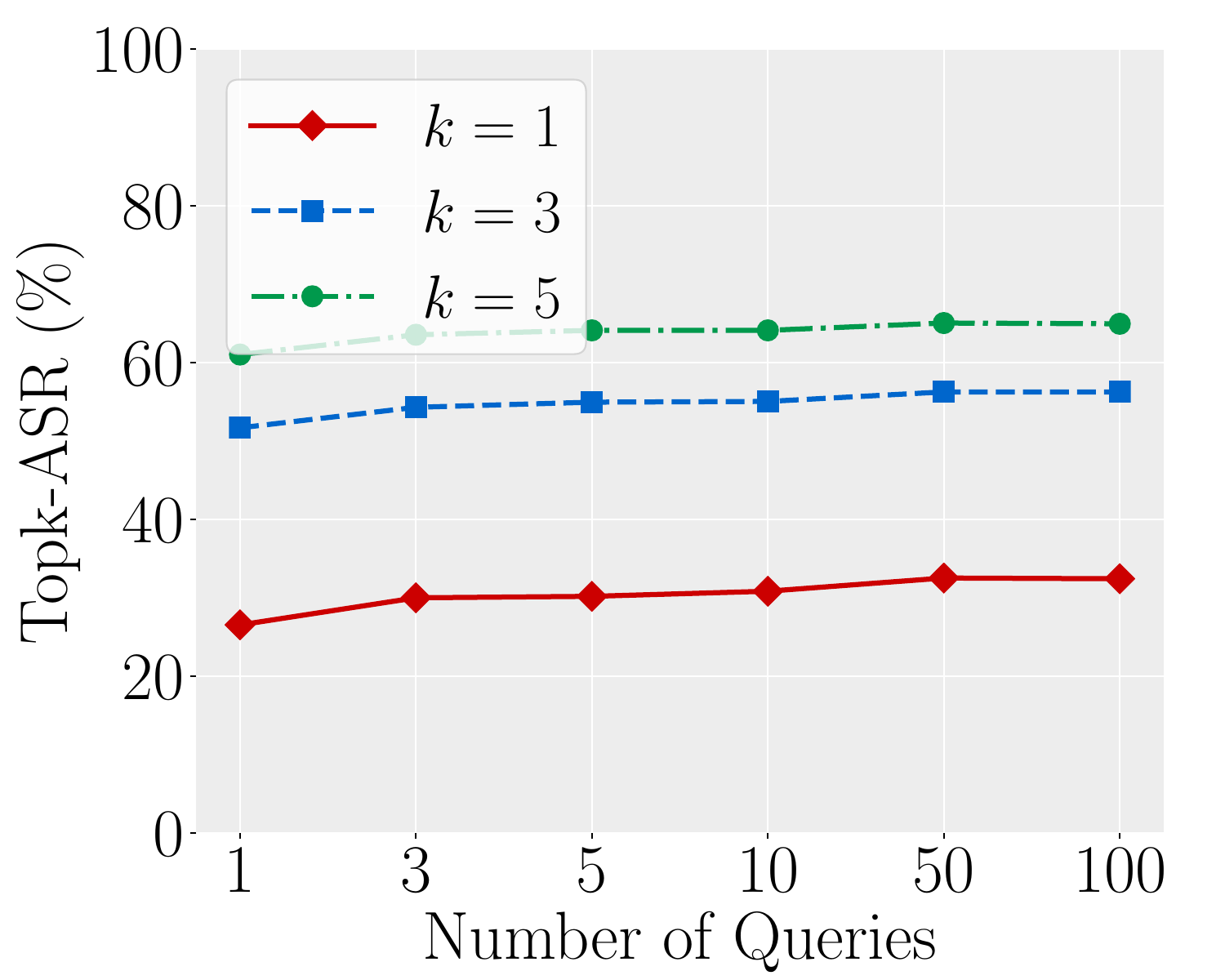}
        \label{fig:querylim_loc_extraction}
        }
        \subfigure [\textsc{TrajExtract} (\textsc{4sq})]{\includegraphics[width=0.22\textwidth]{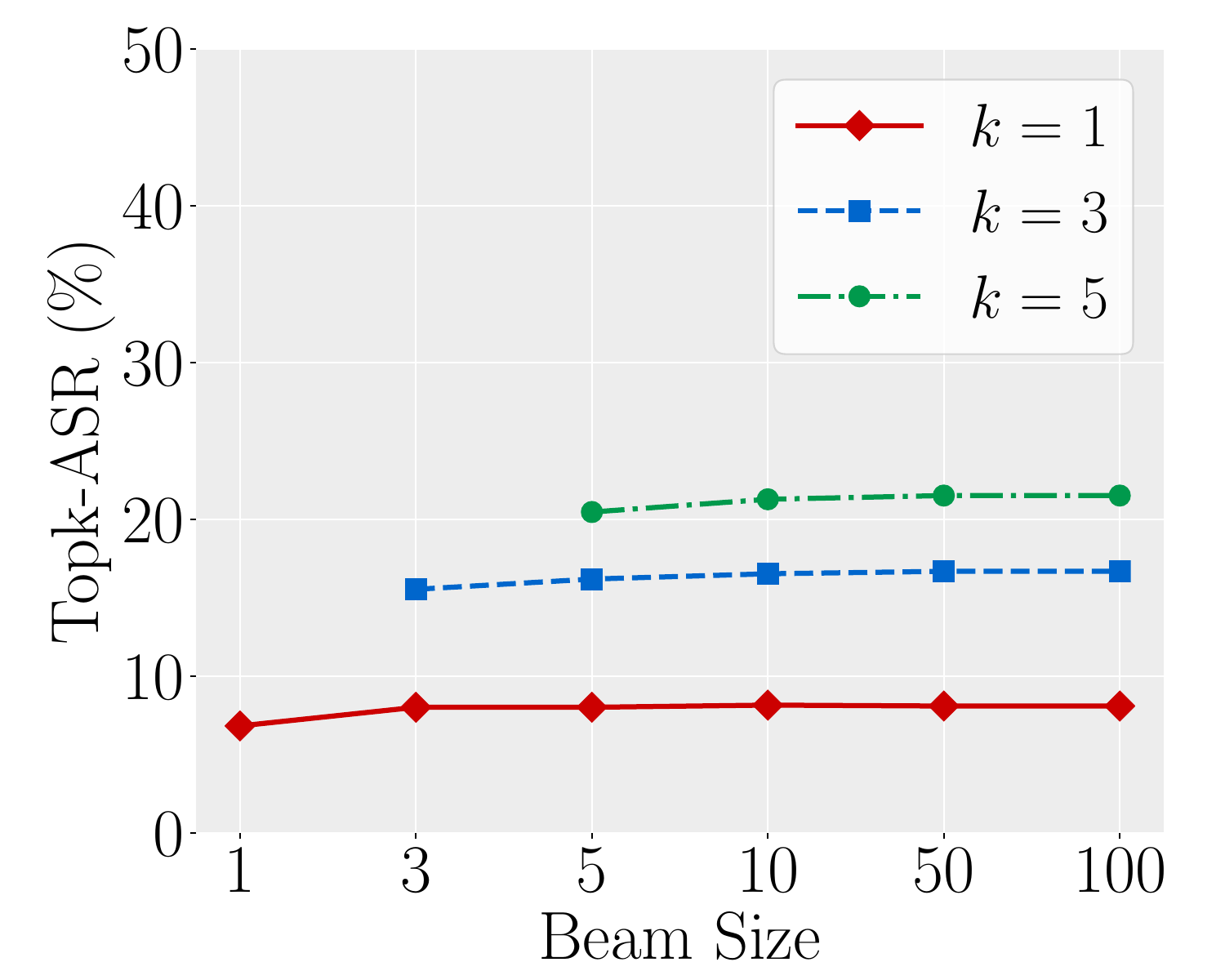}
        \label{fig:querylim_traj_extraction}
        }
        \subfigure [\textsc{LocExtract} (\textsc{Gowalla})] {
            \includegraphics[width=0.22\textwidth]{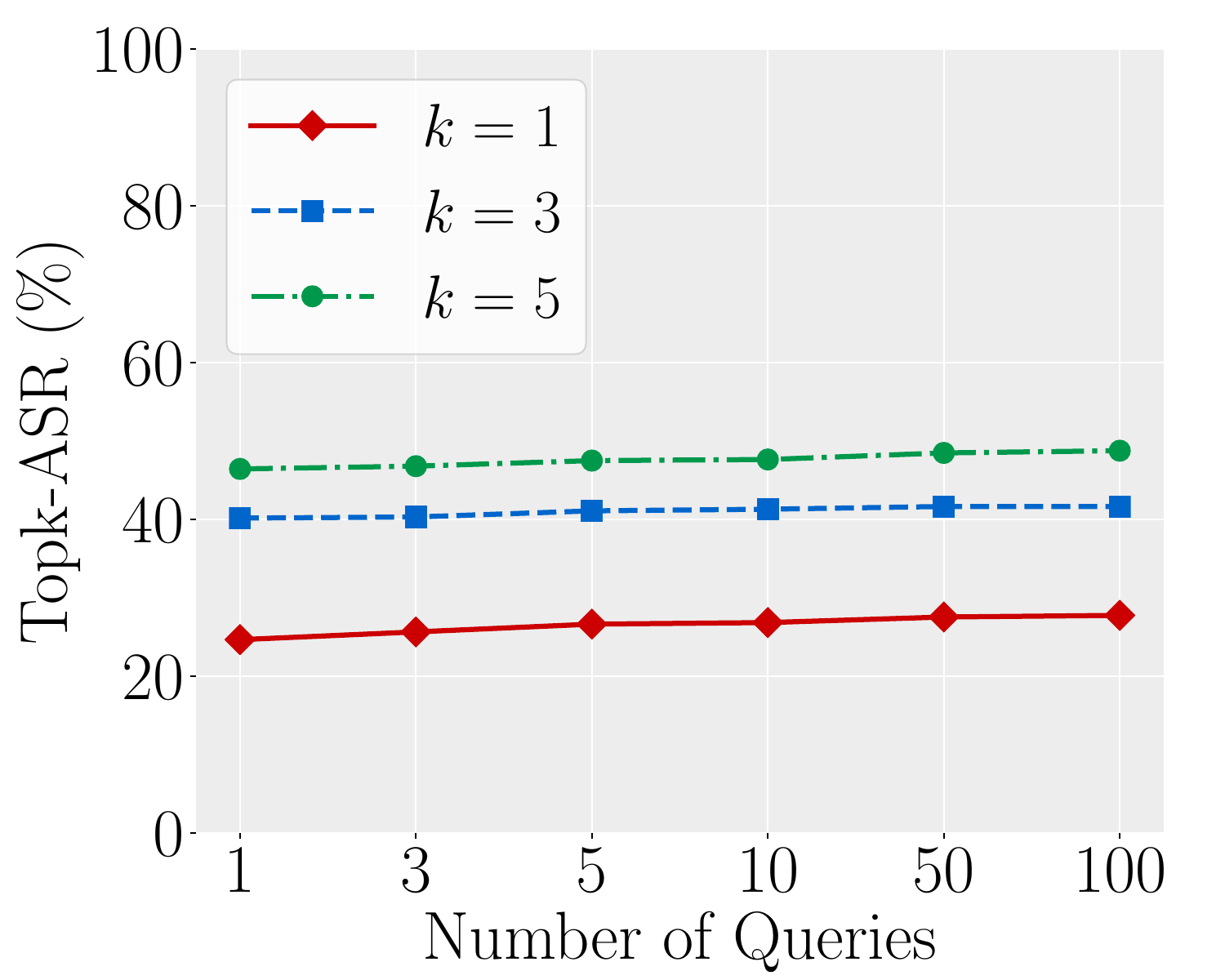}
            \label{fig:querylim_loc_extraction_g}
        }
        \subfigure [\textsc{TrajExtract}~(\textsc{Gowalla})] {
            \includegraphics[width=0.22\textwidth]{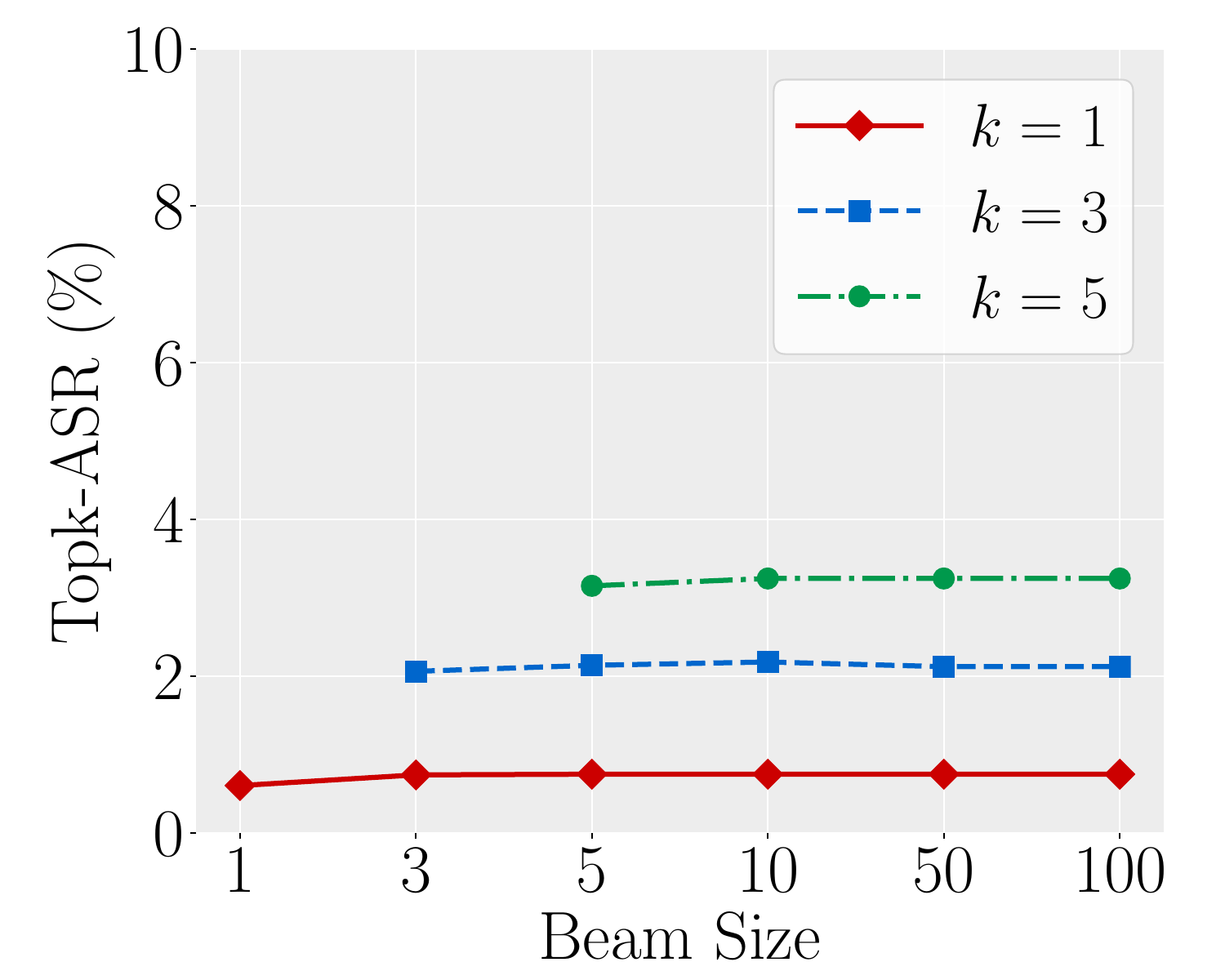}
            \label{fig:querylim_traj_extraction_g}
        }
        \vspace{-2mm}
        \caption{Our \textsc{LocExtract} is effective with a small number of queries and \textsc{TrajExtract} is effective with a small beam size (i.e., both attacks are effective within a small query budget).}
        \label{fig:querylim_extraction}
\end{figure*}

\noindent
\textbf{Data extraction attacks are effective given a limited number of queries}
In a realistic attack scenario, the adversary may encounter query limitations imposed by the victim model, allowing the adversary to query the model for only a certain number of queries. Figure~\ref{fig:querylim_extraction} illustrates that our data extraction attacks are effective given a limited number of queries. For example, as shown in Figure~\ref{fig:querylim_loc_extraction}, a mere $q=50$ query is sufficient for the adversary to achieve a high ASR and infer a user's frequently visited location. In terms of \textsc{TrajExtract} attack (Figure~\ref{fig:querylim_traj_extraction}), the adversary can opt for a small beam width of $\beta=10$, requiring only 1000 queries to extract a trajectory of length $n=4$. This practicality of our data extraction attack holds true even when the query limit is very small.

\noindent
\textbf{Appropriate query 
 timestamp improves the effectiveness of data extraction attacks}
POI recommendation models rely on temporal information to make accurate location predictions. However, obtaining the same timestamps as training for attack can be challenging and is an unrealistic assumption.
Therefore, in our data extraction attack setup, we set the query timestamp to $t=0.5$ (i.e., the middle of the day). 

\begin{figure}[!t]
        \centering
        \subfigure [\textsc{LocExtract} (\textsc{Gowalla})] {
            \includegraphics[width=0.22\textwidth]{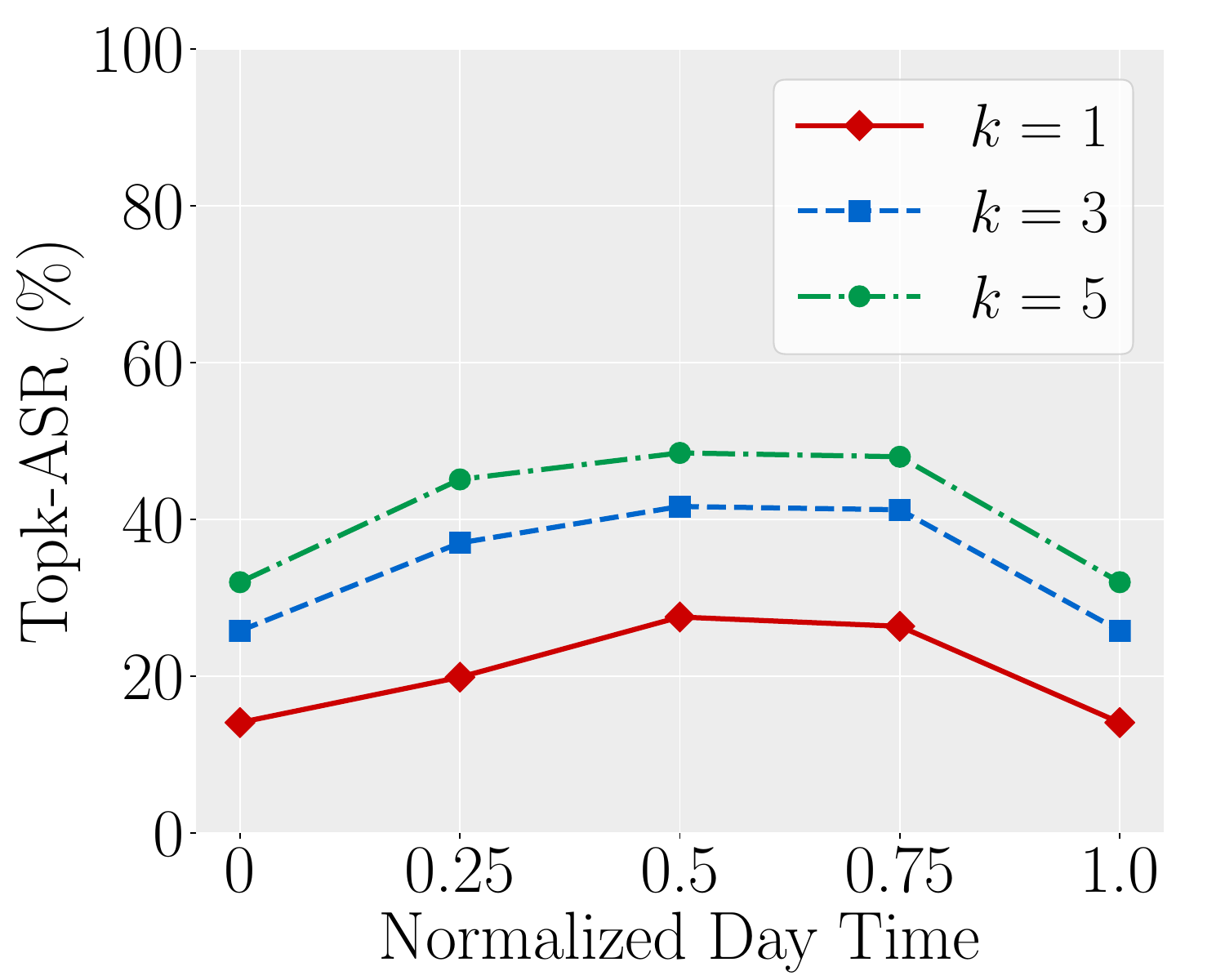}
        }
        \subfigure [\textsc{TrajExtract} (\textsc{Gowalla})] {
            \includegraphics[width=0.22\textwidth]{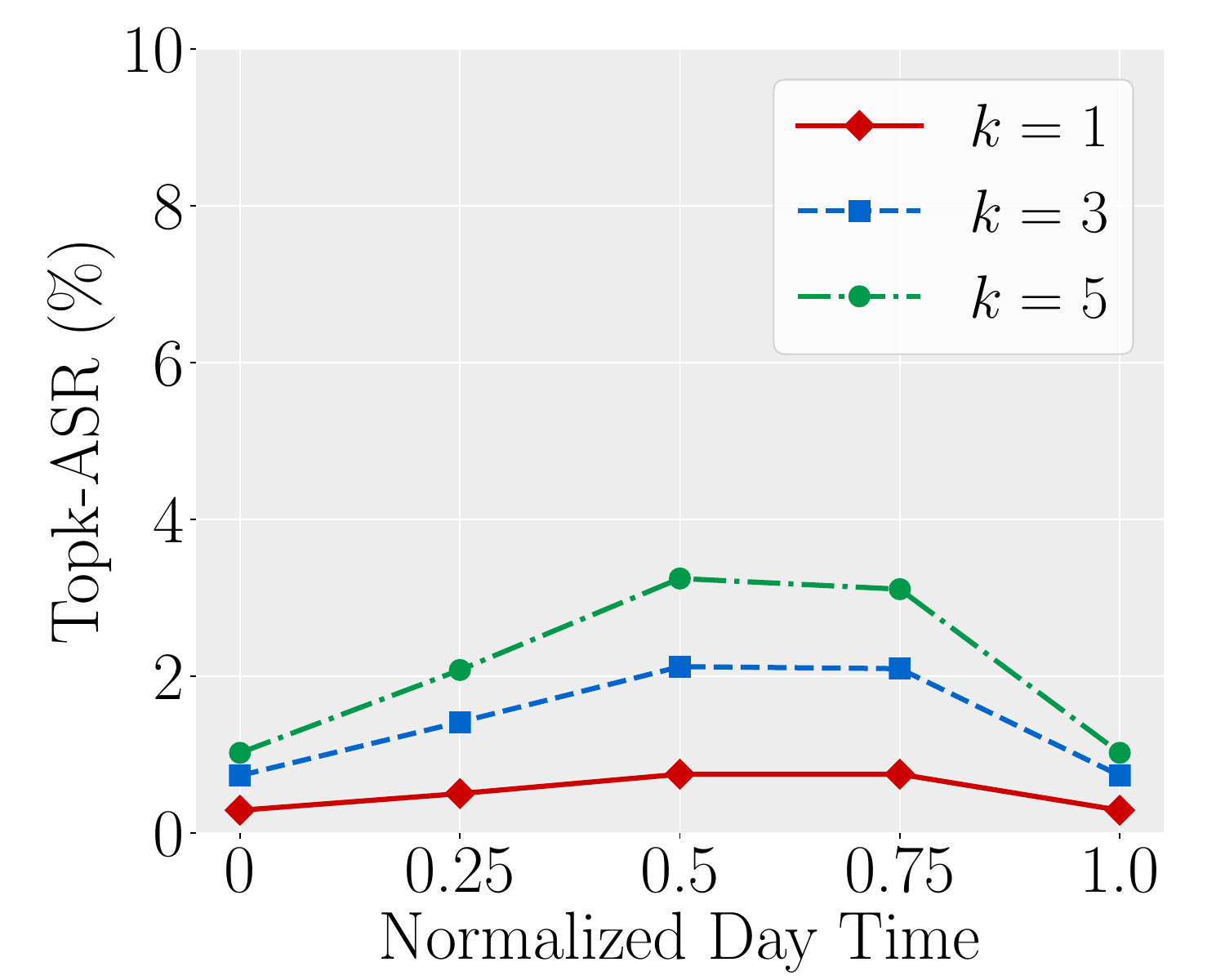}
        }
        \vspace{-2mm}
        \caption{The optimal query timestamp can significantly improve the performance of \textsc{LocExtract} and \textsc{TrajExtract} (Gowalla).}
        \label{fig:dummytime_gow}
\end{figure}

To analyze the effect of how different query timestamps affect data extraction attack performance, we conduct extraction attacks and vary different timestamps that represent various sections within a 24-hour window in the experiments. 
The results, illustrated in Figure~\ref{fig:dummytime} and Figure~\ref{fig:dummytime_gow}, indicate that utilizing timestamps corresponding to common check-in times, such as the middle of the day or late afternoon, yields better attack outcomes.
This finding aligns with the rationale that users are more likely to engage in check-ins during the daytime or after work hours. %

\begin{figure}[!t]
    \centering
    \subfigure [\textsc{4sq}] {\includegraphics[width=0.22\textwidth]{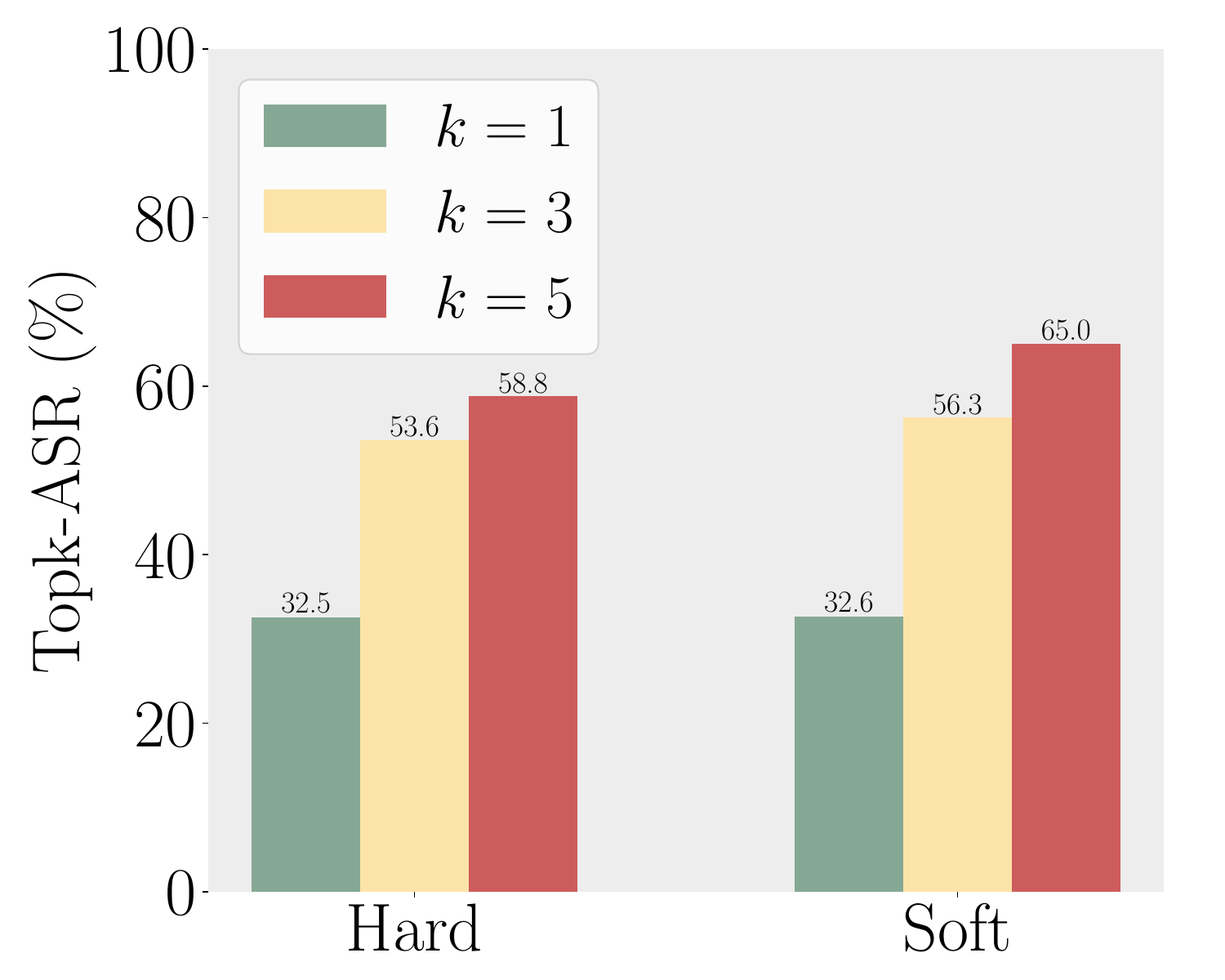}}
    \subfigure [\textsc{Gowalla}] {\includegraphics[width=0.22\textwidth]{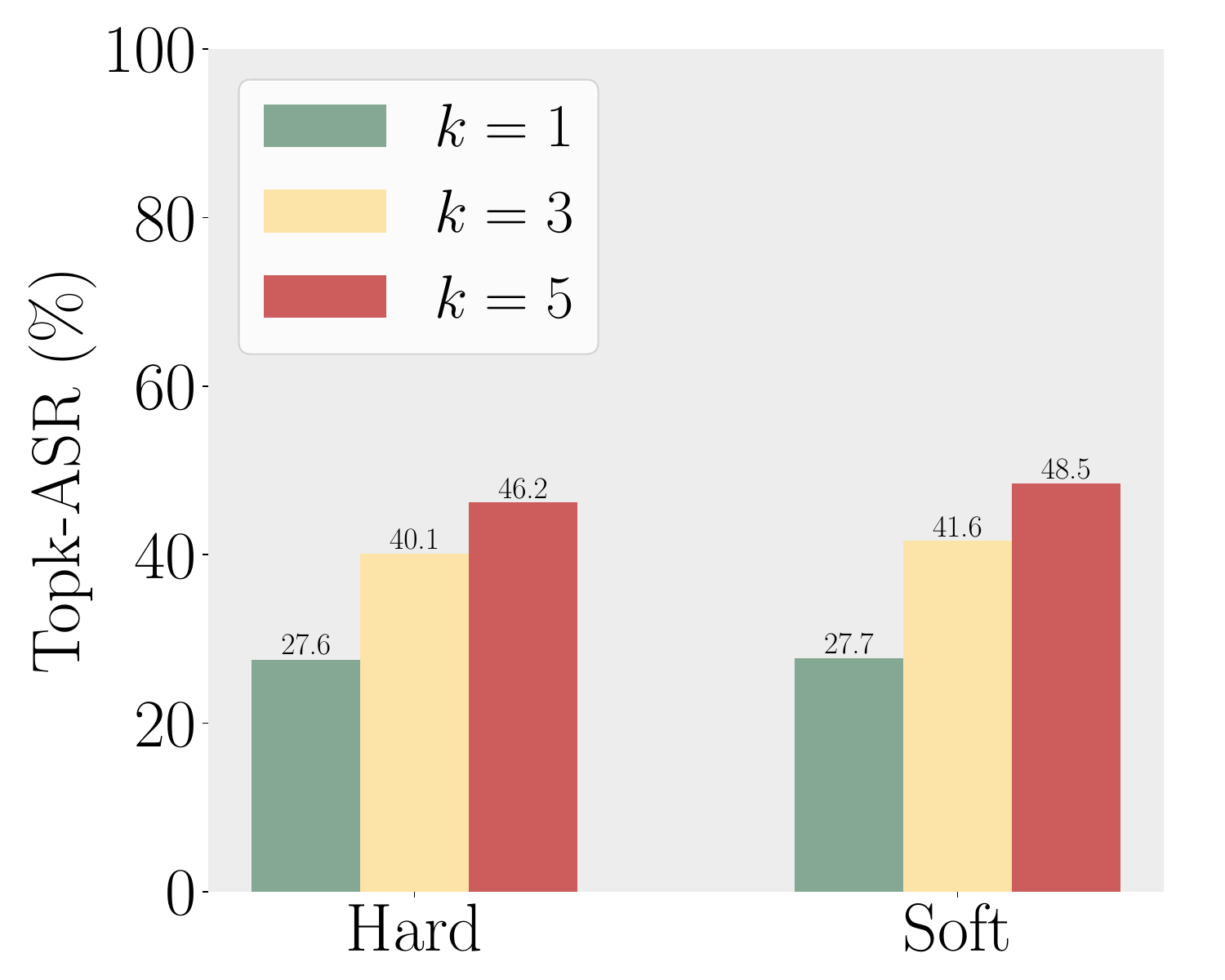}}
    \vspace{-2mm}
    \caption{Comparing soft voting with hard voting for logits aggregation in \textsc{LocExtract}. Soft voting has larger improvements over hard voting as $k$ increases.}
    \label{fig:hard}
\end{figure}
\begin{figure}[h]
        \centering
        \subfigure [\textsc{4sq}]{\includegraphics[width=0.22\textwidth]{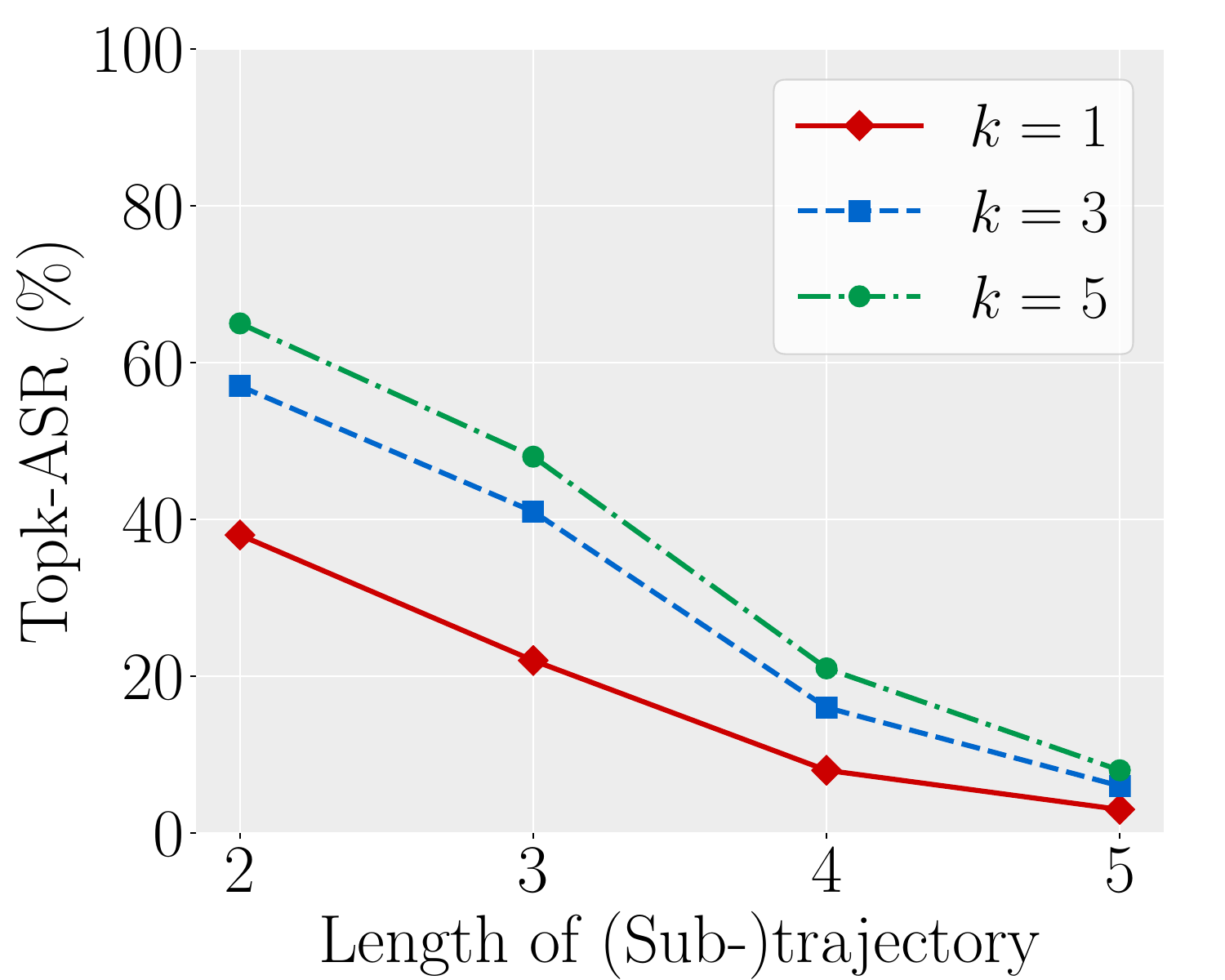}}
         \subfigure [\textsc{Gowalla}]{\includegraphics[width=0.22\textwidth]{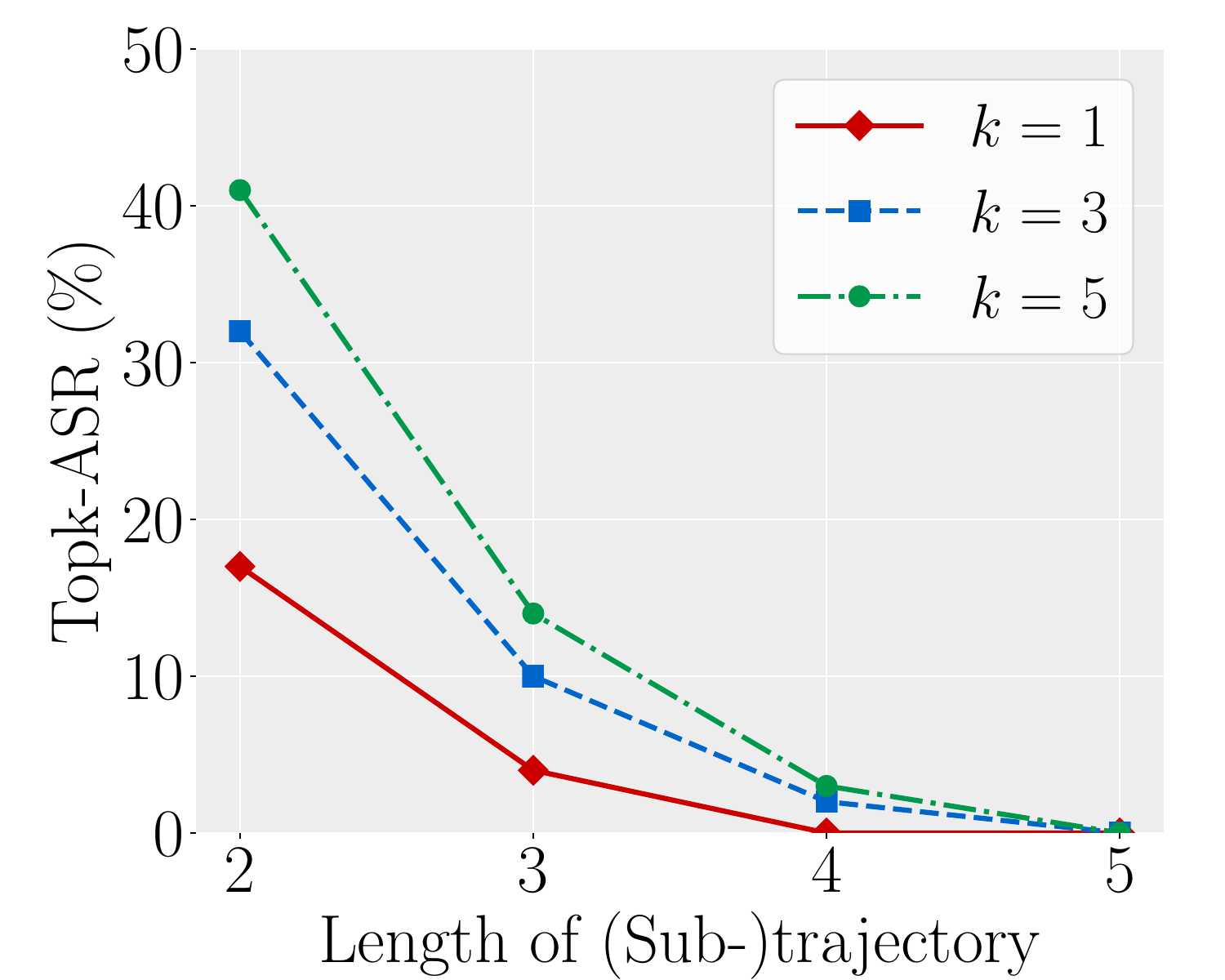}}
         \vspace{-2mm}
        \caption{The location sequences of shorter (sub-)trajectories are more vulnerable to \textsc{TrajExtract}.}
        \label{fig:trajlen-att2}
\end{figure}

\noindent
\textbf{Soft voting improves \textsc{LocExtract}}
For \textsc{LocExtract}, we have the option to employ either hard voting or soft voting to determine the most frequently occurring location. Hard-voting ensembles make predictions based on a majority vote for each query while soft-voting ensembles consider the average predicted probabilities and select the top-k locations with the highest probabilities. From the experimental results depicted in Figure \ref{fig:hard}, we observe that there is not a substantial difference in ASR-1 when using hard voting or soft voting. However, employing soft voting yields better ASR-3 and ASR-5 results.

\noindent
\textbf{Location sequences of shorter trajectories are more vulnerable to \textsc{TrajExtract}}
For \textsc{TrajExtract}, we conduct an ablation study to extract trajectories of varying lengths $n$. The results, illustrated in Figure~\ref{fig:trajlen-att2}, indicate that the attack achieves higher ASR on shorter trajectories than longer ones. This observation can be attributed to our assumption that the attacker possesses prior knowledge of a starting location. As the prediction moves further away from the starting location, its influence on subsequent locations becomes weaker. Consequently, predicting locations farther from the starting point becomes more challenging, decreasing the attack's success rate for longer trajectories. Moreover, the extraction of long trajectories presents additional difficulties. With each step, the probability of obtaining an incorrect location prediction increases, amplifying the challenges the attack algorithm faces.

\begin{figure*}[h]
        \subfigure [\textsc{LocMIA} (\textsc{4sq})] {\includegraphics[width=0.22\textwidth]{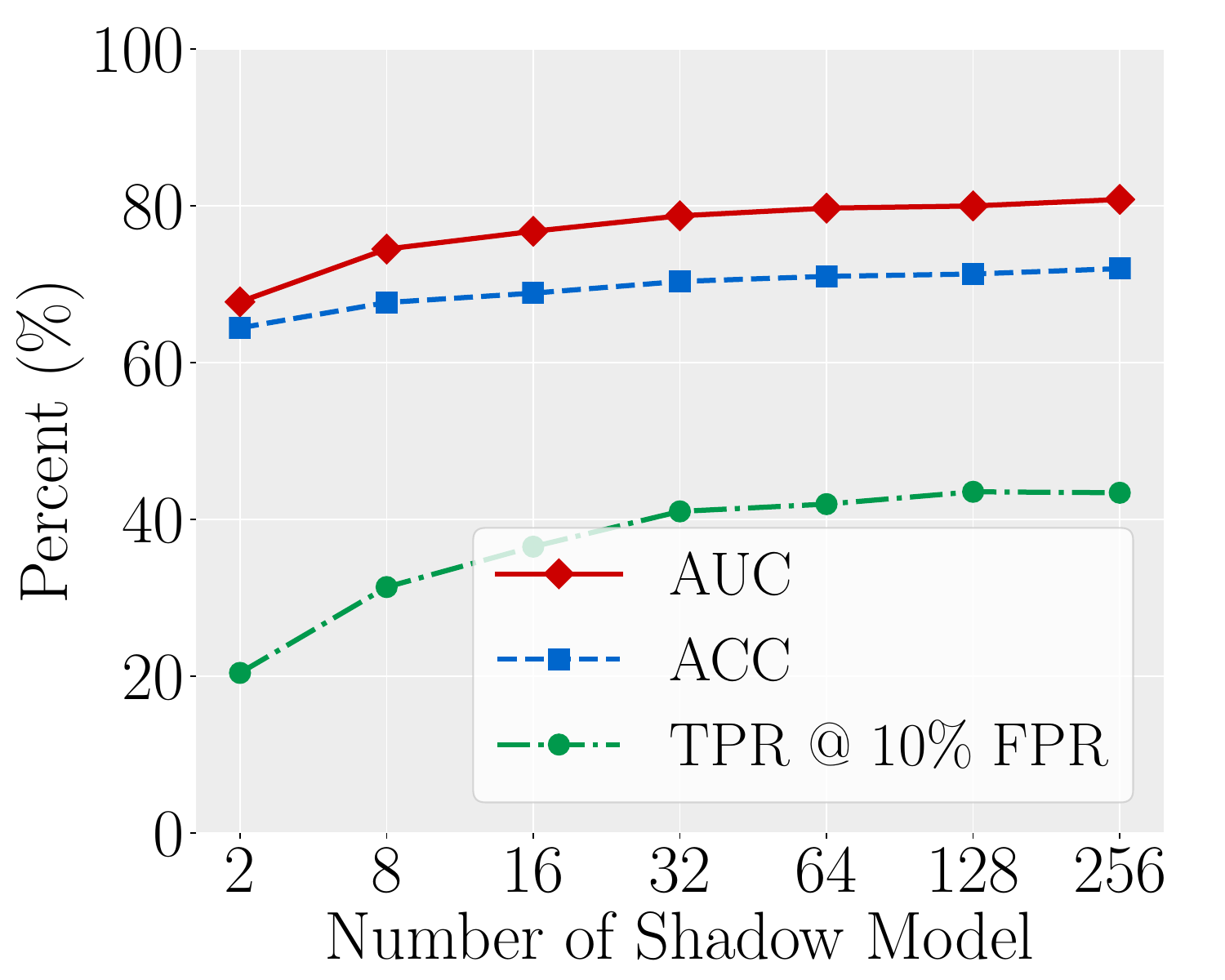}}
        \subfigure [\textsc{TrajMIA} (\textsc{4sq})]{\includegraphics[width=0.22\textwidth]{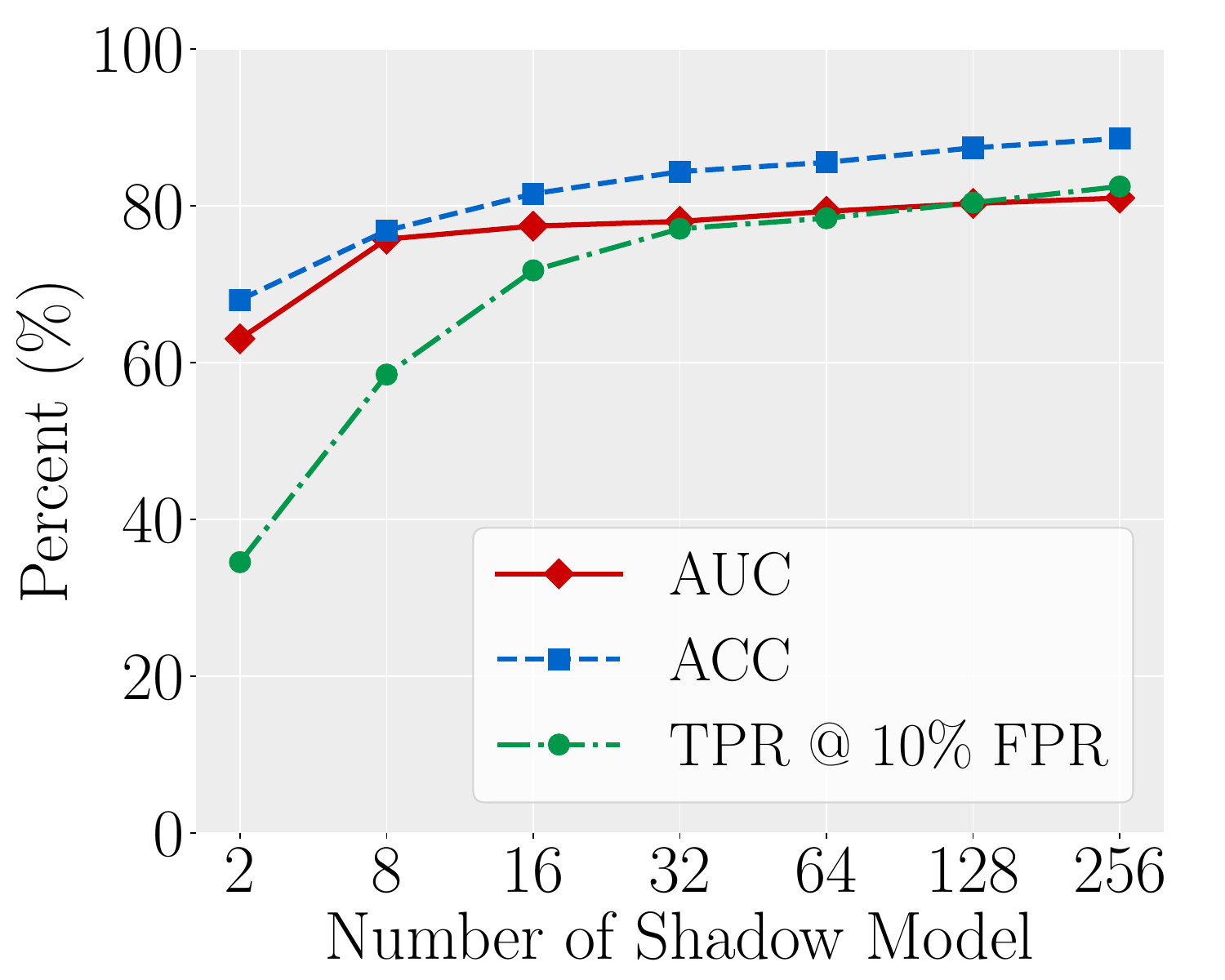}}
        \subfigure [\textsc{LocMIA} (\textsc{Gowalla})] {
            \includegraphics[width=0.22\textwidth]{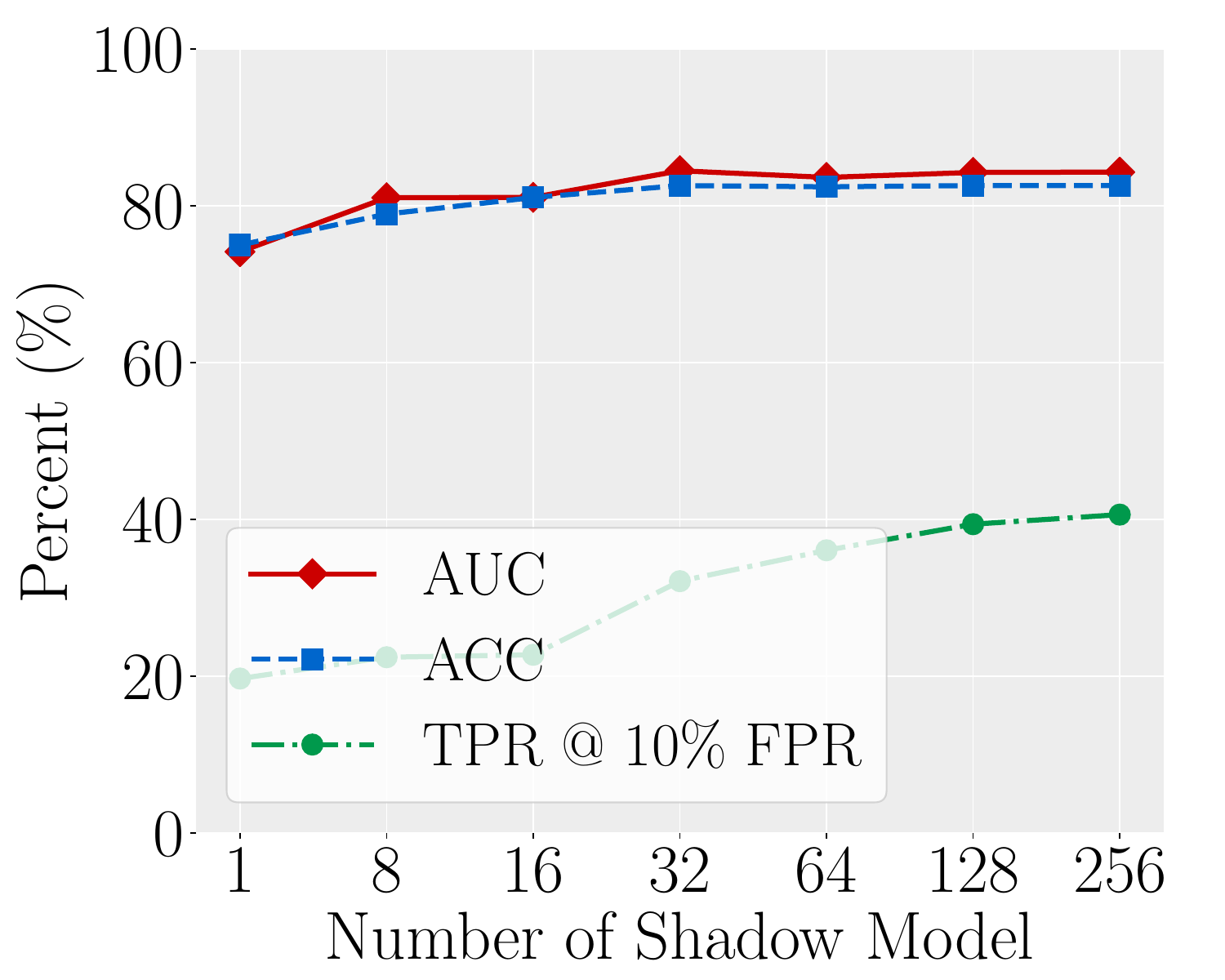}
        }
        \subfigure [\textsc{TrajMIA} (\textsc{Gowalla})] {
            \includegraphics[width=0.22\textwidth]{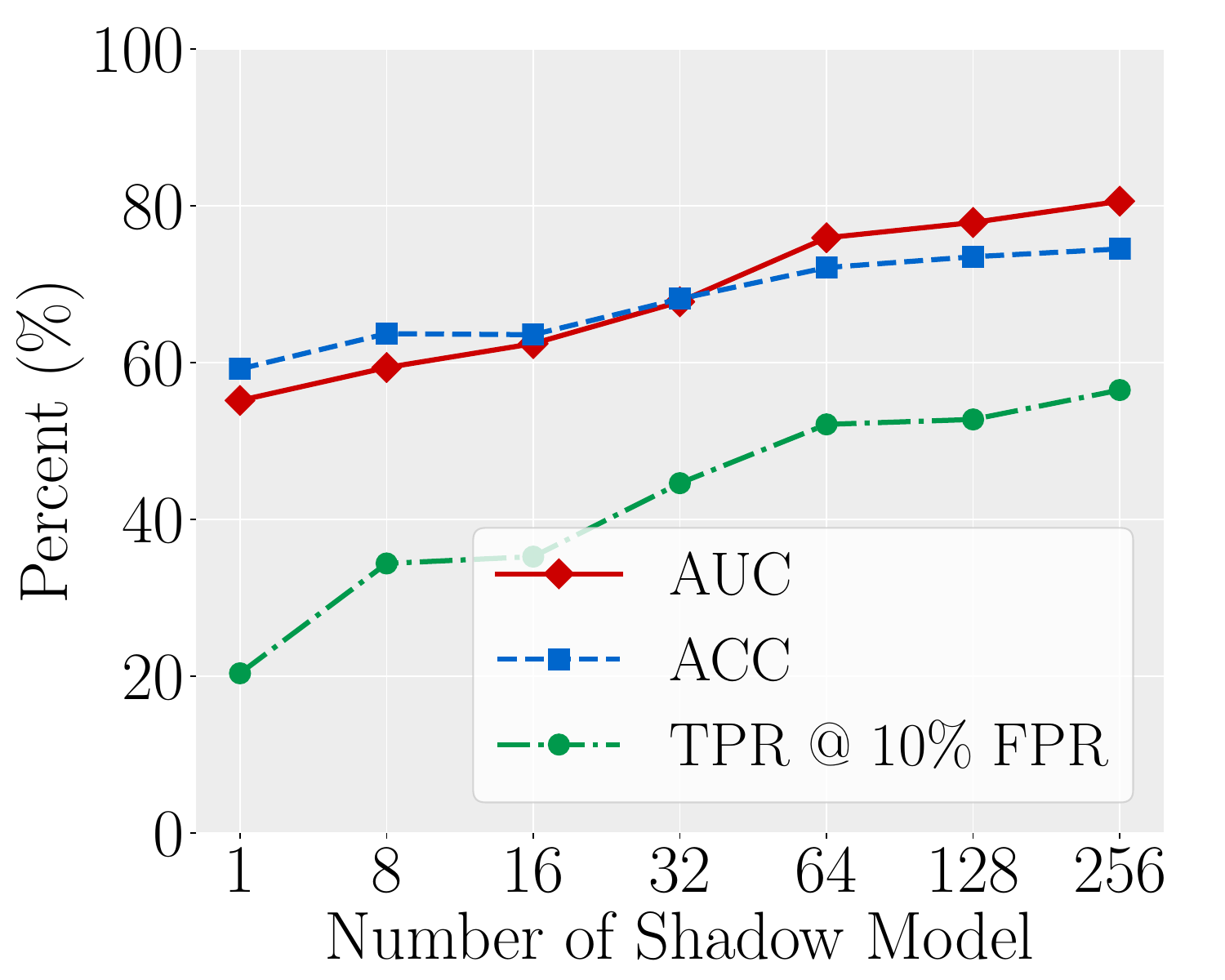}
        }
        \vspace{-2mm}
        \caption{The attack performance of \textsc{LocMIA} and \textsc{TrajMIA} significantly improves as the number of shadow model increases.}
        \label{fig:shadownum}
\end{figure*}

\subsection{Ablation Study on Our Membership Inference Attacks}
\label{Ablation:MIA}
\noindent
\textbf{A larger number of shadow models improves the effectiveness of MIAs}
As mentioned in~\cite{carlini2022membership}, it has been observed that the attack's performance of LiRA tends to improve as the number of shadow models increases. Consistently, our attacks also follow this pattern, as depicted in Figure~\ref{fig:shadownum}. Both location-level MIA and trajectory-level MIA show enhanced performance as we incorporate more shadow models. This improvement is because an increased number of shadow models allows for a better approximation of the distributions for $f_{in}$ and $f_{out}$, thereby simulating the victim model more accurately.

\noindent
\textbf{\textsc{LocMIA} is effective given a limited number of queries}
Since our \textsc{LocMIA} involves multiple queries to explore locations preceding the target location, as well as the corresponding timestamps, it is essential to consider potential limitations on the number of queries in real-world scenarios. Thus, we conduct experiments to investigate the impact of query limits on \textsc{LocMIA}. The results, depicted in Figure~\ref{fig:querylim_mia_loc}, indicate that our attack remains effective even with a limited number of queries for different location choices. The further increase in query locations would not significantly improve attack results.

We also conduct experiments with different settings for the number of query timestamps, denoted as $n_t$. The rationale behind this step is that %
the adversary does not possess information about the real timestamp used to train the victim model. To simulate the effect of selecting the correct timestamp, we perform experiments with varying timestamps to identify the timestamp that yielded the highest confidence score for the targeted location. Based on empirical observations from our experiment on the \textsc{4sq} dataset (see Figure ~\ref{fig:querylim_mia_time}), increasing the number of query timestamps tends to yield better overall results in practice.

\begin{figure}[t!]
    \centering
    \includegraphics[width=0.22\textwidth]{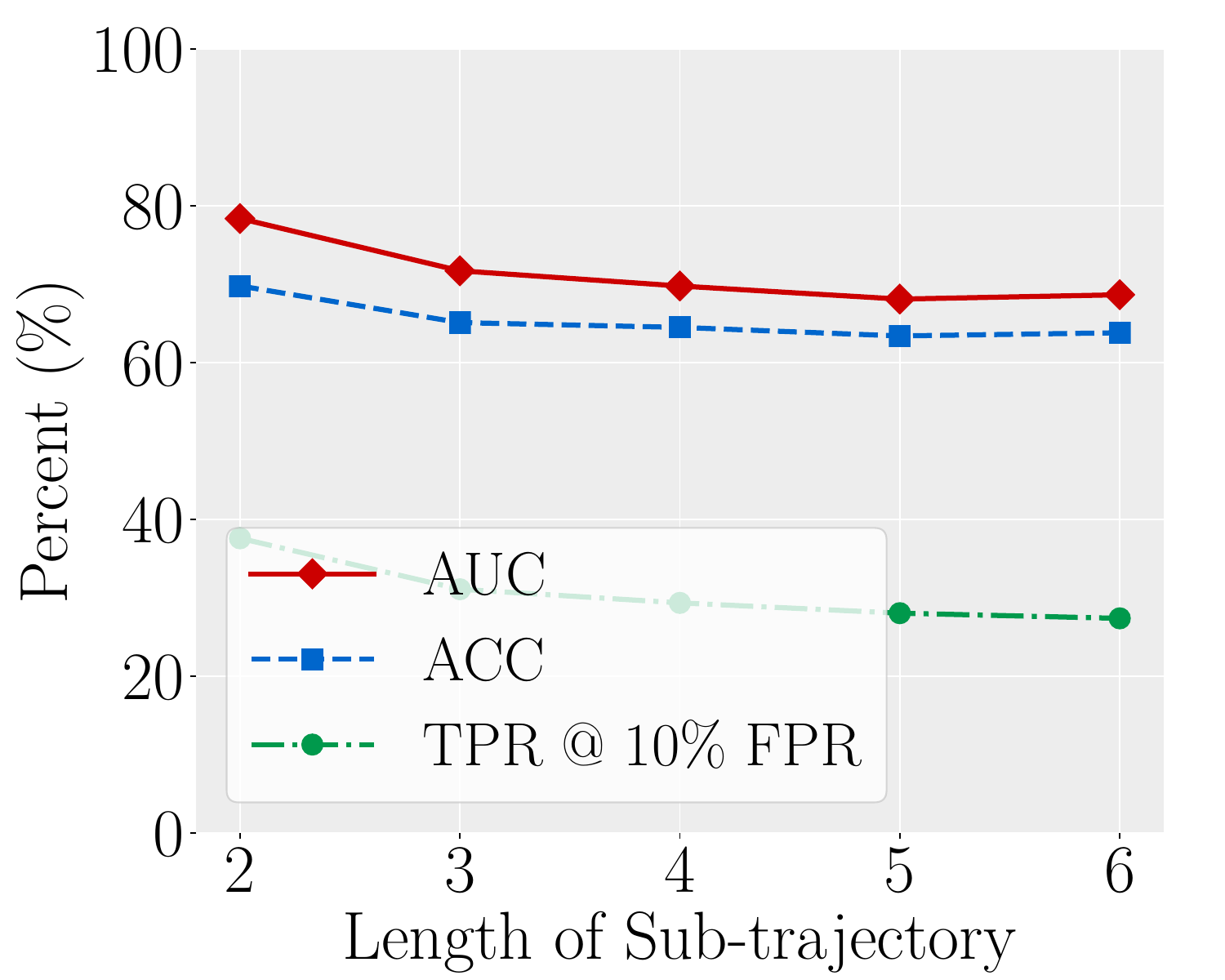}
    \vspace{-2mm}
    \caption{The longer trajectories are less vulnerable to \textsc{TrajMIA}.}
    \label{fig:trajmia_querylen}
\end{figure}

\noindent
\textbf{\textsc{TrajMIA} is less effective as the length of target trajectory increases}
From Figure~\ref{fig:trajmia_querylen}, we note that the attack performance drops as target trajectories are longer. This decline happens because all trajectory query scores influence the attack. Lengthier trajectories introduce increased randomness for the query, affecting the outcome of \textsc{TrajMIA}. \

\section{Defense} \label{sec:defenses}
We evaluate existing defenses against our privacy attacks, including standard techniques to reduce overfitting and differential privacy based defense. Detailed descriptions of defense techniques can be found in Section~\ref{sec:defense}. Then, in~\ref{defense_metric}, we describe the evaluation metrics to measure the defense performance from two perspectives.
In~\ref{defense_results}, we compare different defenses and analyze the numerical results with detailed explanations.

\subsection{Defense Metrics}
\label{defense_metric}
Our inference attacks extract different sensitive information about the training dataset from the victim model, as summarized in Table~\ref{sensitive_info}. To this end, we evaluate defense mechanisms in terms of their performance in preventing each attack from stealing the corresponding sensitive information. Specifically, we measure their defense performance on protecting \textit{all the sensitive information} and \textit{a targeted subset of sensitive information} for each attack, respectively. Here, we define the targeted subset of sensitive information as the mobility data a defender wants to protect in practice (e.g., some selected user-location pairs in \textsc{LocMIA}). We introduce this metric because not all the mobility data are sensitive or equally important. Take \textsc{LocMIA} as an example: Since the utility of POI recommendation is highly related to the model's memorization of user-location pairs, a user may want the model to recognize most of the POIs in his trajectory history while hiding those that are very likely to leak his personal identity (e.g., home). In other words, not all the mobility data need to be protected and it's more important to evaluate how defense mechanisms perform on the targeted subset of sensitive information.

To this end, we jointly measure the defense performance in protecting all the sensitive information and the targeted subset of sensitive information for each attack. Based on different attack objectives, we construct a different targeted subset of sensitive information for measurement by randomly sampling a portion of (e.g., 30\%) the most common locations in \textsc{LocExtract}, location sequences in \textsc{TrajExtract}, user-location pairs in \textsc{LocMIA} and trajectory sequences in \textsc{TrajMIA}. It is noted that we randomly sample 30\% of sensitive information in each attack to construct the targeted subset for the ease of experiments. In practice, the defender may have more personalized choices based on user-specific requirements, which we leave as future work.

\begin{figure*}[h!]
\centering
\subfigure[\textsc{LocExtract}] {\includegraphics[width=0.23\textwidth]{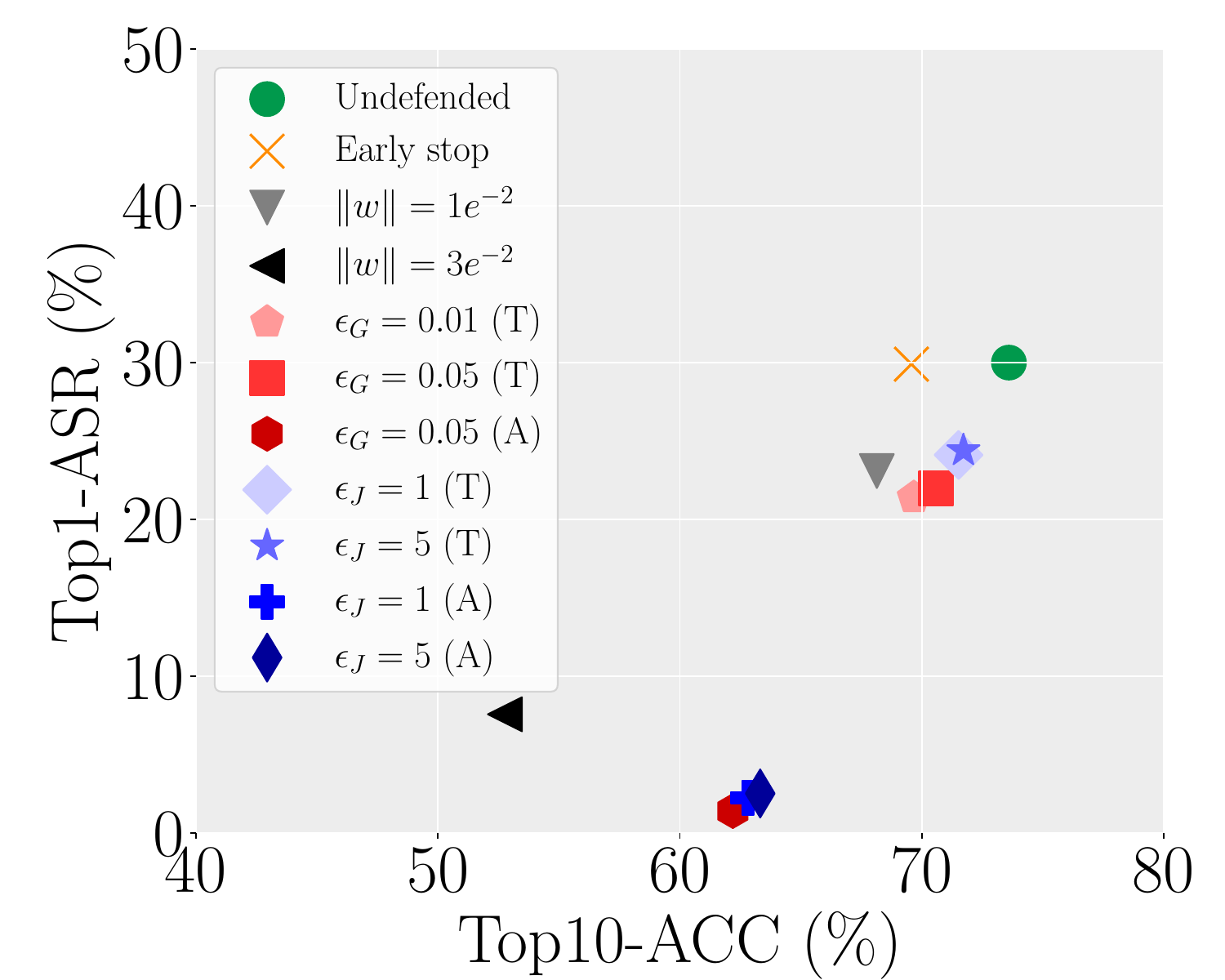}\label{attack1_a}}
\subfigure[\textsc{TrajExtract}] {\includegraphics[width=0.23\textwidth]{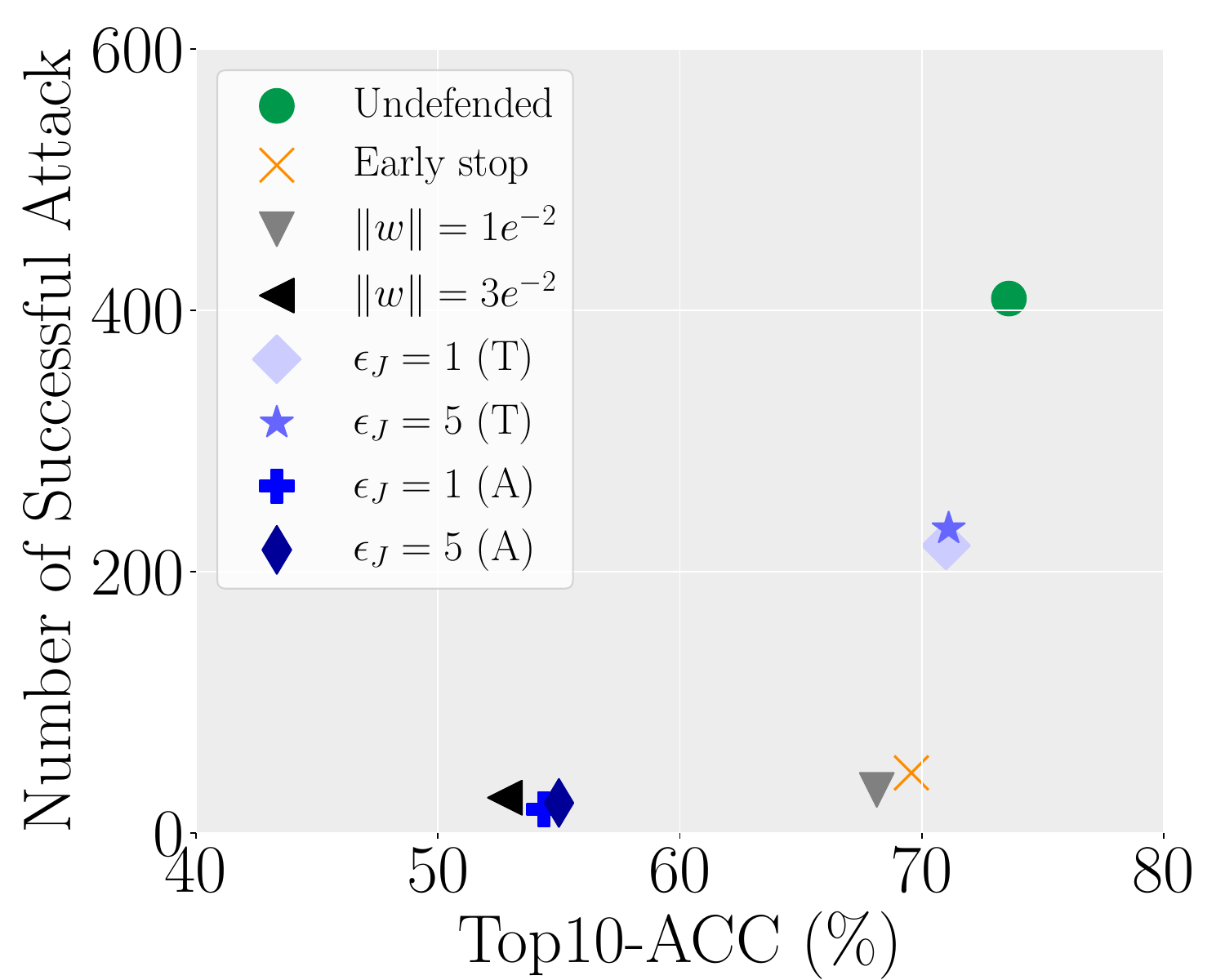}\label{attack2_a}}
\subfigure[\textsc{LocMIA}] {\includegraphics[width=0.23\textwidth]{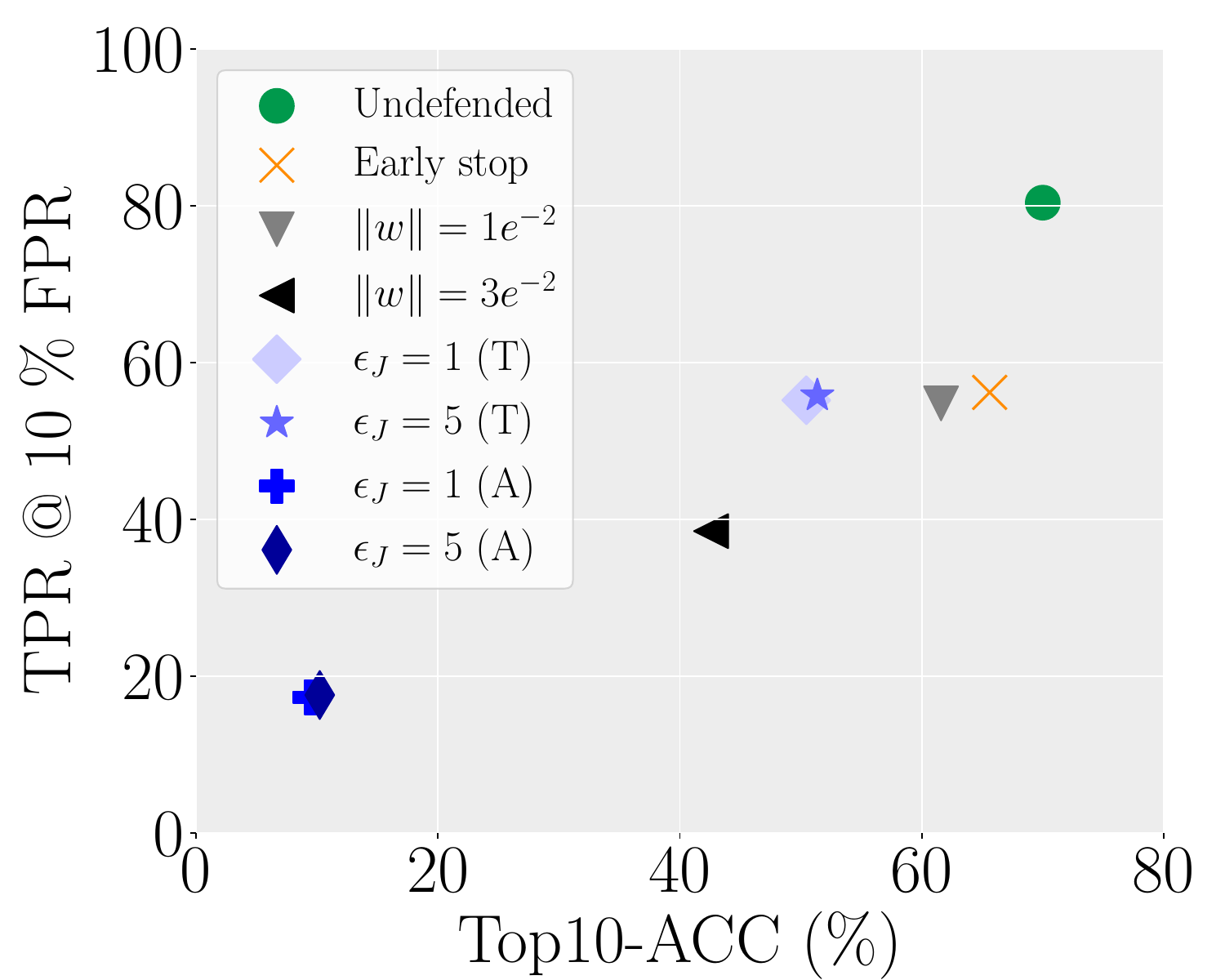}\label{attack3_a}}
\subfigure[\textsc{TrajMIA}] {\includegraphics[width=0.23\textwidth]{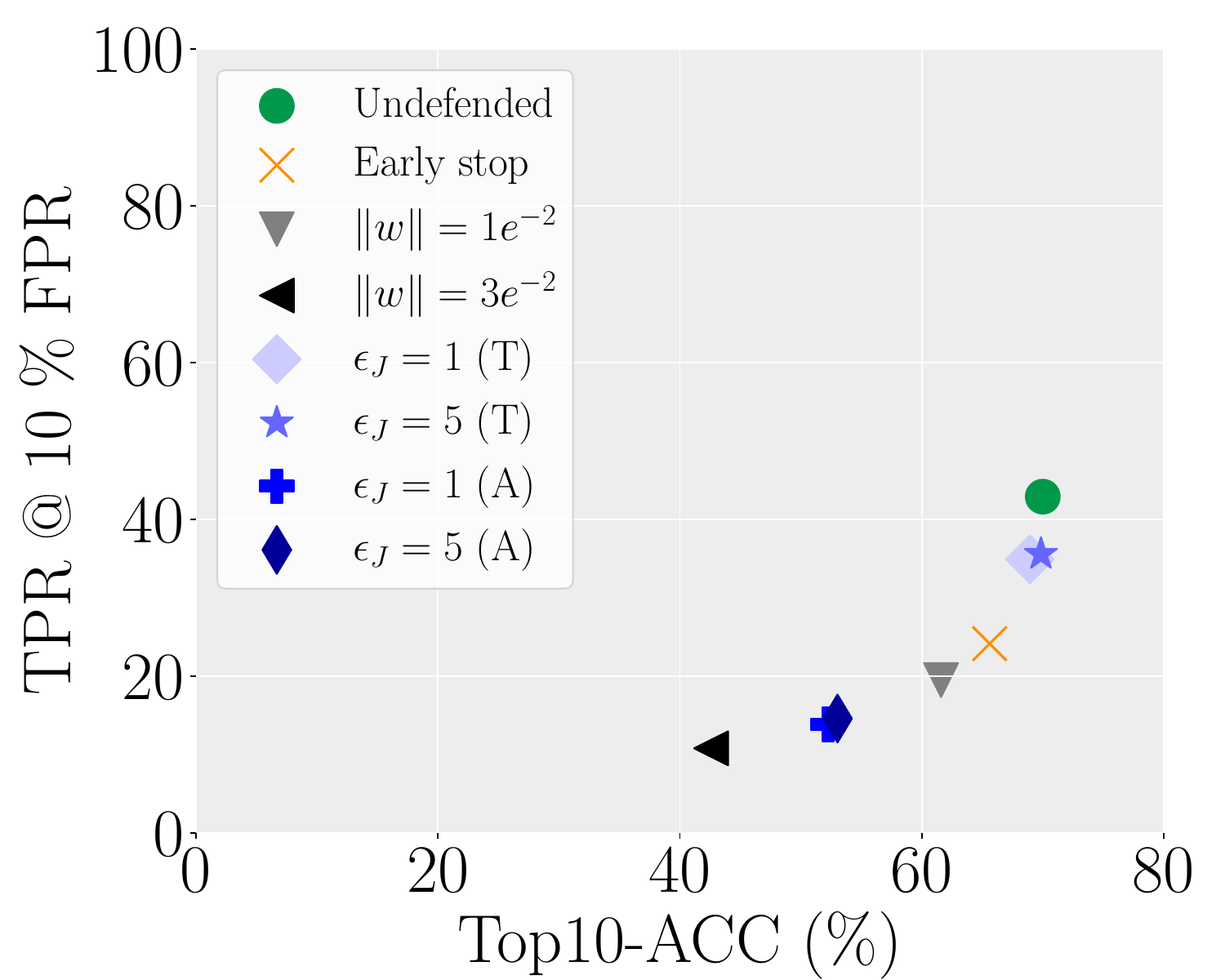}\label{attack4_a}}
\vspace{-2mm}
\caption{Defense performance on protecting all corresponding sensitive information for each attack.}
\label{defense}
\end{figure*}

\begin{figure*}[h!]
\centering
\subfigure[\textsc{LocExtract}]{\includegraphics[width=0.23\textwidth]{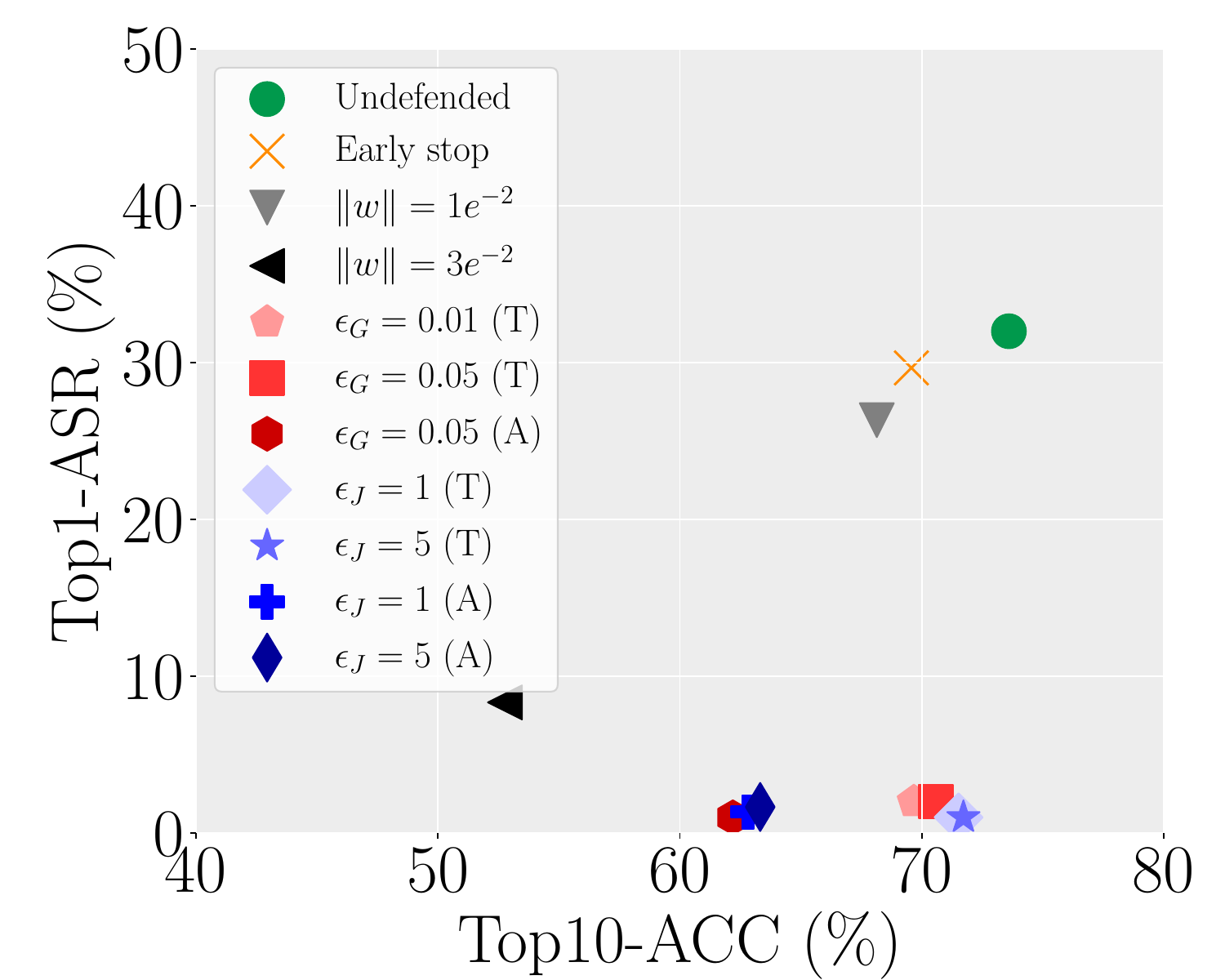}\label{attack1_t}}
\subfigure[\textsc{TrajExtract}] {\includegraphics[width=0.23\textwidth]{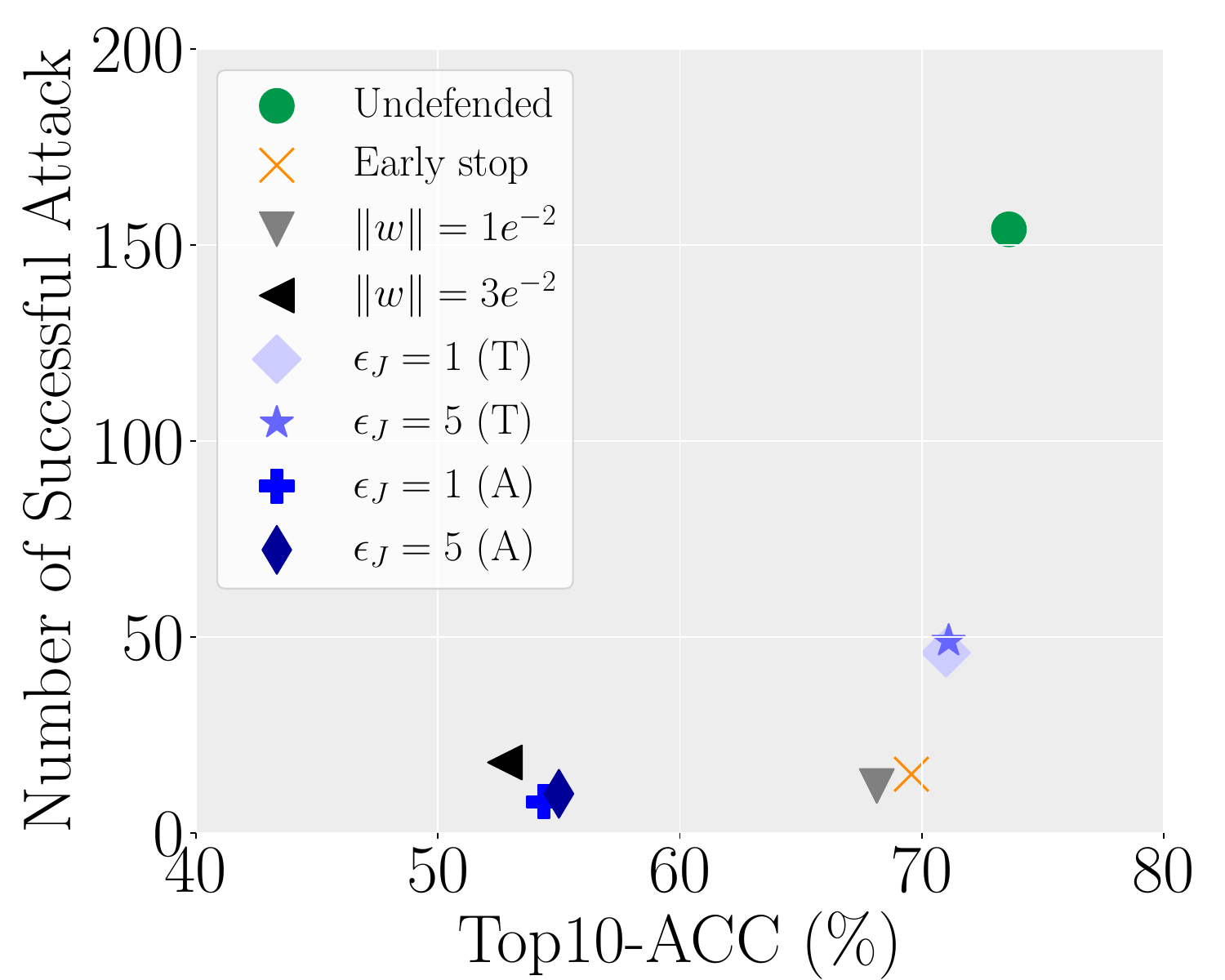}\label{attack2_t}}
\subfigure[\textsc{LocMIA}] {\includegraphics[width=0.23\textwidth]{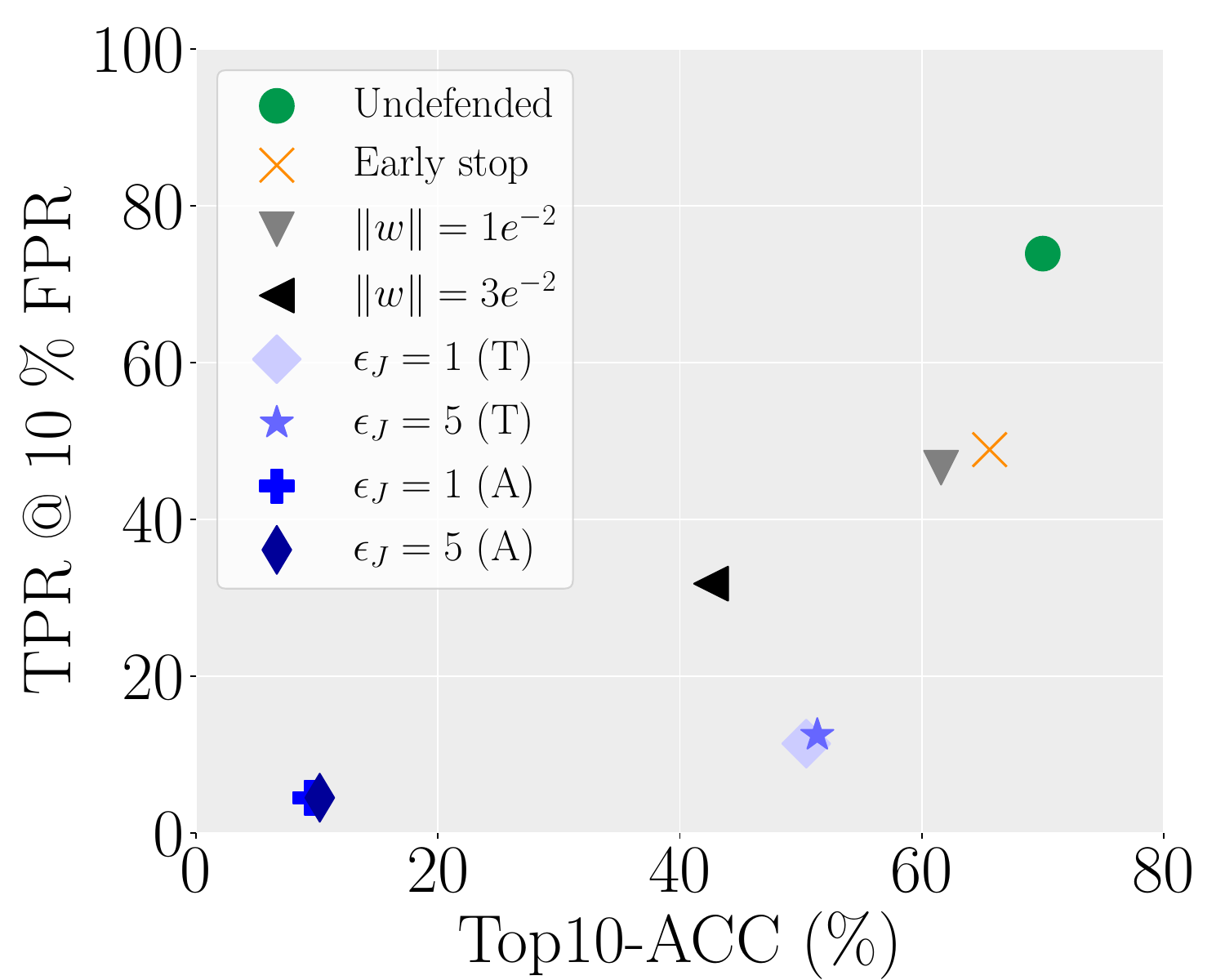}\label{attack3_t}}
\subfigure[\textsc{TrajMIA}] {\includegraphics[width=0.23\textwidth]{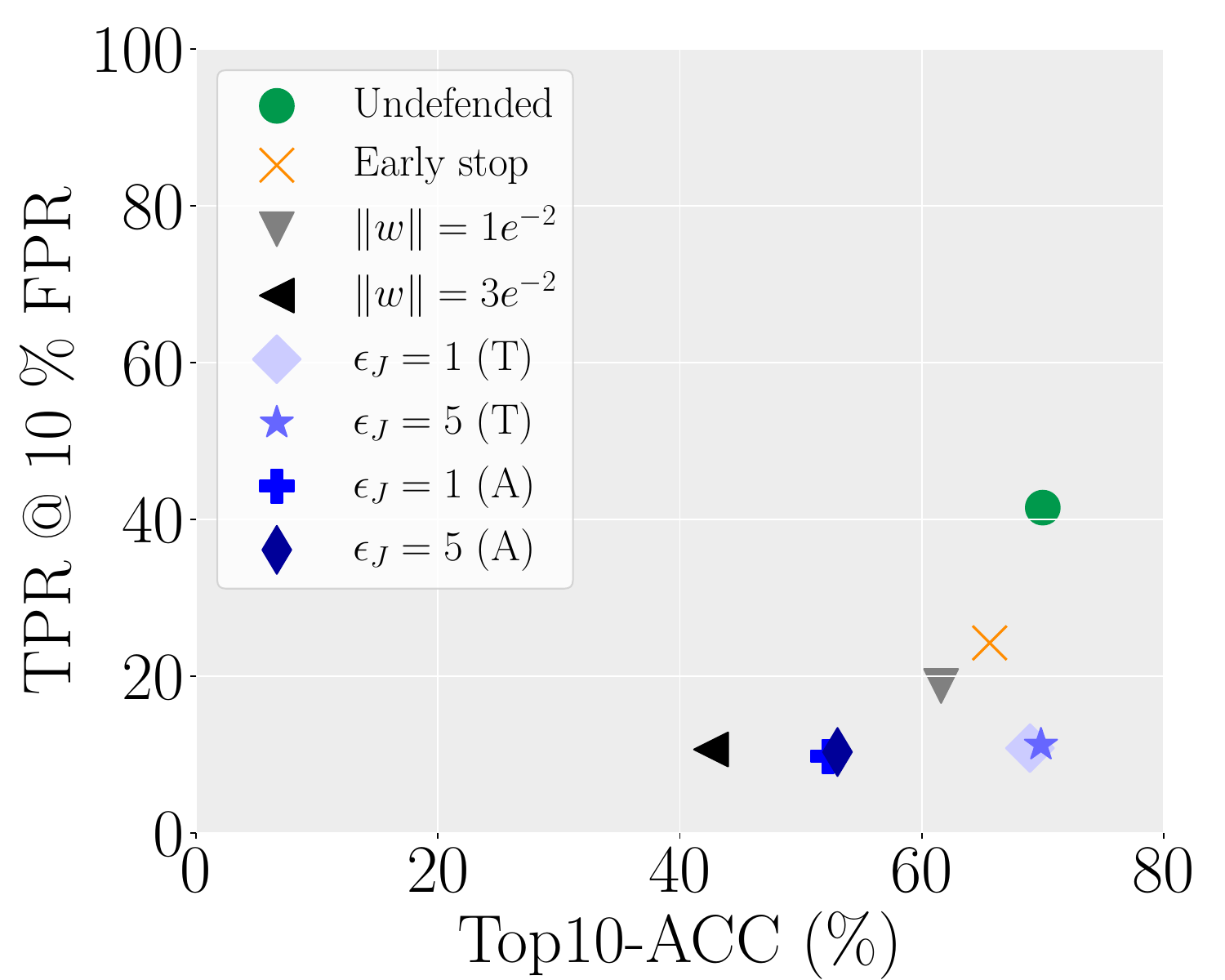}\label{attack4_t}}
\vspace{-2mm}
\caption{Defense performance on protecting the targeted subset of sensitive information for each attack.}
\label{defense2}
\end{figure*}

\subsection{Defense Setup and Results}
\label{defense_results}
\noindent
\textbf{Setup}
GETNext models trained on the 4sq dataset are used for experiments. For $L_2$ regularization, we use weight decay $\| w\| = 1e^{-2}$ and $3e^{-2}$. For early stopping, we stop the training after 5 epochs. For JFT, we mask sensitive information that needs to be protected in phase-I. Then in phase-II, we use DP-SGD~\citep{ACGMM16} with different $\epsilon_{J}$ (1 and 5) to finetune the model. The $C$ and $\delta$ are set to 10 and $1e^{-3}$. For Geo-Ind against \textsc{LocExtract}, we apply different $\epsilon_{G}$ (0.01 and 0.05) to replace each sensitive POI with its nearby location such that the original POI is indistinguishable from any location within $r=400$ meters. Since both JFT and Geo-Ind can be used to protect different amounts of sensitive information, we either protect nearly all the sensitive information or only the targeted subset of sensitive information for each attack, denoted by suffixes (A) and (T). 

\noindent
\textbf{\textsc{LocExtract}} Figure~\ref{attack1_a} shows the defense results on protecting all the sensitive information for \textsc{LocExtract}. From the figure, we observe that DP-based defenses achieve better performance than standard techniques. Both JFT (A) and Geo-Ind (A) reduce ASR from 30\% to 1\% with only a 10\% drop in utility. The reason is that these methods allow the defender to selectively protect common locations only. Besides, when protecting the same amount of sensitive information, JFT achieves slightly better accuracy than Geo-Ind because it involves phase-II training to further optimize the model. Moreover, although the ASR is still high for JFT (T) and Geo-Ind (T) in Figure~\ref{attack1_a}, we notice that they can substantially reduce the attack performance on the targeted subset of sensitive information, as shown in Figure~\ref{attack1_t}. This allows a defender to protect the targeted subset with negligible utility drop. {Figure~\ref{defense3} further shows that Geo-Ind can predict nearby locations of a protected POI as prediction results to maintain its usage.

\noindent
\textbf{\textsc{TrajExtract}} Figures~\ref{attack2_a} and~\ref{attack2_t} show that all the defenses can well protect location (sub-)sequences from being extracted by \textsc{TrajExtract}. This is because sequence-level extraction is a challenging task that pretty much relies on memorization.

\noindent
\textbf{\textsc{LocMIA}} Figures~\ref{attack3_a} and~\ref{attack3_t} show that none of the existing defenses can be used to protect user-location membership information. While JFT (A) reduces the TPR@10\%FPR  to less than 20\%, it significantly sacrifices the model's utility. The reason is that the defender needs to redact a large number of user-location pairs so as to protect them. As a result, the model may learn from wrong sequential information in the phase-I training, leading to a large utility drop. Even for protecting the targeted subset with 30\% of total user-location pairs only, there's still a 20\% drop in utility.

\noindent
\textbf{\textsc{TrajMIA}} Figures~\ref{attack4_a} and~\ref{attack4_t} show the utility-privacy trade-off of different defenses against \textsc{TrajMIA} We notice that JFT (T) can effectively mitigate the MIA on the targeted subset of trajectory sequences with a small degradation in accuracy. However, the utility drop is still large if a defender aims to protect the membership information of all trajectory sequences. 

\noindent
\textbf{Summary} Existing defenses provide a certain degree of guarantee in mitigating the privacy risks of ML-based POI recommendations. However, it is still challenging to remove all the vulnerabilities within a reasonable utility drop. This is because existing POI recommendation models heavily rely on memorizing user-specific trajectory patterns to make predictions, which lack semantic information as guidance. As a result, defense mechanisms such as DP-SGD can easily compromise the utility of the protected model due to the noises added to the gradients. Moreover, defenses such as JFT are not general for all inference attacks since each attack steals different sensitive information. To this end, our evaluation calls for more advanced mechanisms to defend against our attacks.

\subsection{Additional Definitions in Defense} 
\begin{definition}[$(\epsilon,\delta)$-DP]
A randomized mechanism $\mathcal{A}$ satisfies $(\epsilon,\delta)$-DP if and only if for any two adjacent datasets $D$ and $D^{\prime}$, we have:
$$ \forall \mathcal{O} \in Range(\mathcal{A}): Pr[\mathcal{A}(D) \in \mathcal{O}] \leq e^{\epsilon} Pr[\mathcal{A}(D^{\prime}) \in \mathcal{O}] + \delta$$
where $Range(A)$ indicates the set of all possible outcomes of mechanism $\mathcal{A}$ and $\delta$ indicates the possibility that plain $\epsilon$-differential privacy is broken.
\label{def_dp}
\end{definition}

\begin{definition}[geo-indistinguishability] 
A mechanism $\mathcal{A}$ satisfies $\epsilon$-geo-indistinguishability iff for all $l$ and $l^{\prime}$ , we have: 
$$d_\mathcal{P}(\mathcal{A}(l),\mathcal{A}(l^{\prime})) \leq \epsilon d(l,l^{\prime})$$
where $d$ denotes the Euclidean metric and $d_\mathcal{P}$ denotes the distance between two output distributions. Enjoying $\epsilon r$-privacy within $r$ indicates that for any $l$ and $l^{\prime}$ such that $d(l,l^{\prime}) \leq r$, mechanism $\mathcal{A}$ satisfies $\epsilon$-geo-indistinguishability.
\label{def_geo}
\end{definition}

\begin{figure}[t!]
\centering
\includegraphics[width=0.3\textwidth]{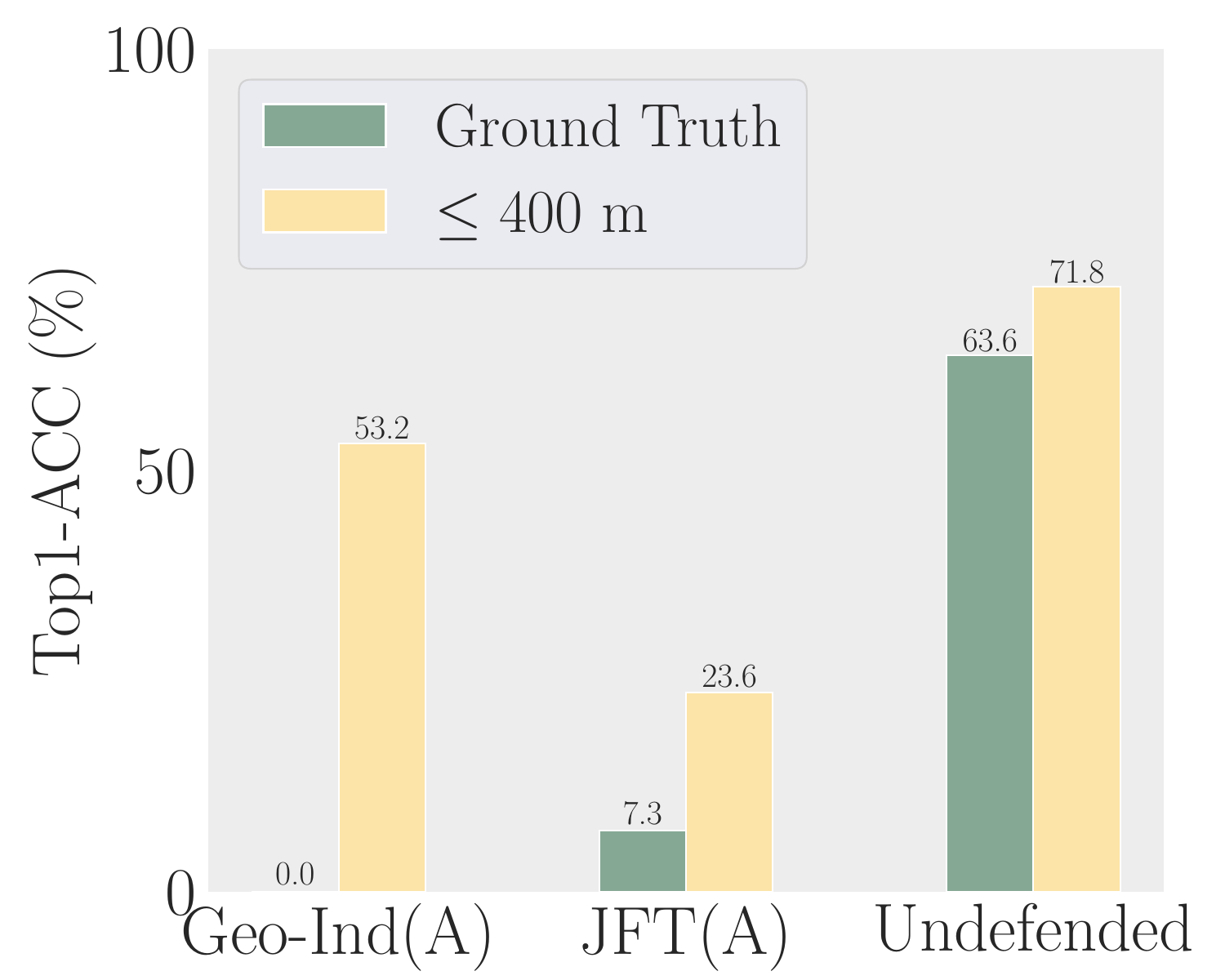}
\caption{For \textsc{LocExtract}, when ground truths are the most common locations, Geo-Ind (A) can predict a nearby location ($\leq 400$ m) of each protected location as the next POI with higher accuracy than JFT (A). The reason is that Geo-Ind applies the Laplacian mechanism to replace each protected POI with its nearby location.}
\label{defense3}
\end{figure}

\section{More Related Work on Defenses against Privacy Attacks}
\label{morerelated}

\noindent
\textbf{Defenses against privacy attacks on mobility data}
As discussed in Section~\ref{rel:mobility}, there have been various studies on stealing sensitive information from mobility data. Consequently, researchers have also explored various approaches to safeguard the privacy of mobility data, including K-Anonymity~\citep{gedik2005location,gruteser2003anonymous}, which aims to generalize sensitive locations by grouping them with other locations, Location Spoofing~\citep{bordenabe2014optimal,hara2016dummy}, which involves sending both real and dummy locations to deceive adversaries, Geo-indistinguishability~\citep{yan2022perturb,andres2013geo}, and local differential privacy (LDP)~\citep{xu2023efficient,bao2021successive}. However, these prior defense mechanisms primarily focus on data aggregation and release processes and can not be directly used in the context of POI recommendation. 
In contrast, our work is the first to concentrate on protecting against privacy breaches originating from deep learning models such as POI recommendation models and that's why we are primarily focusing on testing defense mechanisms related to the model training process as mentioned in Section~\ref{sec:defenses}.

\noindent
\textbf{Defenses against privacy attacks on deep neural networks}
There are also multiple works on protecting the privacy of deep learning models, with some notable examples including regularization and early stopping, which are commonly employed techniques to mitigate overfitting \citep{GBC16}. Another approach is differentially private stochastic gradient descent (DP-SGD) \citep{ACGMM16}, which achieves differential privacy by introducing noise during the gradient descent process while training the model. Additionally, selective differential privacy (S-DP) has been proposed to safeguard the privacy of specific subsets of a dataset with a guarantee of differential privacy \citep{SCLJY22,shi2022just}. However, these methods have primarily been tested on image or language-related models and require customization to fit into the usage of POI recommendation models. In our work, we focus on adapting these defense mechanisms to POI recommendation models by developing privacy definitions that are specifically tailored to the attacks we propose. In addition, we follow the concept of selective DP~\citep{shi2022just} to relax the original DP and selectively protect the sensitive information (e.g., most common locations) in mobility data.

\end{document}